%% file: main.tex
\pdfoutput=1

\documentclass[10pt,twocolumn,letterpaper]{article}
\usepackage[T1]{fontenc}
\usepackage[pagenumbers]{cvpr}
\makeatletter
\renewcommand\LenToUnit[1]{\strip@pt\dimexpr#1*\p@/\unitlength\relax}
\makeatother
\usepackage{graphicx}
\usepackage{amsmath}
\usepackage{amssymb}
\usepackage{booktabs}
\usepackage{colortbl}
\definecolor{oursrow}{HTML}{EAF2FB}
\newcommand{\oursrow}{\rowcolor{oursrow}}
\usepackage{multirow}
\usepackage{subcaption}
\usepackage{pifont}
\usepackage{microtype}
\definecolor{cvprblue}{rgb}{0.21,0.49,0.74}
\usepackage[pagebackref,breaklinks,colorlinks,allcolors=cvprblue]{hyperref}

\def\confName{CVPR}
\def\confYear{2027}

\newcommand{\cmark}{\textcolor{green!55!black}{\ding{51}}}
\newcommand{\xmark}{\textcolor{red!70!black}{\ding{55}}}

\title{Geo2Gcode: From 3D Geometry to CNC Programs through Structured Machining Primitives}
\author{
Sangmin Hong$^{1*}$\quad Dohee Cho$^{2*}$\quad Mohsen Yavartanoo$^{2}$\quad Kyoung Mu Lee$^{1,2}$\\
$^{1}$IPAI, Seoul National University\quad $^{2}$Dept. of ECE \& ASRI, Seoul National University\\
{\tt\small \{mchiash2, jdh12245, myavartanoo, kyoungmu\}@snu.ac.kr}
}

\newif\ifincludelegacyclaims
\includelegacyclaimsfalse
\begin{document}

\maketitle
{\let\thefootnote\relax\footnotetext{$^*$Equal contribution.}}

\begin{abstract}
Turning a 3D shape into a Computer Numerical Control~(CNC) program is still largely manual. For every part, a machinist chooses setups, tools, step-downs, and toolpath patterns. We present Geo2Gcode, which converts a target mesh directly into standard G-code through \emph{structured machining primitives}. Each primitive couples an indexed machining direction, a closed per-view reconstruction of the target, and explicit tool-radius, layer-height, and clearing-pattern parameters, and compiles deterministically into layerwise toolpaths via signed-distance queries. Most of the fidelity comes from this compiler. A lightweight selector trained with a voxel stock simulator fills in primitive parameters without per-part manual settings, approaching the finest fixed setting at about a quarter of its toolpath length. In simulation on ABC parts, Geo2Gcode reaches roughing fidelity comparable to an expert-tuned commercial CAM setup with a moderately longer toolpath, and clearly improves on a default CAM configuration. Generated programs run in a commercial G-code simulator, and one part is machined on a physical CNC mill. These results demonstrate that roughing programs can be generated directly from geometry without per-part CAM tuning.
\end{abstract}

\section{Introduction}
\label{sec:intro}

3D vision can now recover or generate shapes from images~\cite{Pixel2Mesh,onet,LRM}, scans~\cite{ACL-SPC,StarryGazer}, and text~\cite{Text2CAD}, yet turning such a shape into a physical part remains largely manual. Additive printing is increasingly automated, but subtractive Computer Numerical Control (CNC) machining, the standard process for durable metal parts, still requires expert programming. Reconstructing a 3D surface and producing a CNC program are different geometric tasks. A reconstruction specifies the desired boundary, whereas a machining program must specify how a finite-radius tool removes surrounding stock from selected directions. Surface regions hidden in one view, tool offsets, layer spacing, and the order of cutting segments all affect the resulting shape. In practice this translation is performed by a machinist in Computer-Aided Manufacturing (CAM) software, part by part. We ask whether it can be automated directly from target geometry, without first recovering a parametric Computer-Aided Design (CAD) model or feature tree.

\begin{figure*}[t]
    \centering
    \includegraphics[width=0.85\linewidth]{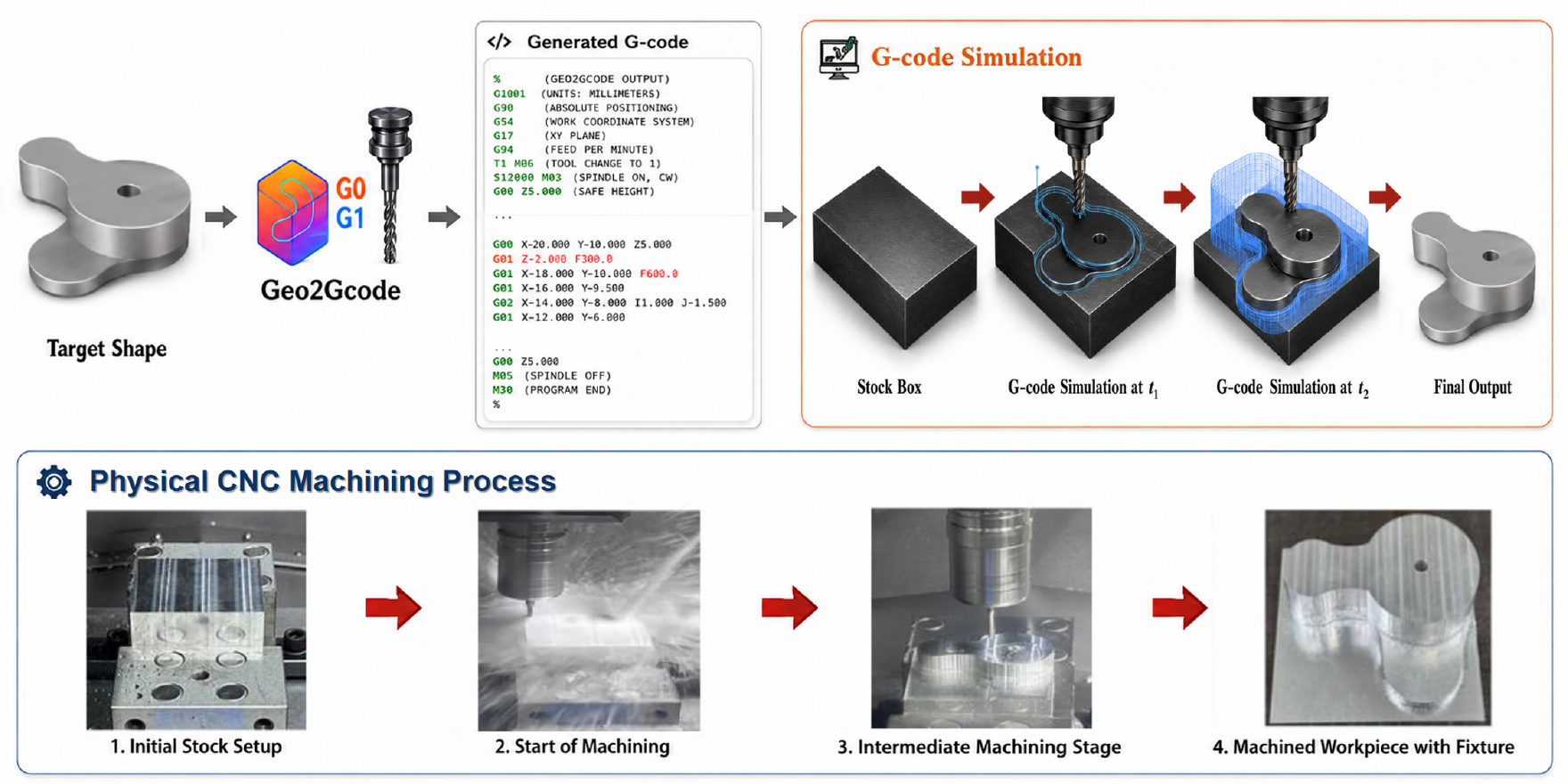}
    \caption{\textbf{Overview of our proposed pipeline.} From a target shape, our system generates G-code with initial setup and tool configuration, then inspects its stock-removal sequence in a CNC simulator. The bottom row shows a qualitative physical machining case, from stock setup to the machined workpiece.}
    \label{fig:pipeline}
\vspace{-3mm}
\end{figure*}

Learning-based shape-program methods either assemble CAD primitives~\cite{CAPRI-Net,SECAD-Net}, which still require a downstream CAM step, or, like CNC-Net~\cite{CNC-Net}, learn low-level cutting operations whose outputs are hard to relate to standard machining practice. Geo2Gcode takes a middle path built on \emph{structured machining primitives}. A machining primitive is the segment of a CNC program executed from one fixed tool direction. It specifies where to cut from, what to cut, and how to cut through (i) an indexed machining direction, (ii) a closed mesh of the target region visible along that direction, and (iii) the tool radius, layer height, and clearing pattern of a conventional layerwise program. A deterministic compiler turns each primitive into G-code, and a full program is a sequence of primitives, one per direction. A primitive is more than a template label, since its geometry fixes tool-center locations, its radius and layer height fix spatial sampling, and its pattern fixes the cutting order. Programs therefore stay machinist-readable and use only standard indexed 3+2-axis motion.

The remaining difficulty is choosing parameters. Because each view machines the stock left by earlier views, good choices depend on the whole sequence, and the number of candidate programs grows exponentially with the number of views. We replace per-part expert trial and error with a selector trained on simulator feedback that picks radius, layer height, and clearing strategy per view.

We summarize our main contributions as follows:
\begin{itemize}
\item \textbf{Structured machining primitives:} Each primitive bundles a direction, a closed per-view target mesh, and layerwise parameters, and compiles deterministically into conventional, machinist-readable G-code. This turns a target mesh into a multi-view CNC program without recovering a CAD model or manual CAM planning.
\item \textbf{Simulation-trained parameter selection:} A cumulative stock simulator scores multi-view programs, and a lightweight selector trained on its feedback fills in per-view parameters with no per-part manual settings. It outperforms random, greedy, and the reported constant settings, and only the finest setting is more accurate, with a 3.6$\times$ longer toolpath.
\item \textbf{Machining-oriented evaluation:} Against commercial CAM on the same meshes, Geo2Gcode reaches comparable simulated fidelity without manual input, and its programs run unmodified in CIMCO, an industrial G-code simulator that replays the controller-level code against a stock model with the actual cutter geometry.
\end{itemize}

\section{Related Work}
\label{sec:relatedworks}

\subsection{Reverse Engineering for 3D Shapes}
Deep learning has enabled primitive-based modeling, approximating shapes with cubes~\cite{Tulsiani_2017_CVPR, Zou_2017_ICCV, niu2018im2struct}, ellipsoids~\cite{genova2019learning}, or deformable primitives~\cite{Cvxnet, 3dias, Neural_parts, huang2023learning}, often using constructive solid geometry (CSG)~\cite{CSG,foley1996computer}. Reinforcement learning has also been applied to sequential primitive assembly~\cite{Csgnet, du2018, Brick-by-brick} and 3D printing~\cite{huang2024toolpath}.
Other approaches detect and fit primitives within point clouds~\cite{SPFN, ParseNet} or infer CSG programs. Recent works compress CAD models via CSG operations without ground-truth assemblies~\cite{UCSG, CSG-Stump, ExtrudeNet, CAPRI-Net, SECAD-Net}.
However, most methods focus on static reconstruction and overlook the sequential material removal essential for CNC machining, motivating frameworks that capture its stepwise nature.

\subsection{CNC Machining for Reverse Engineering}
Generating optimized G-code remains manual, requiring expert tuning of parameters and toolpaths. Classical setup planning selects orientations with visibility or accessibility analysis~\cite{Woo1994Visibility, zhong2025deepmill} but leaves toolpath parameters to the user. Learning-based Computer-Aided Process Planning (CAPP) with PSO or SVMs~\cite{HSIEH20133, DITTRICH201949} depends on pre-labeled CAD models and generalizes poorly.
Route planning methods~\cite{BALIC20021171, kukreja2022optimum} require extensive CAD-toolpath pairs, limiting generalization.
CNC-Net~\cite{CNC-Net} proposed a self-supervised framework that learns operations from 3D models without labeled operation sequences. Geo2Gcode instead restricts decisions to explicit machining primitives built from standard CAM patterns, so every learned decision maps to an interpretable program, and it evaluates the emitted G-code through cumulative material-removal simulation.

\begin{figure*}[t]
\centering
    \includegraphics[width=0.85\linewidth]{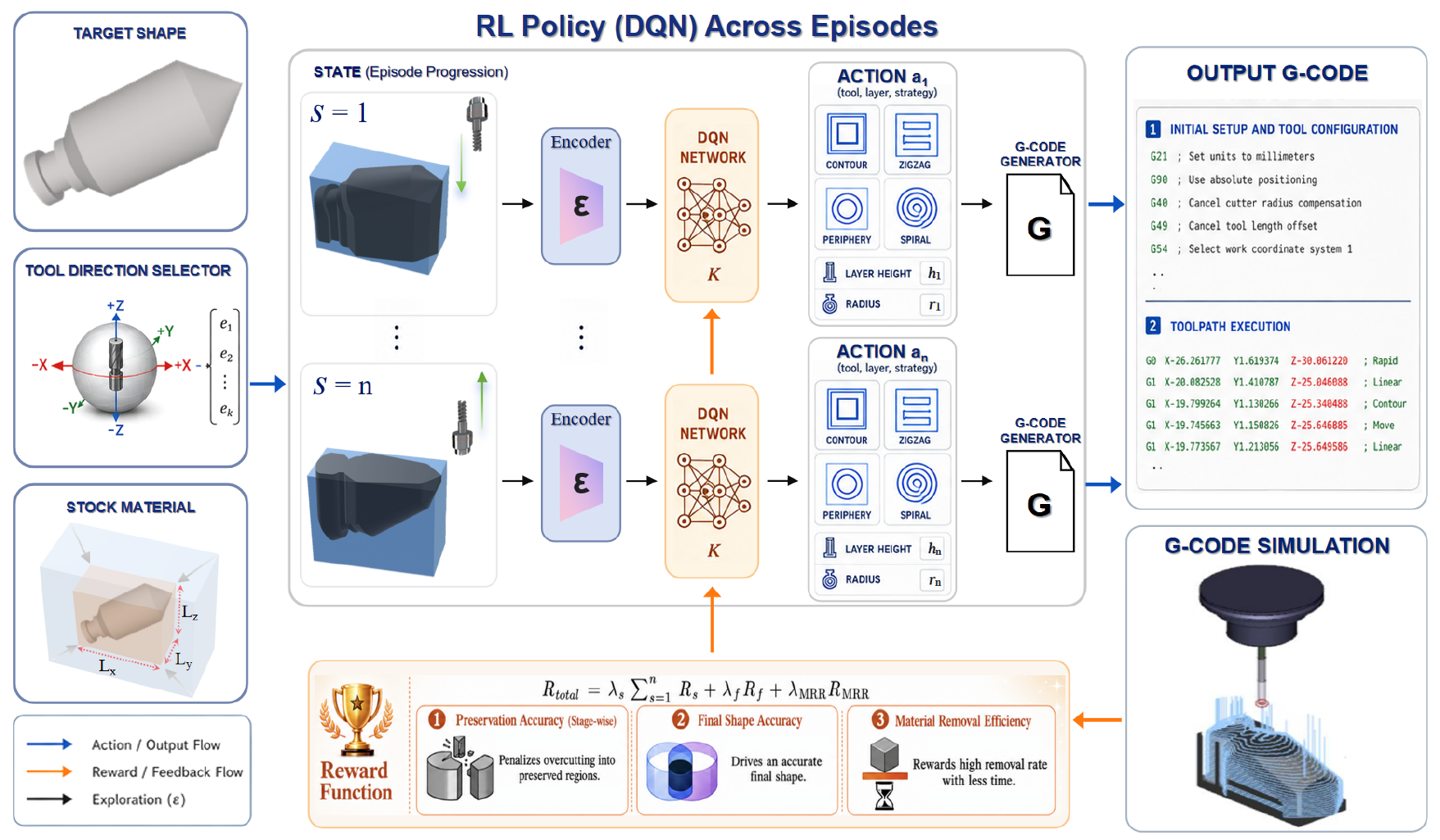}
    \par\vspace{1mm}{\small\textbf{Mandatory pass:} Contour \qquad \textbf{DQN-selectable clearing:} Zigzag, Periphery, or Spiral}
    \caption{
        \textbf{The framework of Geo2Gcode.} The orientation module first produces an ordered set of views. At each view, the contour pass is mandatory, while the DQN selects one of three clearing strategies together with tool radius and layer height. The resulting G-code is simulated to compute a reward balancing reconstruction fidelity and simulated cutting time.
    }
    \label{fig:framework}
\vspace{-3mm}
\end{figure*}

\section{Method}
\label{sec:method}

As shown in Figure~\ref{fig:framework}, Geo2Gcode represents a structured machining primitive as a tuple $(v_s,\mathcal{S}_s,a_s)$ of a direction, its closed target reconstruction, and radius/height/traversal parameters. Given this tuple, the path generator deterministically emits layerwise G-code, whose execution updates the residual stock carried into the next view.

For parameter selection, we use a Deep Q-Network (DQN)~\cite{DQN} in a finite-horizon RL task with one decision per selected orientation. The representation and compiler define what programs can be generated, and the DQN selects a program within that family rather than generating individual tool motions, which keeps every decision interpretable.
The pipeline begins with a 3D shape representation $\mathcal{S}$ of the target object.
An angle selection module first determines machining orientations \( V = \{v_s\}_{s=1}^n \) that increase sampled surface coverage. Each selected angle provides a different indexed view of the object.
For each view direction $v_s$, a visibility-based reconstruction step forms $\mathcal{S}_s$ by retaining visible target regions and filling occluded regions. This supplies a closed per-view target for toolpath construction, but it is not by itself a certificate of finite-cutter accessibility.
Once the shape $\mathcal{S}_s$ is reconstructed for a given machining angle $v_s$, an encoder $\mathcal{E}$ extracts a compact latent representation of the geometry:
\begin{equation}
\mathbf{z}_s = \mathcal{E}(\mathcal{S}_s),
\end{equation}
where $\mathbf{z}_s\in\mathbb{R}^d$, the DGCNN feature of the reconstructed view $\mathcal{S}_s$, is the entire policy observation, and neither the view index nor the evolving stock $V_{s-1}$ is included. The stock is maintained by the simulator and enters learning through transitions and rewards.
From $\mathbf{z}_s$, the DQN selects tool radius, layer height, and clearing strategy as one discrete action:
\begin{equation}
a_s = \{r_s^{\text{tool}}, h_s^{\text{layer}}, s_s^{\text{strategy}}\} \in \mathcal{A}.
\end{equation}
The generated parameters are then translated into executable G-code as machine-readable instructions through a G-code generator.
At the end of the episode, the complete G-code sequence, corresponding to all selected machining angles, is then sent to a CNC machine simulator for validation.
The simulator propagates action-dependent residual stock between views to compute intermediate and terminal rewards. Because the observation is the per-view target rather than the stock, the selector is a target-conditioned contextual policy that cannot react to unobserved stock states. We still train it with bootstrapped multi-view returns rather than a per-view bandit objective, so each view's action is credited with its effect on the rewards of later views.
The DQN policy is updated using this feedback, refining its decision-making for future episodes.

\subsection{Tool Orientation Selection}
The goal of the tool orientation selection module is to identify a small set of orientations with high sampled surface coverage.
Since a shape may be occluded from a single direction, we select orientations $V = \{v_s\}_{s=1}^n$ that increase visibility while keeping the number of indexed setups small.
A surface point $p$ is considered visible from a candidate orientation $v$ when its ray test is unobstructed. Visibility is a geometric proxy rather than a complete cutter-accessibility test, in the spirit of visibility maps~\cite{Woo1994Visibility} and complementary to learned accessibility prediction~\cite{zhong2025deepmill}.
Mathematically, visibility is determined by casting a ray from $p$ along $-v$ and checking for intersections with the object surface. If the ray hits another surface point before exiting, $p$ is occluded in direction $v$.
To select tool directions, we use an iterative greedy approach. At each step $i$, the next orientation $v_i$ is chosen to maximize the number of uncovered visible points:
\begin{equation}
v_i = \arg\max_{v} \sum_{p \in P_{\text{uncovered}}} \mathbf{1} (\text{Vis}(p, v)),
\end{equation}
where $P_{\text{uncovered}}$  is the set of surface points not yet covered, and $\mathbf{1}(\text{Vis}(p, v))$ is $1$ if $p$ is visible from $v$, and $0$ otherwise.
The reported configuration evaluates 100 Fibonacci-sphere candidate directions and stops at 90\% sampled coverage or five selected views, yielding $V = \{v_1, \ldots, v_n\}$.

\subsection{Visibility-Based Reconstruction}
\begin{figure}[t]
\centering
    \includegraphics[width=0.9\linewidth]{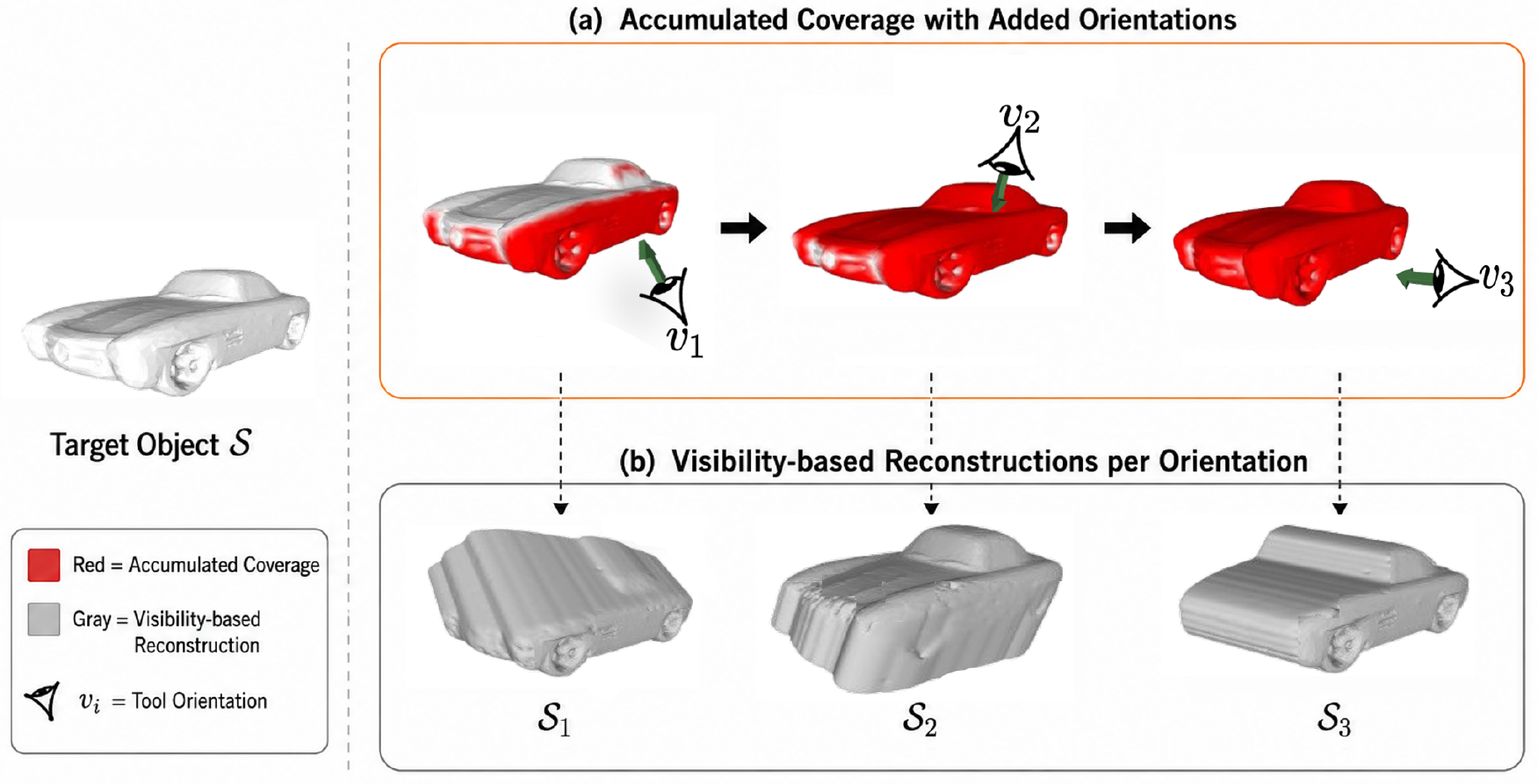}
    \caption{\textbf{Tool orientation selection process and visibility-based reconstructions.}}
    \label{fig:orientation_selection}
\vspace{-3mm}
\end{figure}

The visibility-based reconstruction processes $\mathcal{S}$ into a closed per-view target $\mathcal{S}_s$ in which visible regions align with $\mathcal{S}$ and occluded regions along $v_s$ are filled, as shown in Figure~\ref{fig:orientation_selection}.
This representation conditions path construction and stock simulation, while downstream machine-level feasibility remains outside the geometric visibility test.
To achieve this, the reconstruction process employs a surface voxelization approach, converting the visible portions of $\mathcal{S}$ into a discrete volumetric grid.
The occluded regions are filled to form a watertight representation.
The reconstructed shape is refined with marching cubes~\cite{LorensenCline1987MarchingCubes} and Laplacian smoothing to yield a smooth, continuous mesh.

\subsection{G-code Generation}
\label{sec:gcodegen}
The G-code generator converts selected parameters into CNC commands, identically for learned, fixed, and random selection.
Machining proceeds layerwise with depth increment $h_s^{\text{layer}}$.
Contour traces the object boundary before clearing surrounding stock.
Zigzag, Periphery, or Spiral removes material still connected to deeper layers.
The selected strategy applies to every layer of one view.
Disconnected path segments are joined by retract-and-reposition moves, and the simulator checks the cutting tool rather than the full holder, fixture, or machine envelope.
In each layer, the tool retracts to a starting position $(x_{s,\text{st}},y_{s,\text{st}},z_{s,\text{st}})$ executing:
\begin{equation}
G_{s,\text{st}} = \texttt{G}0 \; X x_{s,\text{st}} \; Y y_{s,\text{st}} \; Z z_{s,\text{st}},
\end{equation}
where $\texttt{G}0$ refers to the transition to the starting position.
This prevents rapid transitions from being treated as cutting moves in the stated simulator.
At each layer $l$, the contour toolpath first traces the target surface by selecting the points whose signed distance to $\mathcal{S}_s$ matches the tool radius, $P_{s,l}^{\text{surf}} = \{ p \in P_l; \space |\text{SDF}_s(p) - r_s^{\text{tool}}| < \epsilon \}$ with a small tolerance $\epsilon$, and visits them in nearest-neighbor order, which detaches the object geometry from the surrounding material. The remaining stock, $P_{s,l}^{\text{out}} = \{ p \in P_l; \space \text{SDF}_s(p) > r_s^{\text{tool}} \}$, is then cleared by one of three strategies that differ in the order in which these points are visited. Since consecutive points are joined by cutting G1 moves, the order also determines the connecting sweeps, so the strategies remove different material. Zigzag sweeps back and forth across the layer, periphery orders them counterclockwise around the starting point, and spiral follows a continuous outward trajectory. In all cases the ordered points are emitted as carving moves
\begin{equation}
G_{s,l,i} = \texttt{G}1 \; Xx_{s,l,i}\; Yy_{s,l,i}\; Zz_{s,l,i},
\label{eq:gcode-g1}
\end{equation}
where $(x_{s,l,i},y_{s,l,i},z_{s,l,i})$ locates $p_{s,l,i}$ and $\texttt{G}1$ denotes cutting. Radius sets candidate positions, height sets depth increments, and strategy sets traversal order. Section~\ref{sec:results} and the generator ablation discuss their geometric effects.

\begin{table*}[t]
\centering
\small
\setlength{\tabcolsep}{5pt}
\begin{tabular}{l c cccc cccc}
    \toprule
    \multirow{2}{*}{\textbf{Method}} &
    \multirow{2}{*}{\textbf{Tuning}} &
    \multicolumn{4}{c}{\textbf{ABC}} & \multicolumn{4}{c}{\textbf{ShapeNet}} \\
    \cmidrule(lr){3-6} \cmidrule(lr){7-10}
    && \textbf{IoU}$\uparrow$ & \textbf{F1}$\uparrow$ & \textbf{CD}$\downarrow$ & \textbf{NC}$\uparrow$
     & \textbf{IoU}$\uparrow$ & \textbf{F1}$\uparrow$ & \textbf{CD}$\downarrow$ & \textbf{NC}$\uparrow$ \\
    \midrule
    CSG-Stump~\cite{CSG-Stump} & {\cmark} & 0.787 & 0.879 & 0.428 & 0.884 & 0.697 & 0.827 & 0.521 & 0.866 \\
    ExtrudeNet~\cite{ExtrudeNet} & {\cmark} & 0.769 & 0.875 & 0.505 & 0.871 & 0.607 & 0.773 & 0.918 & 0.844 \\
    CAPRI-Net~\cite{CAPRI-Net} & {\cmark} & 0.768 & 0.866 & \textbf{0.312} & 0.914 & 0.700 & 0.824 & \textbf{0.447} & \textbf{0.895} \\
    SECAD-Net~\cite{SECAD-Net} & {\cmark} & 0.776 & 0.867 & 0.398 & 0.900 & 0.650 & 0.784 & 2.405 & 0.852 \\
    CNC-Net~\cite{CNC-Net} & {\xmark} & 0.780 & 0.889 & 1.127 & 0.864 & 0.698 & 0.817 & 2.154 & 0.844 \\
    CNC-Net~\cite{CNC-Net} & {\cmark} & 0.824 & 0.901 & 0.509 & 0.893 & 0.740 & 0.850 & 1.562 & 0.863 \\
    \midrule
    \oursrow Ours & {\xmark} & 0.837 & 0.907 & 0.493 & 0.914 & 0.766 & 0.858 & 0.678 & 0.879 \\
    \oursrow Ours & {\cmark} & \textbf{0.848} & \textbf{0.912} & 0.460 & \textbf{0.922} & \textbf{0.773} & \textbf{0.863} & 0.669 & 0.880 \\
    \bottomrule
\end{tabular}
\caption{\textbf{Geometric fidelity on ABC~\cite{Abc} (50 test shapes) and ShapeNet~\cite{Shapenet} (130 test shapes).} Tuning: per-shape adaptation (five episodes for ours). Baselines output CAD primitives (top four) or CNC operations (CNC-Net), whereas ours outputs executable G-code. CD is scaled by $10^3$.}
\label{tab:quant}
\vspace{-3mm}
\end{table*}

\subsection{Reward Calculation}
Let $V_t$ be target occupancy and $V_s$ residual stock after view $s$, initialized to a full stock $V_0$. Each G1 move sweeps a sampled flat disk of radius $r$ (the cutter end face) together with the tool axis over one layer height, and sets the swept voxels to zero. The preservation, terminal fidelity, and removal-efficiency rewards are
\begin{equation}
\begin{aligned}
\mathcal R_s &= |V_s\cap V_t|/|V_t|,\\
\mathcal R_f &= \operatorname{IoU}(V_n,V_t),\\
\mathcal R_{\mathrm{MRR}} &= |V_0\setminus V_n|/t_{\mathrm{cut}}.
\end{aligned}
\end{equation}
Here $t_{\mathrm{cut}}$ is the simulated cutting time $T$, defined as the total length of the emitted toolpath over all views, in normalized unit-cube coordinates, divided by a constant feed of 100 normalized units per time unit. All moves, including rapids and retracts, are charged at this feed, so $T$ is a path-length proxy, not calibrated machine seconds. The total reward is
\begin{equation}
\mathcal R_{\mathrm{total}}=\lambda_s\sum_{s=1}^n\mathcal R_s+\lambda_f\mathcal R_f
+\lambda_{\mathrm{MRR}}\mathcal R_{\mathrm{MRR}}.
\label{eq:reward}
\end{equation}
The preservation term penalizes loss of target material, terminal IoU balances missing and excess stock, and the efficiency term favors less simulated cutting time. Removing $\mathcal{R}_{\mathrm{MRR}}$ shows its effect on the accuracy--time trade-off (Sec.~\ref{sec:ablation}).

\subsection{Implementation Details}
\label{sec:implementation}
We train for 2,000 episodes, sampling one object with replacement and taking one action per orientation. The stock is the target's bounding box plus a small margin, voxelized at $256^3$ (about 0.4\,mm per 100\,mm) so voxel size adapts to each shape, and the tool starts 0.1 above the shape. The action space uses radii $\{0.005,0.01,0.02,0.03,0.04,0.05,0.1\}$ and layer heights $\{1/64,1/128\}$ in normalized units, and each view applies a mandatory Contour pass followed by Zigzag, Spiral, or Periphery clearing. When emitting G-code, normalized coordinates are rescaled to the physical job size. DGCNN~\cite{dgcnn} maps 16,384 points to 1024 features, and a DQN~\cite{DQN} with two 128-unit fully connected layers selects actions. We use Adam with $\text{lr}=10^{-3}$, $\gamma=0.95$, a target network, experience replay, $\epsilon$ decaying from 1.0 to 0.01, and $\lambda_s=\lambda_f=1$, $\lambda_{\text{MRR}}=0.1$. Ablations are retrained without the named reward term or generator, and every ablation and orientation row uses the same five-episode adaptation as the full model. Separate policies are trained on ABC and ShapeNet. On a single Quadro RTX 8000, pretraining takes about 30\,h and five-episode adaptation takes about 10\,min per shape. The generated G-code is additionally inspected in the CIMCO G-code simulator~\cite{cimco}.

\section{Experiments}
\label{sec:experiments}

\begin{figure*}[t]
    \centering

    \begin{subfigure}[t]{\textwidth}
        \centering

        \begin{subfigure}[b]{0.10\textwidth}
            \includegraphics[clip,width=\textwidth, page=1]{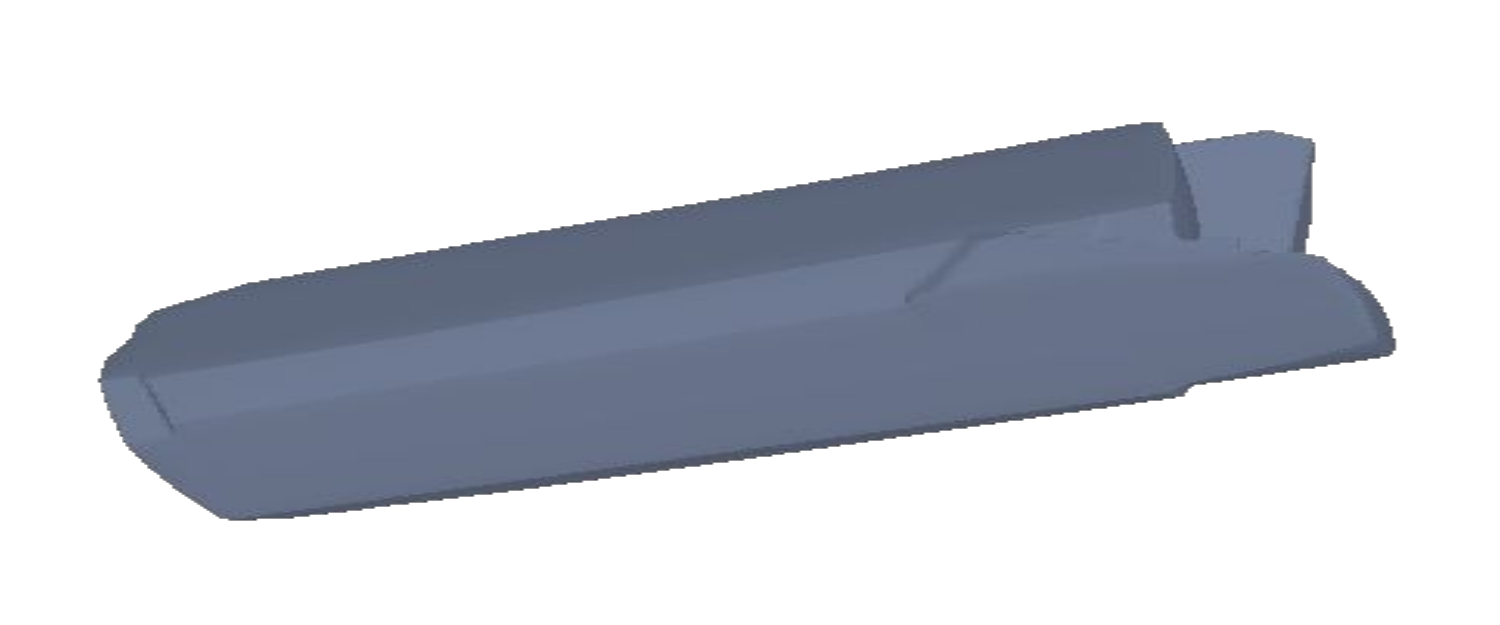}
        \end{subfigure}
        \hfill
        \begin{subfigure}[b]{0.10\textwidth}
            \includegraphics[clip,width=\textwidth, page=2]{figure_assets/Geo2Gcode_abc_qual1_final.pdf}
        \end{subfigure}
        \hfill
        \begin{subfigure}[b]{0.10\textwidth}
            \includegraphics[clip,width=\textwidth, page=3]{figure_assets/Geo2Gcode_abc_qual1_final.pdf}
        \end{subfigure}
        \hfill
        \begin{subfigure}[b]{0.10\textwidth}
            \includegraphics[clip,width=\textwidth, page=4]{figure_assets/Geo2Gcode_abc_qual1_final.pdf}
        \end{subfigure}
        \hfill
        \begin{subfigure}[b]{0.10\textwidth}
            \includegraphics[clip,width=\textwidth, page=5]{figure_assets/Geo2Gcode_abc_qual1_final.pdf}
        \end{subfigure}
        \hfill
        \begin{subfigure}[b]{0.10\textwidth}
            \includegraphics[clip,width=\textwidth, page=6]{figure_assets/Geo2Gcode_abc_qual1_final.pdf}
        \end{subfigure}
        \hfill
        \begin{subfigure}[b]{0.10\textwidth}
            \includegraphics[clip,width=\textwidth, page=7]{figure_assets/Geo2Gcode_abc_qual1_final.pdf}
        \end{subfigure}

        \begin{subfigure}[b]{0.12\textwidth}
            \includegraphics[trim=100 0 80 0,clip,width=\textwidth, page=1]{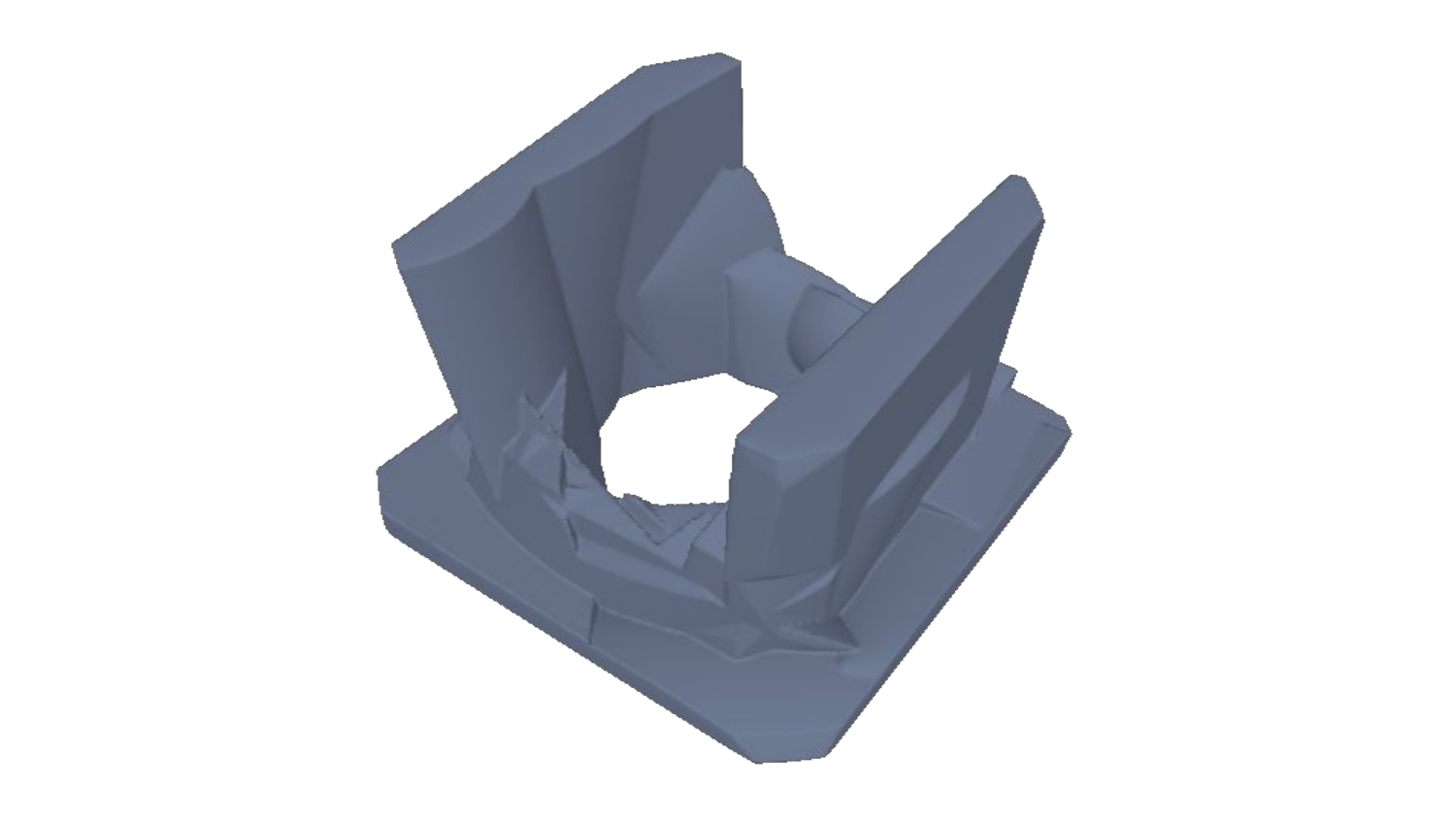}
        \end{subfigure}
        \hfill
        \begin{subfigure}[b]{0.12\textwidth}
            \includegraphics[trim=100 0 80 0,clip,width=\textwidth, page=2]{figure_assets/Geo2Gcode_abc_qual2_final.pdf}
        \end{subfigure}
        \hfill
        \begin{subfigure}[b]{0.12\textwidth}
            \includegraphics[trim=100 0 80 0,clip,width=\textwidth, page=3]{figure_assets/Geo2Gcode_abc_qual2_final.pdf}
        \end{subfigure}
        \hfill
        \begin{subfigure}[b]{0.12\textwidth}
            \includegraphics[trim=100 0 80 0,clip,width=\textwidth, page=4]{figure_assets/Geo2Gcode_abc_qual2_final.pdf}
        \end{subfigure}
        \hfill
        \begin{subfigure}[b]{0.12\textwidth}
            \includegraphics[trim=100 0 80 0,clip,width=\textwidth, page=5]{figure_assets/Geo2Gcode_abc_qual2_final.pdf}
        \end{subfigure}
        \hfill
        \begin{subfigure}[b]{0.12\textwidth}
            \includegraphics[trim=100 0 80 0,clip,width=\textwidth, page=6]{figure_assets/Geo2Gcode_abc_qual2_final.pdf}
        \end{subfigure}
        \hfill
        \begin{subfigure}[b]{0.12\textwidth}
            \includegraphics[trim=100 0 80 0,clip,width=\textwidth, page=7]{figure_assets/Geo2Gcode_abc_qual2_final.pdf}
        \end{subfigure}

        \begin{subfigure}[b]{0.12\textwidth}
            \includegraphics[trim=100 0 80 0,clip,width=\textwidth, page=1]{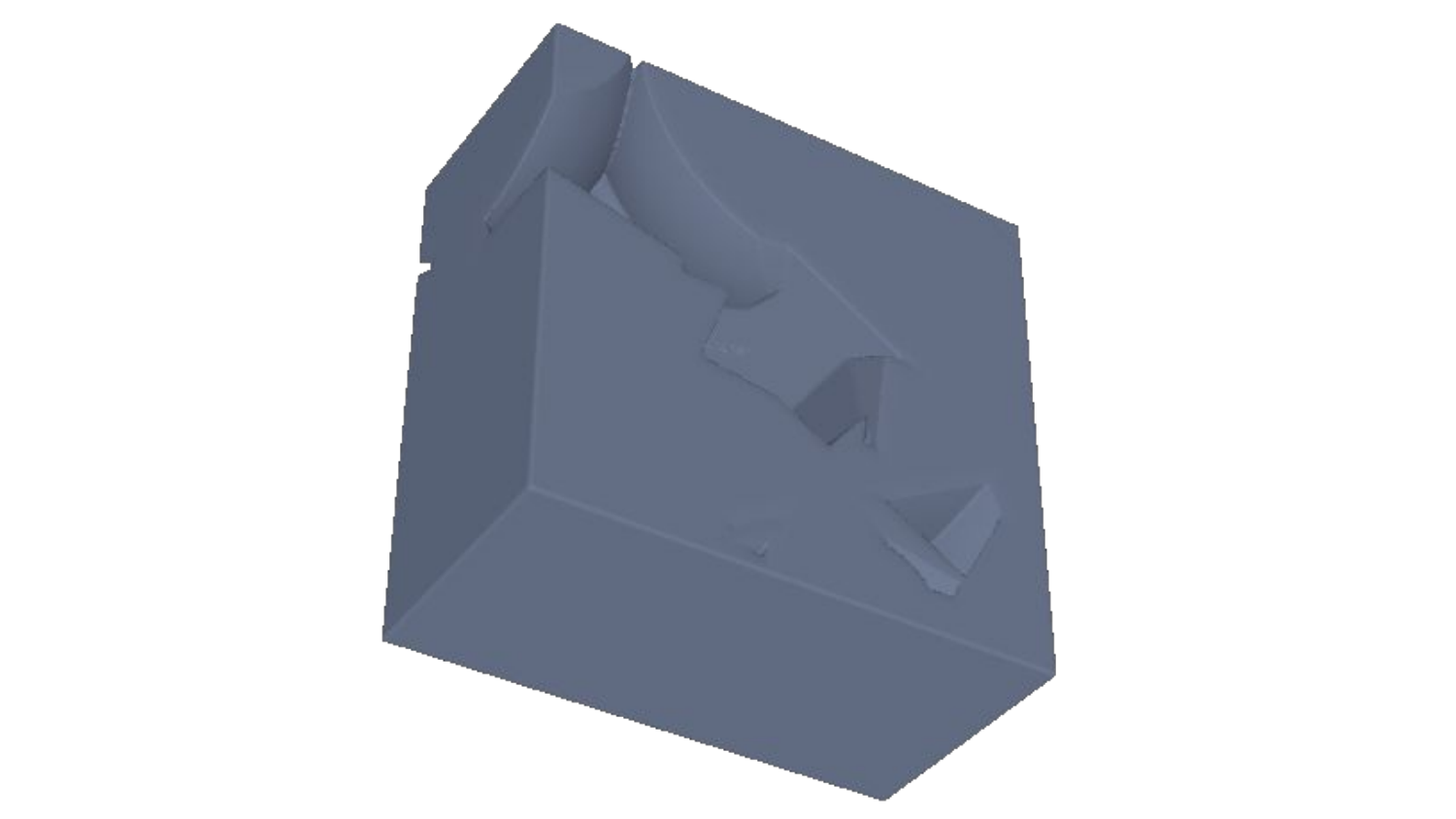}
        \end{subfigure}
        \hfill
        \begin{subfigure}[b]{0.12\textwidth}
            \includegraphics[trim=100 0 80 0,clip,width=\textwidth, page=2]{figure_assets/Geo2Gcode_abc_qual3_final.pdf}
        \end{subfigure}
        \hfill
        \begin{subfigure}[b]{0.12\textwidth}
            \includegraphics[trim=100 0 80 0,clip,width=\textwidth, page=3]{figure_assets/Geo2Gcode_abc_qual3_final.pdf}
        \end{subfigure}
        \hfill
        \begin{subfigure}[b]{0.12\textwidth}
            \includegraphics[trim=100 0 80 0,clip,width=\textwidth, page=4]{figure_assets/Geo2Gcode_abc_qual3_final.pdf}
        \end{subfigure}
        \hfill
        \begin{subfigure}[b]{0.12\textwidth}
            \includegraphics[trim=100 0 80 0,clip,width=\textwidth, page=5]{figure_assets/Geo2Gcode_abc_qual3_final.pdf}
        \end{subfigure}
        \hfill
        \begin{subfigure}[b]{0.12\textwidth}
            \includegraphics[trim=100 0 80 0,clip,width=\textwidth, page=6]{figure_assets/Geo2Gcode_abc_qual3_final.pdf}
        \end{subfigure}
        \hfill
        \begin{subfigure}[b]{0.12\textwidth}
            \includegraphics[trim=100 0 80 0,clip,width=\textwidth, page=8]{figure_assets/Geo2Gcode_abc_qual3_final.pdf}
        \end{subfigure}

        \begin{subfigure}[b]{0.12\textwidth}
            \includegraphics[trim=100 0 80 0,clip,width=\textwidth, page=1]{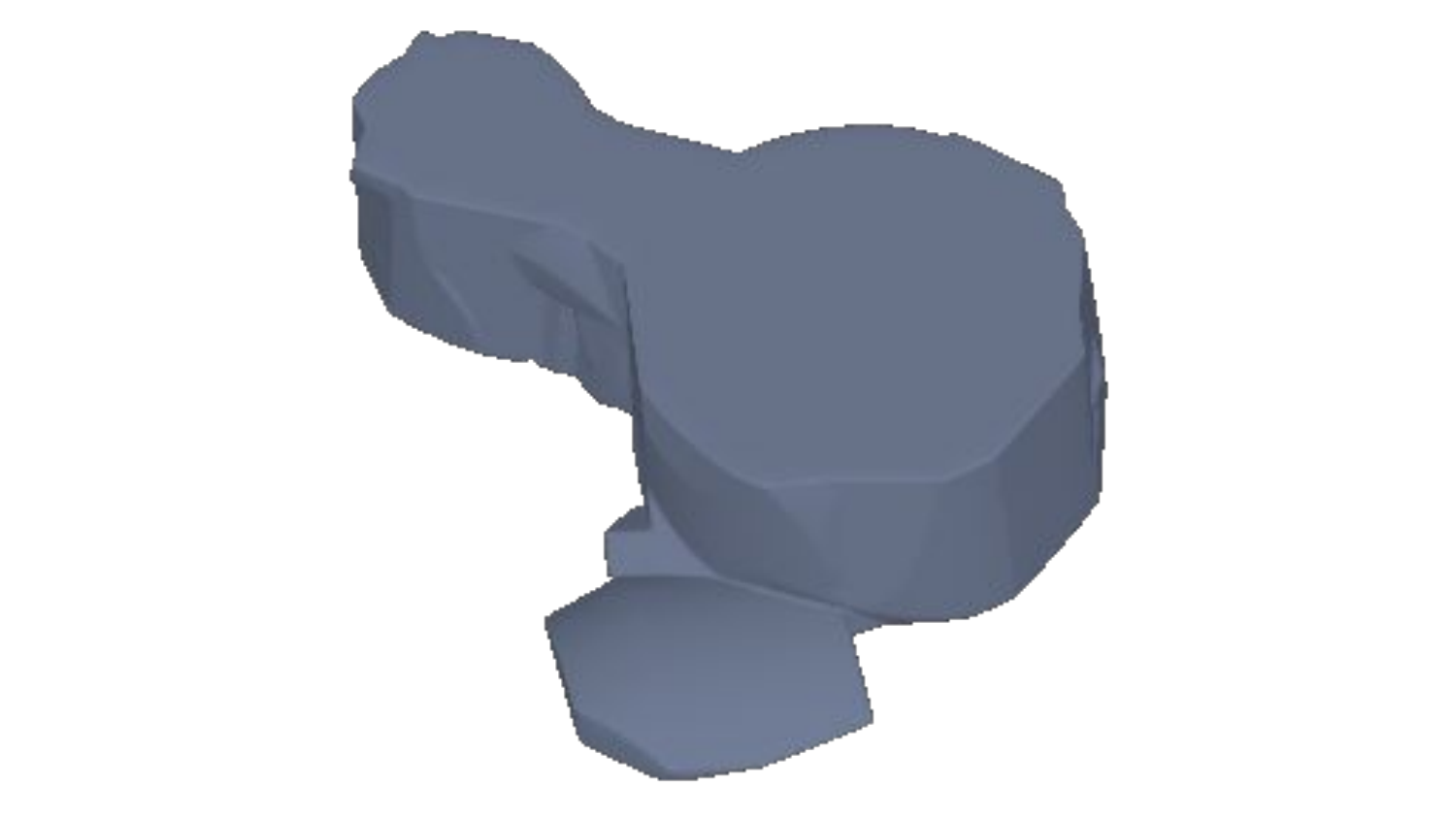}
            \caption*{\footnotesize\mbox{CSG-Stump}}
        \end{subfigure}
        \hfill
        \begin{subfigure}[b]{0.12\textwidth}
            \includegraphics[trim=100 0 80 0,clip,width=\textwidth, page=2]{figure_assets/Geo2Gcode_abc_qual4_final.pdf}
            \caption*{\footnotesize\mbox{ExtrudeNet}}
        \end{subfigure}
        \hfill
        \begin{subfigure}[b]{0.12\textwidth}
            \includegraphics[trim=100 0 80 0,clip,width=\textwidth, page=3]{figure_assets/Geo2Gcode_abc_qual4_final.pdf}
            \caption*{\footnotesize\mbox{CAPRI-Net}}
        \end{subfigure}
        \hfill
        \begin{subfigure}[b]{0.12\textwidth}
            \includegraphics[trim=100 0 80 0,clip,width=\textwidth, page=4]{figure_assets/Geo2Gcode_abc_qual4_final.pdf}
            \caption*{\footnotesize\mbox{SECAD-Net}}
        \end{subfigure}
        \hfill
        \begin{subfigure}[b]{0.12\textwidth}
            \includegraphics[trim=100 0 80 0,clip,width=\textwidth, page=5]{figure_assets/Geo2Gcode_abc_qual4_final.pdf}
            \caption*{\footnotesize\mbox{CNC-Net}}
        \end{subfigure}
        \hfill
        \begin{subfigure}[b]{0.12\textwidth}
            \includegraphics[trim=100 0 80 0,clip,width=\textwidth, page=6]{figure_assets/Geo2Gcode_abc_qual4_final.pdf}
            \caption*{\footnotesize\mbox{Ours}}
        \end{subfigure}
        \hfill
        \begin{subfigure}[b]{0.12\textwidth}
            \includegraphics[trim=100 0 80 0,clip,width=\textwidth, page=8]{figure_assets/Geo2Gcode_abc_qual4_final.pdf}
            \caption*{\footnotesize\mbox{Target (GT)}}
        \end{subfigure}

        \caption{Qualitative results on the ABC~\cite{Abc} dataset.}
        \label{fig:qual_abc}
    \end{subfigure}

    \begin{subfigure}[t]{\textwidth}
        \centering

        \begin{subfigure}[b]{0.10\textwidth}
            \includegraphics[clip,width=\textwidth, page=1]{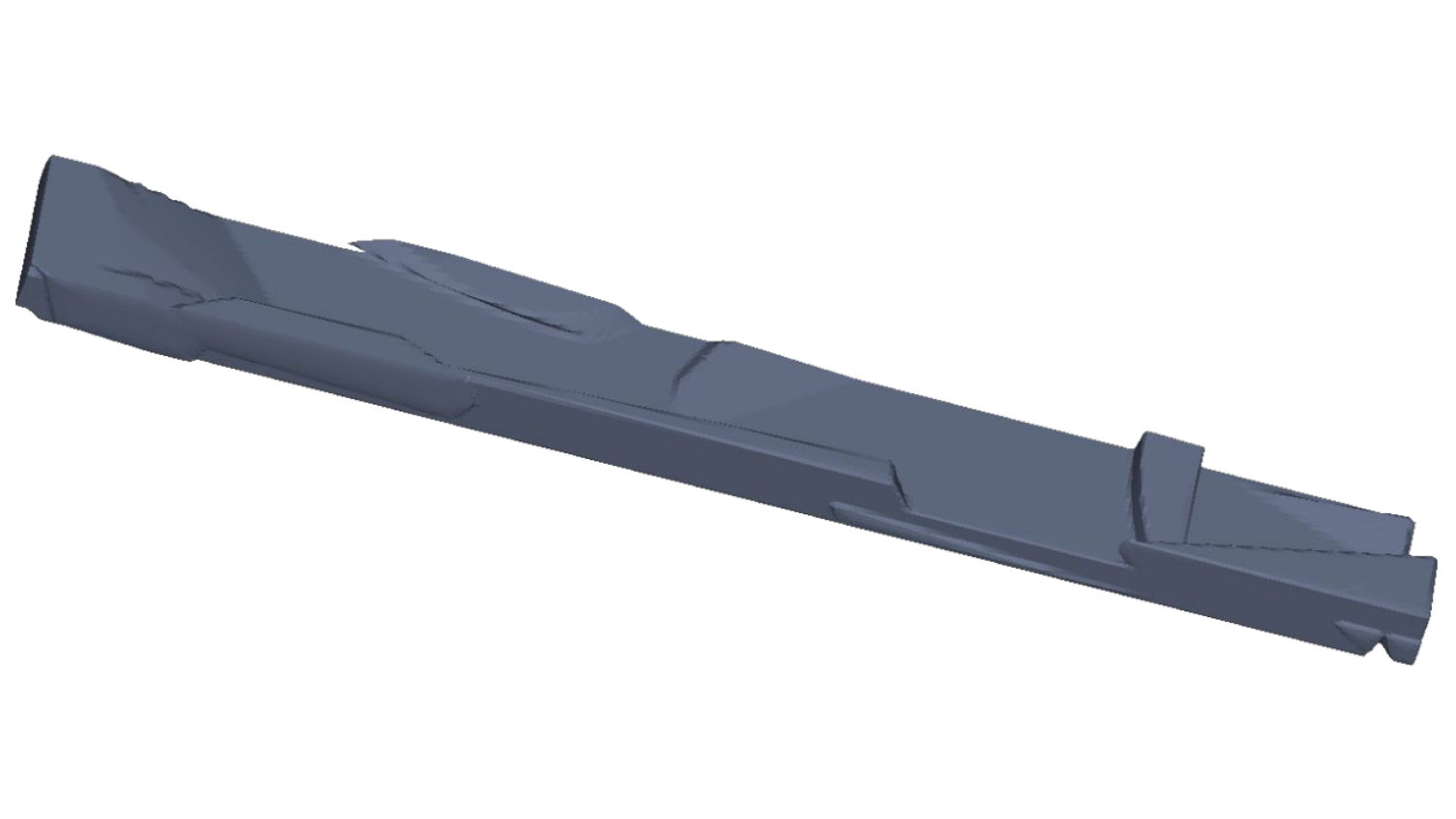}
        \end{subfigure}
        \hfill
        \begin{subfigure}[b]{0.10\textwidth}
            \includegraphics[clip,width=\textwidth, page=2]{figure_assets/Geo2Gcode_shapenet_qual1_final.pdf}
        \end{subfigure}
        \hfill
        \begin{subfigure}[b]{0.10\textwidth}
            \includegraphics[clip,width=\textwidth, page=3]{figure_assets/Geo2Gcode_shapenet_qual1_final.pdf}
        \end{subfigure}
        \hfill
        \begin{subfigure}[b]{0.10\textwidth}
            \includegraphics[clip,width=\textwidth, page=4]{figure_assets/Geo2Gcode_shapenet_qual1_final.pdf}
        \end{subfigure}
        \hfill
        \begin{subfigure}[b]{0.10\textwidth}
            \includegraphics[clip,width=\textwidth, page=5]{figure_assets/Geo2Gcode_shapenet_qual1_final.pdf}
        \end{subfigure}
        \hfill
        \begin{subfigure}[b]{0.10\textwidth}
            \includegraphics[clip,width=\textwidth, page=6]{figure_assets/Geo2Gcode_shapenet_qual1_final.pdf}
        \end{subfigure}
        \hfill
        \begin{subfigure}[b]{0.10\textwidth}
            \includegraphics[clip,width=\textwidth, page=7]{figure_assets/Geo2Gcode_shapenet_qual1_final.pdf}
        \end{subfigure}

        \begin{subfigure}[b]{0.10\textwidth}
            \includegraphics[width=\textwidth, page=1]{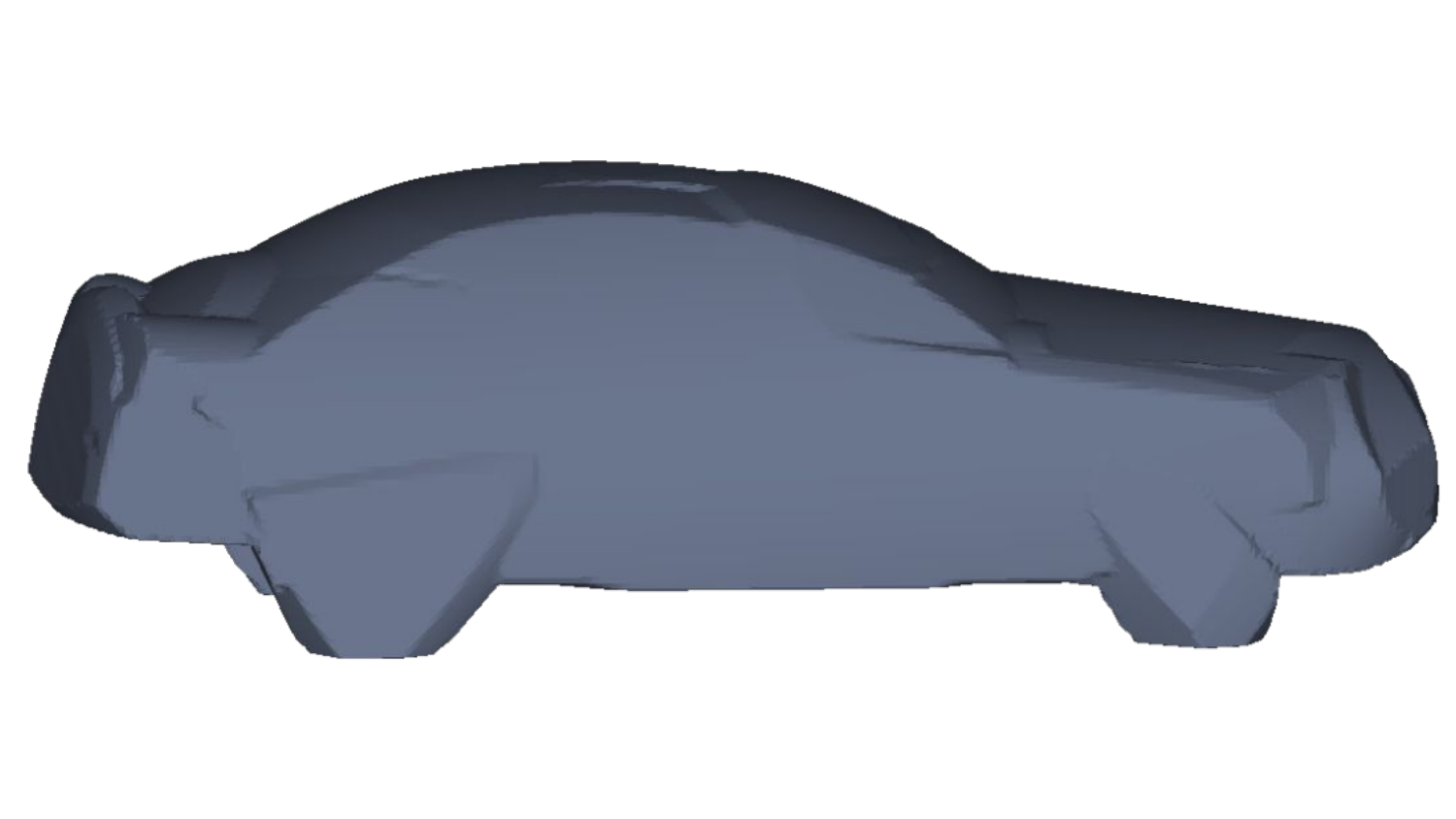}
        \end{subfigure}
        \hfill
        \begin{subfigure}[b]{0.10\textwidth}
            \includegraphics[width=\textwidth, page=2]{figure_assets/Geo2Gcode_shapenet_qual2_final.pdf}
        \end{subfigure}
        \hfill
        \begin{subfigure}[b]{0.10\textwidth}
            \includegraphics[width=\textwidth, page=3]{figure_assets/Geo2Gcode_shapenet_qual2_final.pdf}
        \end{subfigure}
        \hfill
        \begin{subfigure}[b]{0.10\textwidth}
            \includegraphics[width=\textwidth, page=4]{figure_assets/Geo2Gcode_shapenet_qual2_final.pdf}
        \end{subfigure}
        \hfill
        \begin{subfigure}[b]{0.10\textwidth}
            \includegraphics[width=\textwidth, page=5]{figure_assets/Geo2Gcode_shapenet_qual2_final.pdf}
        \end{subfigure}
        \hfill
        \begin{subfigure}[b]{0.10\textwidth}
            \includegraphics[width=\textwidth, page=6]{figure_assets/Geo2Gcode_shapenet_qual2_final.pdf}
        \end{subfigure}
        \hfill
        \begin{subfigure}[b]{0.10\textwidth}
            \includegraphics[width=\textwidth, page=7]{figure_assets/Geo2Gcode_shapenet_qual2_final.pdf}
        \end{subfigure}

        \begin{subfigure}[b]{0.12\textwidth}
            \includegraphics[trim=100 0 80 0,clip,width=\textwidth, page=1]{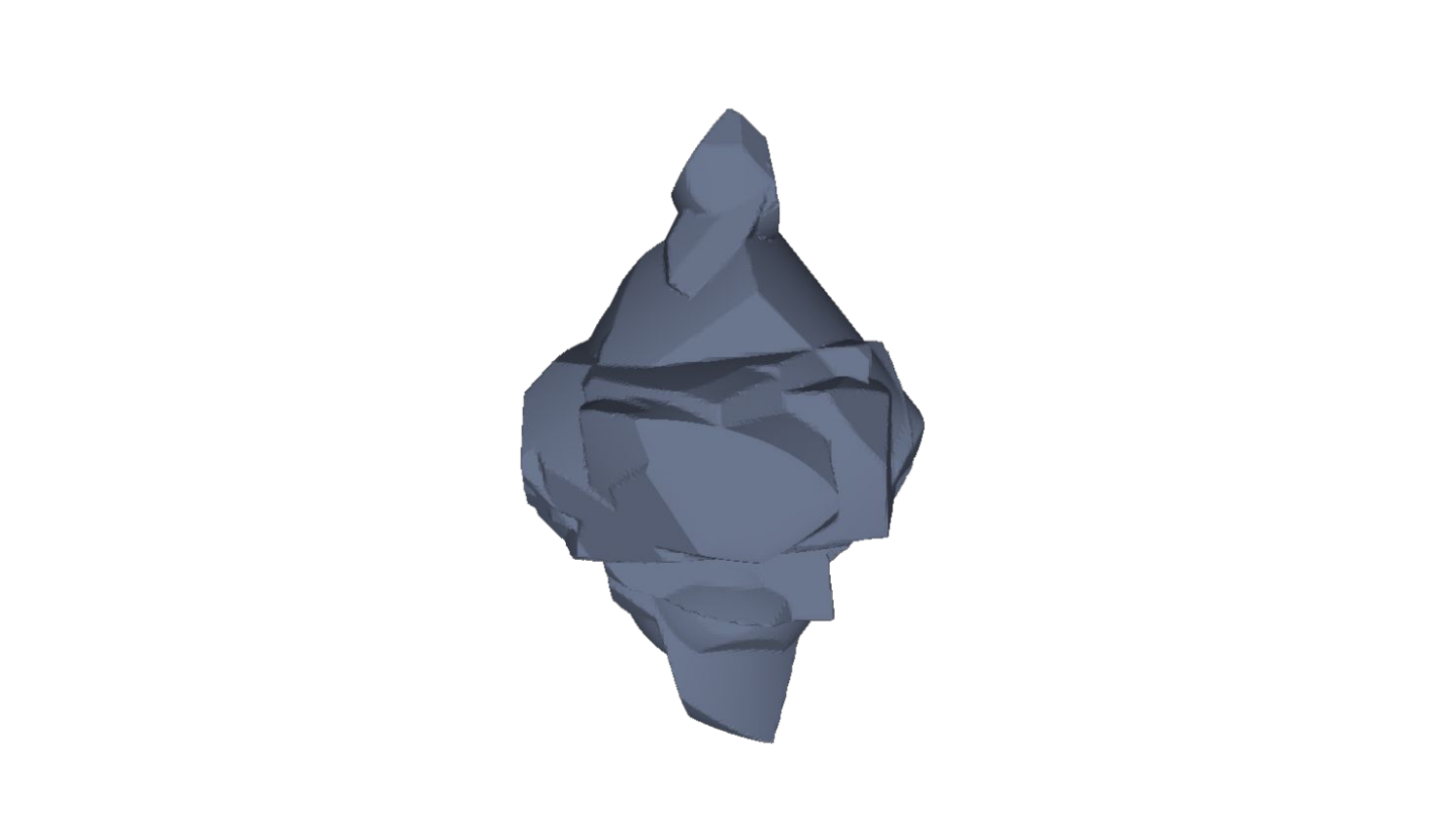}
        \end{subfigure}
        \hfill
        \begin{subfigure}[b]{0.12\textwidth}
            \includegraphics[trim=100 0 80 0,clip,width=\textwidth, page=2]{figure_assets/Geo2Gcode_shapenet_qual4_final.pdf}
        \end{subfigure}
        \hfill
        \begin{subfigure}[b]{0.12\textwidth}
            \includegraphics[trim=100 0 80 0,clip,width=\textwidth, page=3]{figure_assets/Geo2Gcode_shapenet_qual4_final.pdf}
        \end{subfigure}
        \hfill
        \begin{subfigure}[b]{0.12\textwidth}
            \includegraphics[trim=100 0 80 0,clip,width=\textwidth, page=4]{figure_assets/Geo2Gcode_shapenet_qual4_final.pdf}
        \end{subfigure}
        \hfill
        \begin{subfigure}[b]{0.12\textwidth}
            \includegraphics[trim=100 0 80 0,clip,width=\textwidth, page=5]{figure_assets/Geo2Gcode_shapenet_qual4_final.pdf}
        \end{subfigure}
        \hfill
        \begin{subfigure}[b]{0.12\textwidth}
            \includegraphics[trim=100 0 80 0,clip,width=\textwidth, page=6]{figure_assets/Geo2Gcode_shapenet_qual4_final.pdf}
        \end{subfigure}
        \hfill
        \begin{subfigure}[b]{0.12\textwidth}
            \includegraphics[trim=100 0 80 0,clip,width=\textwidth, page=7]{figure_assets/Geo2Gcode_shapenet_qual4_final.pdf}
        \end{subfigure}

        \begin{subfigure}[b]{0.12\textwidth}
            \includegraphics[trim=100 0 80 0,clip,width=\textwidth, page=1]{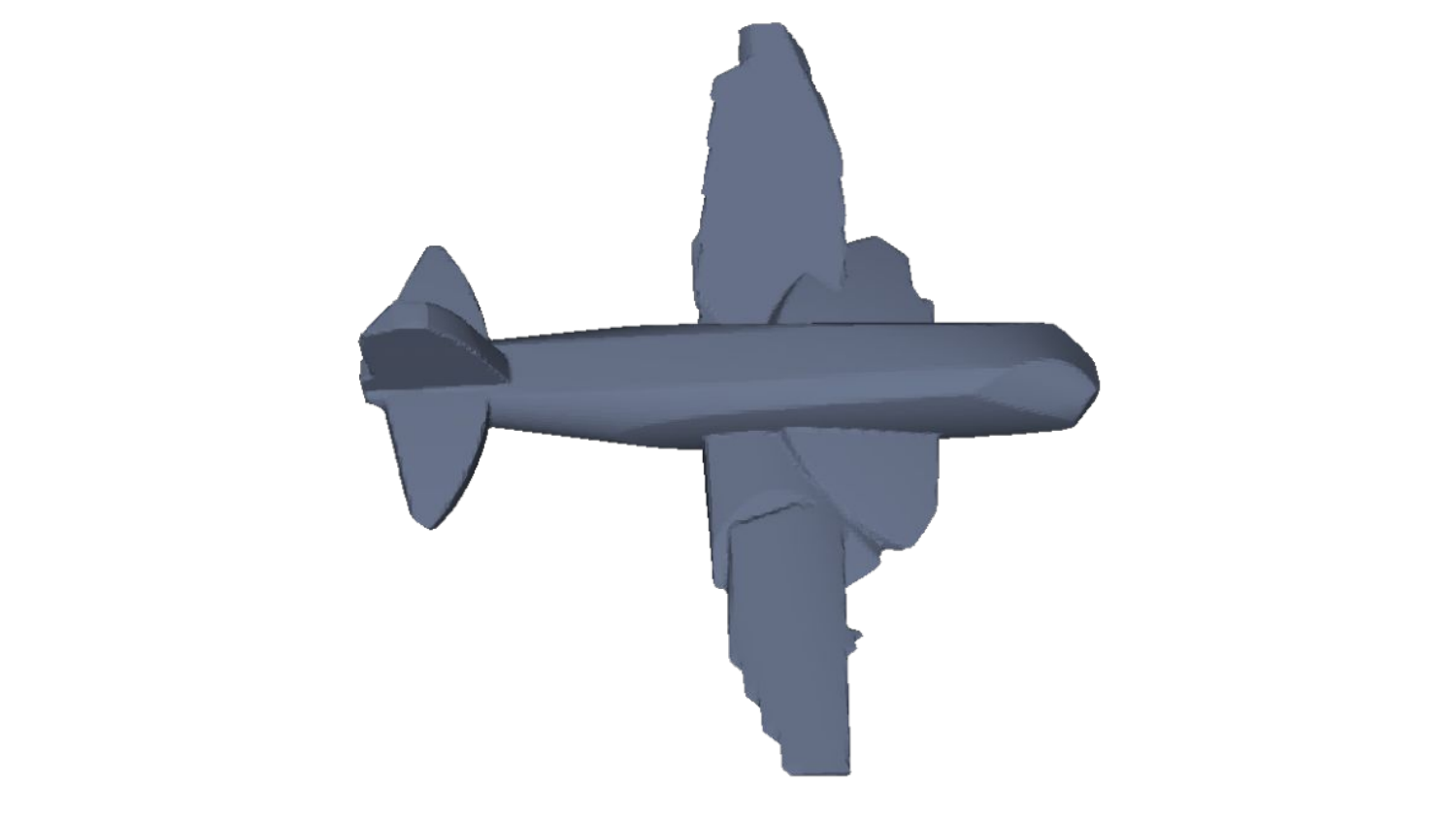}
            \caption*{\footnotesize\mbox{CSG-Stump}}
        \end{subfigure}
        \hfill
        \begin{subfigure}[b]{0.12\textwidth}
            \includegraphics[trim=100 0 80 0,clip,width=\textwidth, page=2]{figure_assets/Geo2Gcode_shapenet_qual6_final.pdf}
            \caption*{\footnotesize\mbox{ExtrudeNet}}
        \end{subfigure}
        \hfill
        \begin{subfigure}[b]{0.12\textwidth}
            \includegraphics[trim=100 0 80 0,clip,width=\textwidth, page=3]{figure_assets/Geo2Gcode_shapenet_qual6_final.pdf}
            \caption*{\footnotesize\mbox{CAPRI-Net}}
        \end{subfigure}
        \hfill
        \begin{subfigure}[b]{0.12\textwidth}
            \includegraphics[trim=100 0 80 0,clip,width=\textwidth, page=4]{figure_assets/Geo2Gcode_shapenet_qual6_final.pdf}
            \caption*{\footnotesize\mbox{SECAD-Net}}
        \end{subfigure}
        \hfill
        \begin{subfigure}[b]{0.12\textwidth}
            \includegraphics[trim=100 0 80 0,clip,width=\textwidth, page=5]{figure_assets/Geo2Gcode_shapenet_qual6_final.pdf}
            \caption*{\footnotesize\mbox{CNC-Net}}
        \end{subfigure}
        \hfill
        \begin{subfigure}[b]{0.12\textwidth}
            \includegraphics[trim=100 0 80 0,clip,width=\textwidth, page=6]{figure_assets/Geo2Gcode_shapenet_qual6_final.pdf}
            \caption*{\footnotesize\mbox{Ours}}
        \end{subfigure}
        \hfill
        \begin{subfigure}[b]{0.12\textwidth}
            \includegraphics[trim=100 0 80 0,clip,width=\textwidth, page=7]{figure_assets/Geo2Gcode_shapenet_qual6_final.pdf}
            \caption*{\footnotesize\mbox{Target (GT)}}
        \end{subfigure}

        \caption{Qualitative results on the ShapeNet~\cite{Shapenet} dataset.}
        \label{fig:qual_shapenet}
    \end{subfigure}

    \caption{\textbf{Qualitative comparisons.} Our outputs are rendered in the CIMCO G-code simulator. Red highlights the simulated machined surface exposed by the cutting operations, and gray denotes stock that remains after simulation.}
    \label{fig:qual_combined}
\vspace{-3mm}
\end{figure*}

\subsection{Datasets}

\noindent\textbf{ABC.} Following CNC-Net~\cite{CNC-Net}, we train on 5,000 normalized single-part ABC~\cite{Abc} CAD objects and, because per-shape tuning is costly, evaluate on 50 random shapes of the 1,000-shape test set.

\noindent\textbf{ShapeNet.} Following CAPRI-Net~\cite{CAPRI-Net}, we train on 35,000 watertight ShapeNet~\cite{Shapenet} shapes over 13 categories and test on 10 shapes per category.

\begin{figure}[t]
    \centering
    \includegraphics[trim=100 0 80 0,clip,width=0.25\columnwidth,page=1]{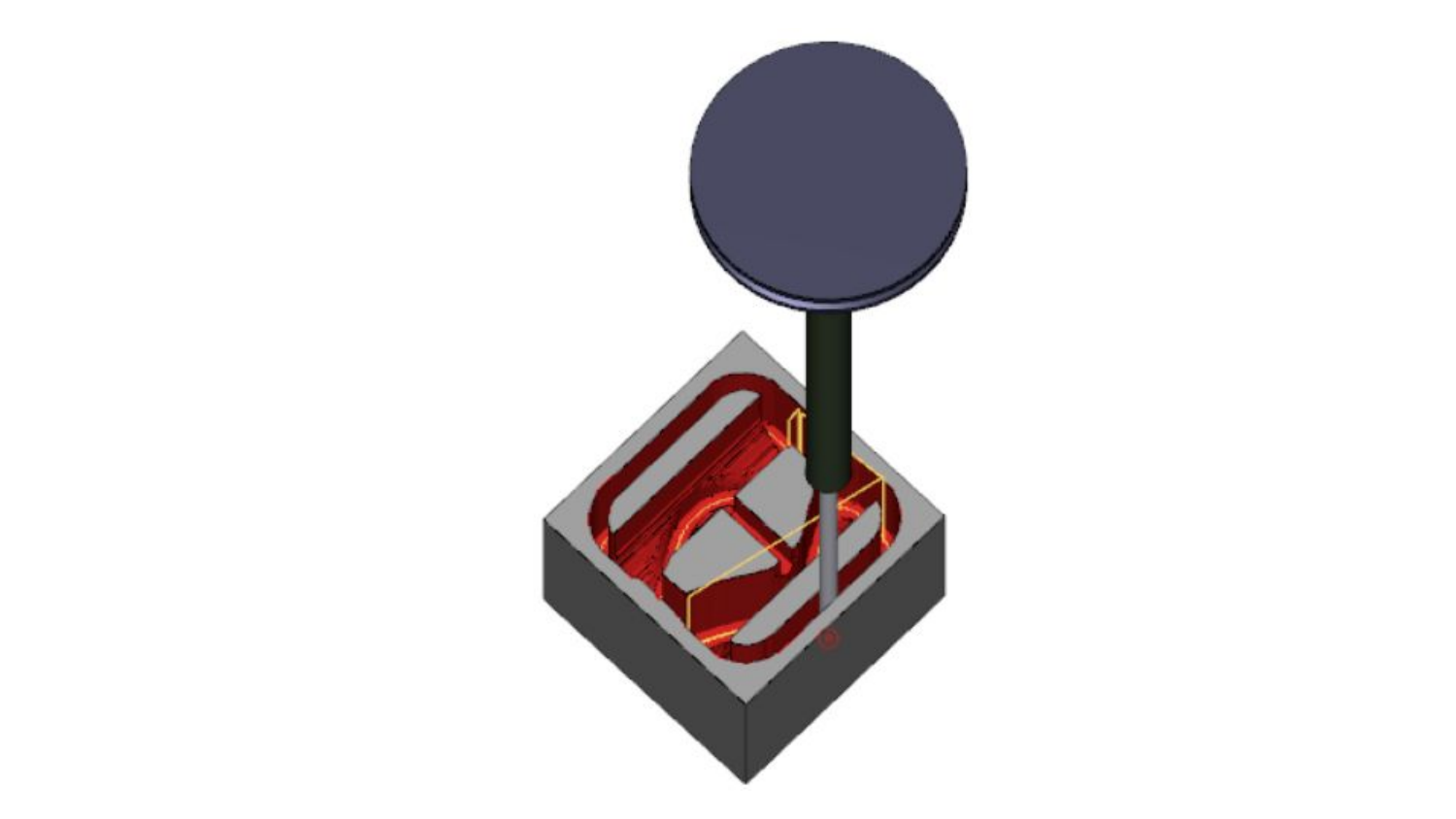}
    \includegraphics[trim=100 0 80 0,clip,width=0.25\columnwidth,page=2]{figure_assets/toolpath_simulation.pdf}
    \includegraphics[trim=100 0 80 0,clip,width=0.25\columnwidth,page=3]{figure_assets/toolpath_simulation.pdf}
    \includegraphics[trim=100 0 80 0,clip,width=0.25\columnwidth,page=4]{figure_assets/toolpath_simulation.pdf}
    \includegraphics[trim=100 0 80 0,clip,width=0.25\columnwidth,page=5]{figure_assets/toolpath_simulation.pdf}
    \includegraphics[trim=100 0 80 0,clip,width=0.25\columnwidth,page=6]{figure_assets/toolpath_simulation.pdf}
    \includegraphics[trim=100 0 80 0,clip,width=0.25\columnwidth,page=7]{figure_assets/toolpath_simulation.pdf}
    \includegraphics[trim=100 0 80 0,clip,width=0.25\columnwidth,page=8]{figure_assets/toolpath_simulation.pdf}
    \includegraphics[trim=100 0 80 0,clip,width=0.25\columnwidth,page=9]{figure_assets/toolpath_simulation.pdf}
    \includegraphics[trim=100 0 80 0,clip,width=0.25\columnwidth,page=10]{figure_assets/toolpath_simulation.pdf}
    \includegraphics[trim=100 0 80 0,clip,width=0.25\columnwidth,page=11]{figure_assets/toolpath_simulation.pdf}
    \includegraphics[trim=100 0 80 0,clip,width=0.25\columnwidth,page=12]{figure_assets/toolpath_simulation.pdf}
    \caption{\textbf{Toolpath simulation} in machining order.}
    \label{fig:toolpath_sim}
\vspace{-3mm}
\end{figure}

\begin{figure}[t]
    \centering
    \includegraphics[trim=140 0 120 0,clip,width=0.24\columnwidth,page=1]{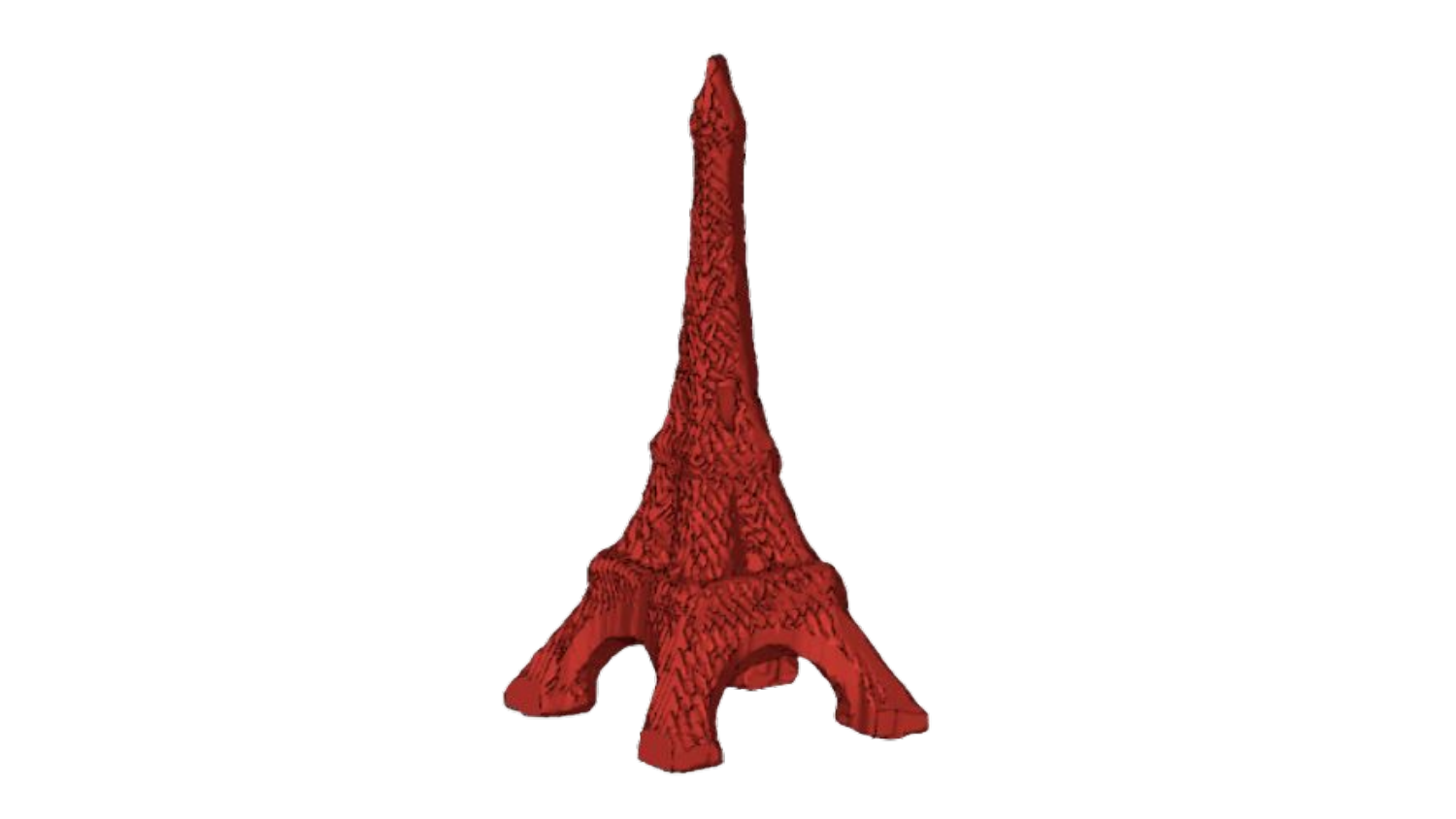}
    \includegraphics[trim=140 0 120 0,clip,width=0.24\columnwidth,page=3]{figure_assets/zero_shot_qual_final.pdf}
    \includegraphics[trim=100 0 80 0,clip,width=0.24\columnwidth,page=4]{figure_assets/zero_shot_qual_final.pdf}
    \includegraphics[trim=100 0 80 0,clip,width=0.24\columnwidth,page=5]{figure_assets/zero_shot_qual_final.pdf}

    \includegraphics[trim=100 0 80 0,clip,width=0.24\columnwidth,page=6]{figure_assets/zero_shot_qual_final.pdf}
    \includegraphics[trim=100 0 80 0,clip,width=0.24\columnwidth,page=7]{figure_assets/zero_shot_qual_final.pdf}
    \includegraphics[trim=100 0 80 0,clip,width=0.24\columnwidth,page=8]{figure_assets/zero_shot_qual_final.pdf}
    \includegraphics[trim=100 0 80 0,clip,width=0.24\columnwidth,page=9]{figure_assets/zero_shot_qual_final.pdf}

    \makebox[0.24\columnwidth][c]{\footnotesize Ours}
    \makebox[0.24\columnwidth][c]{\footnotesize Target}
    \makebox[0.24\columnwidth][c]{\footnotesize Ours}
    \makebox[0.24\columnwidth][c]{\footnotesize Target}
    \caption{\textbf{Zero-shot generalization} to out-of-domain shapes.}
    \label{fig:zero_shot}
\vspace{-3mm}
\end{figure}

\begin{figure}[t]
    \centering
    \includegraphics[width=\columnwidth]{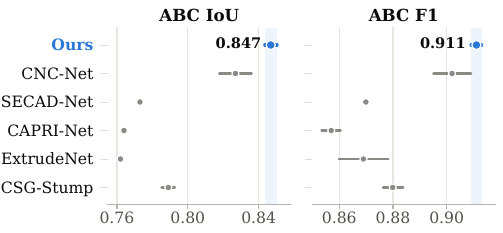}
    \caption{\textbf{Repeated-run 95\% confidence intervals on ABC} (ten random seeds per method, same shapes). Means differ slightly from the single-run values in Table~\ref{tab:quant}.}
    \label{fig:ci}
\vspace{-3mm}
\end{figure}

\subsection{Evaluation Protocol and Metrics}
\label{sec:metrics}
We simulate G-code on a $256^3$ grid over $[-0.5,0.5]^3$, where swept tool volumes remove stock voxels and Marching Cubes converts the residual stock without post-processing. Each method receives the same target object in the input representation its implementation prescribes. Since our method also carves directly against the target, Tables~\ref{tab:quant} and~\ref{tab:large_main} measure how faithfully each output representation reproduces a known target, not a matched-input ablation. Tables~\ref{tab:parameter_controls} and~\ref{tab:CAM} provide controlled comparisons under identical inputs and simulation. Baselines use public implementations or are retrained on the same split, follow their own per-shape tuning protocols, and are meshed and evaluated identically.

\noindent\textbf{Metrics.} Following~\cite{CAPRI-Net, CNC-Net}, we report Intersection over Union (IoU) and F1-score on occupancies over the same $256^3$ grid, and symmetric Chamfer Distance~(CD) and Normal Consistency~(NC) on 8,000 uniformly sampled surface points per object, with CD scaled by 1,000.

\subsection{Quantitative and Qualitative Results}
\label{sec:results}
\noindent\textbf{Settings.} No tuning applies the frozen pretrained policy, while tuning adds five automatic adaptation episodes per target that continue DQN updates on that shape (about ten minutes, excluded from $T$).

\noindent\textbf{Comparison with reconstruction and CNC baselines.} This comparison is context, and the main machining evidence is in Sections~\ref{sec:param_select} and~\ref{sec:cam}. Although CAD reconstruction methods do not emit G-code, we include them because they represent the mesh-to-CAD stage of the conventional mesh$\rightarrow$CAD$\rightarrow$CAM$\rightarrow$G-code route. Geo2Gcode achieves the highest IoU and F1 on both datasets and outperforms CNC-Net, the closest machining-based method, on all four metrics (Table~\ref{tab:quant}). The CNC-Net comparison is the like-for-like one, since CAD methods solve a different task, and the ABC IoU advantage holds across ten random seeds (Figure~\ref{fig:ci}). CAD methods such as CAPRI-Net keep lower CD because smooth analytic surfaces favor surface metrics~\cite{CNC-Net}. Qualitatively (Figures~\ref{fig:qual_abc}, \ref{fig:qual_shapenet}, and~\ref{fig:toolpath_sim}), our CIMCO-simulated~\cite{cimco} outputs retain thin, localized structures that CAD-reconstruction baselines often miss, since primitives carve directly against the target geometry.

\input{main_evidence}

\begin{table}[t]
    \centering
    \small
    \setlength{\tabcolsep}{2.5pt}
    \begin{tabular}{lccccc}
        \toprule
        \textbf{Method} & \textbf{IoU}$\uparrow$ & \textbf{F1}$\uparrow$ & \textbf{CD}$\downarrow$ & \textbf{NC}$\uparrow$ & \textbf{T}$\downarrow$ \\
        \midrule
        MeshCAM (non-expert) & 0.783 & 0.854 & 1.474 & 0.868 & 1488.6 \\
        MeshCAM (expert)     & 0.833 & 0.904 & 0.555 & 0.910 & \textbf{375.7} \\
        \oursrow Ours (no tuning)     & 0.837 & 0.907 & 0.493 & 0.914 & 451.3 \\
        \oursrow Ours (5-ep. tuning)  & \textbf{0.848} & \textbf{0.912} & \textbf{0.460} & \textbf{0.922} & 423.1 \\
        \midrule
        \oursrow Ours, axis-aligned   & 0.834 & 0.905 & 0.501 & 0.920 & 451.6 \\
        \bottomrule
    \end{tabular}
    \caption{\textbf{Comparison with commercial CAM (MeshCAM) on the same 50 ABC meshes.} The expert setting is tuned per part by a machinist, whereas ours involves no manual input. $T$ is the normalized cutting-time proxy. Axis-aligned: ours with each part's expert $\pm x/\pm y/\pm z$ setups (no tuning).}
    \label{tab:CAM}
\vspace{-3mm}
\end{table}

\begin{table}[t]
    \centering
    \small
    \setlength{\tabcolsep}{4pt}
    \begin{tabular}{llccc>{\columncolor{oursrow}}c}
        \toprule
        & \textbf{Metric} & \textbf{6-axes} & \textbf{Rand(5)} & \textbf{Rand(6)} & \textbf{Ours} \\
        \midrule
        \multirow{2}{*}{Coverage} & ABC      & 92.67 & 83.52 & 89.11 & \textbf{96.42} \\
                                  & ShapeNet & 91.27 & 79.77 & 84.73 & \textbf{92.75} \\
        \bottomrule
    \end{tabular}
    \caption{\textbf{Effect of tool orientation selection.} Coverage: \% of sampled surface points visible from the selected views.}
    \label{tab:abl_direction}
\vspace{-3mm}
\end{table}

\begin{table}[t]
    \centering
    \small
    \setlength{\tabcolsep}{3pt}
    \begin{tabular}{lcc>{\columncolor{oursrow}}c}
        \toprule
        \textbf{Proxy} & \textbf{Non-exp.} & \textbf{Expert} & \textbf{Ours} \\
        \midrule
        Gouge $\leq$0.1\,mm (\%)$\uparrow$ & 89.61 & 92.31 & \textbf{98.57} \\
        Mean P95 overcut (mm)$\downarrow$ & 0.249 & 0.262 & \textbf{0.217} \\
        Local dim.\ variation$\downarrow$ & 0.678 & 0.726 & \textbf{0.228} \\
        Junction angle ($^\circ$)$\downarrow$ & 52.4 & 39.9 & \textbf{27.1} \\
        Curvature consistency$\downarrow$ & 1.15 & 7.57 & \textbf{0.69} \\
        Accel.\ violation rate$\downarrow$ & \textbf{1.8e-6} & 4.0e-4 & 1.3e-5 \\
        \bottomrule
    \end{tabular}
    \caption{\textbf{Machining-quality proxies versus MeshCAM} on all 50 ABC parts, scaled to 100\,mm and weighted equally. Overcut is measured from the G-code against target-mesh samples, not the voxel grid. Path proxies assume 100\,mm/min feed and 500\,mm/s$^2$ acceleration.}
    \label{tab:machining}
\vspace{-3mm}
\end{table}

\subsection{Comparison with Commercial CAM}
\label{sec:cam}
Commercial CAM is the practical alternative, so we compare against MeshCAM on the same 50 ABC meshes (Table~\ref{tab:CAM}), given without CAD metadata. Its G-code is replayed in our simulator and timed with the same routine, so all rows share one evaluation pipeline. The expert setup also includes finishing passes, which can only favor it. Because $T$ charges every tool the same feed, it measures toolpath length rather than machining time and favors small tools. The non-expert setting uses one top setup and a 6\,mm flat end mill (our median radius at 100\,mm scale) for parallel roughing and finishing. In the expert setting, one machinist chooses per part 1--5 axis-aligned setups (mean 2.0), parallel roughing with mostly 20\,mm flat end mills (our largest radius), and parallel, waterline, or pencil finishing with 4--8\,mm tools. All MeshCAM tools are flat end mills, as simulated.

Without any manual input, Geo2Gcode matches the expert in IoU (0.837 vs.\ 0.833) with a 20\% longer toolpath, and it clearly outperforms the non-expert configuration in both fidelity and time. It also uses fewer setups (1.8 views per part vs.\ 2.0), and since each view uses one tool, it needs only 1.8 tools per part on average. Real machining time is not evaluated. With adaptation, it gives the best fidelity on every metric. Restricting our views to the expert's axis-aligned setups barely changes the result, so the comparison does not rely on non-axis orientations.

\noindent\textbf{Machining quality.} IoU does not capture overcut or path quality, so Table~\ref{tab:machining} measures them directly from the emitted G-code on the same 50 parts, using an overcut proxy that models each cutter position as a sphere of the tool radius. It approximates flat end mills but is applied identically to all programs (local variation and curvature consistency are standard deviations of neighboring signed errors and of turning angle per unit length). Overcut is evaluated on target-mesh samples, independent of the voxel grid. Our programs overcut the target less than the expert's, with 98.6\% of sampled surface points within 0.1\,mm versus 92.3\%, and the advantage holds on most parts (Wilcoxon $p{<}10^{-4}$). This is not from cutting less, since our IoU, which penalizes leftover stock, is at least the expert's. Our paths are also the smoothest, with the smallest junction angles and most consistent curvature, whereas non-expert MeshCAM has the fewest acceleration violations. Higher fidelity therefore comes neither from gouging the target nor from rougher paths.

\noindent\textbf{Computational cost.} Orientation selection (5.08\,s) and G-code translation (10.1\,s) per part are faster than MeshCAM (18.8--29.3\,s).

\noindent\textbf{Physical machining.} We executed one generated program, scaled to the stock, on a FANUC Robodrill with AL6061 stock and a flat endmill (Figure~\ref{fig:pipeline}). The controller accepted the program without edits to cutting coordinates or path ordering, while an operator handled workholding and clearance. This single run shows controller acceptance, not dimensional accuracy or collision safety.

\begin{table}[t]
    \centering
    \small
    \setlength{\tabcolsep}{3pt}
    \begin{tabular}{lccccc}
        \toprule
        \textbf{Excluded} & \textbf{IoU}$\uparrow$ & \textbf{F1}$\uparrow$ & \textbf{CD}$\downarrow$ & \textbf{NC}$\uparrow$ & \textbf{T}$\downarrow$ \\
        \midrule
        Spiral      & 0.797 & 0.876 & 0.615 & 0.911 & 508.1 \\
        Periphery   & 0.806 & 0.882 & 0.565 & 0.914 & 529.0 \\
        Zigzag      & 0.811 & 0.886 & 0.611 & 0.917 & 441.3 \\
        Contour     & 0.756 & 0.849 & 0.891 & 0.847 & 444.4 \\
        \midrule
        \oursrow None (Ours) & \textbf{0.848} & \textbf{0.912} & \textbf{0.460} & \textbf{0.922} & \textbf{423.1} \\
        \bottomrule
    \end{tabular}
    \caption{\textbf{Effect of toolpath strategies (ABC, $256^3$).} Each row retrains without the named pattern, and all rows use five-episode adaptation.}
    \label{tab:abl_toolpath}
\vspace{-3mm}
\end{table}

\begin{table}[t]
\centering\small
\setlength{\tabcolsep}{3pt}
\begin{tabular}{lcccc}
\toprule
\textbf{Method} & \textbf{IoU}$\uparrow$ & \textbf{F1}$\uparrow$ & \textbf{CD}$\downarrow$ & \textbf{NC}$\uparrow$ \\
\midrule
CNC-Net (per-shape tuned) & 0.630 & 0.765 & 1.234 & 0.780 \\
\oursrow Ours (zero-shot) & \textbf{0.770} & \textbf{0.866} & \textbf{0.710} & \textbf{0.859} \\
\bottomrule
\end{tabular}
\caption{\textbf{ShapeNet-to-OOD transfer on 20 unseen objects.}}
\label{tab:ood_main}
\vspace{-3mm}
\end{table}

\subsection{Zero-Shot Results on Out-of-Domain Data}
We test ShapeNet-trained models on 20 unseen objects, e.g., Bunny, Teapot, and Eiffel Tower (Figure~\ref{fig:zero_shot}, Table~\ref{tab:ood_main}). CNC-Net is fine-tuned per target (about 50\,min), while ours is not. Ours still achieves higher IoU and lower CD, showing that target-compiled primitives transfer to unseen geometry without per-shape optimization.

\begin{table}[t]
    \centering
    \small
    \setlength{\tabcolsep}{3pt}
    \begin{tabular}{lccccc}
        \toprule
        \textbf{Excluded} & \textbf{IoU}$\uparrow$ & \textbf{F1}$\uparrow$ & \textbf{CD}$\downarrow$ & \textbf{NC}$\uparrow$ & \textbf{T}$\downarrow$ \\
        \midrule
        $\mathcal{R}_s$          & 0.792 & 0.871 & 0.603 & 0.903 & \textbf{413.6} \\
        $\mathcal{R}_\text{MRR}$ & \textbf{0.851} & \textbf{0.914} & \textbf{0.456} & 0.921 & 774.8 \\
        \midrule
        \oursrow None (Ours)              & 0.848 & 0.912 & 0.460 & \textbf{0.922} & 423.1 \\
        \bottomrule
    \end{tabular}
    \caption{\textbf{Effect of reward terms (ABC).} Each row retrains without the named term, and all rows use five-episode adaptation.}
    \label{tab:abl_sampling}
\vspace{-3mm}
\end{table}

\subsection{Ablation Studies}
\label{sec:ablation}

\subsubsection{Effect of Tool Orientation Selection.}
\label{subsec:tool_ori}
Table~\ref{tab:abl_direction} compares six principal-axis views, random sets of five or six views, and our greedy selection. With at most five views, our selection covers the most surface points on both datasets, and downstream it matches the setups a machinist chose per part (Table~\ref{tab:CAM}). Orientation selection thus replaces manual setup planning without loss of fidelity.

\subsubsection{Effect of Toolpath Strategies.}
Table~\ref{tab:abl_toolpath} retrains the selector with one pattern removed from the library. Removing Contour causes the largest drop (0.848 to 0.756 IoU), confirming that the surface-following pass carries most of the fidelity. Removing any clearing pattern lowers IoU and increases $T$, so the patterns are complementary. The compiler does not forbid gouging G1 connections, but they are simulated and penalized by $\mathcal{R}_s$.

\subsubsection{Effect of Reward Functions.}
With $\mathcal{R}_f$ retained in Equation~\eqref{eq:reward}, removing the preservation term $\mathcal{R}_s$ clearly lowers fidelity (0.848 to 0.792 IoU), whereas removing $\mathcal{R}_\text{MRR}$ barely changes IoU but increases $T$ by 83\%.

\section{Conclusion}
\label{sec:conclusion}
We presented Geo2Gcode, which turns a 3D shape directly into a CNC program without a CAD model or manual CAM planning. Structured machining primitives keep every program machinist-readable and leave only a few parameter choices, which a simulator-trained selector makes automatically. Experiments against reconstruction baselines and commercial CAM show that this structure enables automation without sacrificing fidelity. Geo2Gcode is comparable to an expert-tuned CAM setup without manual input, its programs run unmodified in an industrial G-code simulator, and a generated program was machined on a physical mill.

\noindent\textbf{Limitations and future work.} Geo2Gcode currently targets roughing of indexed 3+2-axis parts, and its simulator models the cutter but not the holder, fixture, or cutting loads, so collision safety and non-axis fixturing are not yet verified. Selector and ablation results are single runs on 50 ABC parts. Next steps are finishing and 5-axis primitives, repeated runs, and measured parts.

\clearpage
\section*{LLM Usage Statement}
\textit{OpenAI Codex assisted with experiment implementation, figure preparation, and manuscript writing. The authors reviewed all generated text and remain responsible for the submission.}

{
    \small
    \bibliographystyle{ieeenat_fullname}
    \bibliography{main}
}

\clearpage
\appendix
\setcounter{section}{0}\setcounter{table}{0}\setcounter{figure}{0}\setcounter{equation}{0}
\renewcommand{\thesection}{S\arabic{section}}
\renewcommand{\thetable}{S\arabic{table}}
\renewcommand{\thefigure}{S\arabic{figure}}
\renewcommand{\theequation}{S\arabic{equation}}
\section*{Supplementary Material}

\noindent This appendix provides additional details, experiments, and analyses that could not be included due to space constraints. The contents are summarized below:

\begin{itemize}\setlength{\itemsep}{0pt}
    \item \textbf{S1.} Toolpath Strategies
    \item \textbf{S2.} G-code Simulation
    \item \textbf{S3.} Reinforcement Learning Formulation
    \item \textbf{S4.} More Qualitative Results
    \item \textbf{S5.} Effect of Machining Precision
    \item \textbf{S6.} Comparison with the RL-based Method
    \item \textbf{S7.} Comparison with DeepMill on Inaccessibility Prediction
    \item \textbf{S8.} Machining-Relevant Metrics
    \item \textbf{S9.} CAD-Surface Overcut Proxies
    \item \textbf{S10.} Physical CNC Machining Results
    \item \textbf{S11.} Analysis on Complex Shapes
    \item \textbf{S12.} Voxel Resolution and Laplacian Smoothing
    \item \textbf{S13.} Baseline Protocols
    \item \textbf{S14.} Ablations for Tool Orientation Selection
    \item \textbf{S15.} Pre-training Time
    \item \textbf{S16.} Per-Sample Fine-Tuning Time
    \item \textbf{S17.} Compact Selector Diagnostic
\end{itemize}

\section{Toolpath Strategies}
\label{sec:supp:toolpath}

This section gives the full formulation of the four toolpath strategies summarized in Section~\ref{sec:gcodegen}. All four operate within a single machining layer $l$ at view direction $v_s$, and differ in which points they select and in the order in which those points are visited. Lengths are in normalized coordinates, and G-code emission rescales them to the job's physical size.

\paragraph{Shape-dependent pattern suitability.}
Contour follows the surface-offset band, while the three clearing templates traverse the exterior candidate region. Their suitability depends on the geometry of that region and the selected radius and layer height. A pattern that suits one shape need not suit another, so restricting the available templates can degrade reconstruction or increase path length. Table~\ref{tab:abl_toolpath} tests this library restriction with retraining, not a paired wrong-versus-optimal-pattern intervention on each shape. Removing mandatory Contour is also distinct from removing one selectable clearing template. The aggregate results therefore support complementary patterns without identifying a unique optimal pattern for every shape.

\paragraph{Why traversal can affect geometry.}
The simulator removes the tool volume swept along G1 segments, not only at their endpoints. For an ordered waypoint list $\pi$, the removed region has the form $\bigcup_i \mathrm{Sweep}(p_{\pi(i)},p_{\pi(i+1)};r)$. Changing the order changes adjacent pairs and may change that union even when the candidate point set is unchanged. Exterior endpoints alone do not certify clearance of an entire connecting segment. Thus, ``different traversal orders'' does not imply equal geometric outcomes. This explains why an order-based template can affect removal, but does not separately attribute Table~\ref{tab:abl_toolpath}'s aggregate differences to connectors, parameter changes, or retraining.

\subsection{Contour Toolpath}
The contour toolpath is designed to refine the object's surface by closely following its geometry.
The process begins with the visibility-reconstructed shape $\mathcal{S}_s$, where a signed distance function is calculated as $\text{SDF}_s(p)$ for each point $p\in P_l$ in the layer $l$ to its surface.
The contour toolpath points $P_{s,l}^{\text{surf}}$ are determined by selecting points where the SDF value is approximately equal to the tool radius, ensuring that the tool follows the object's surface while maintaining an appropriate clearance:
\begin{equation}
P_{s,l}^{\text{surf}} = \{ p \in P_l; \space \left| \text{SDF}_s(p) - r_s^{\text{tool}} \right| < \epsilon \},
\label{eq:supp:contour}
\end{equation}
where $\epsilon$ is a small tolerance to account for numerical precision.
The sorting process follows a nearest-neighbor approach, starting from a starting point and iteratively selecting the closest unprocessed point until all points are visited and the G-code is updated:
\begin{equation}
G_{s,l,i} = \texttt{G}1 \; Xx_{s,l,i}\; Yy_{s,l,i}\; Zz_{s,l,i},
\label{eq:supp:gcode-g1}
\end{equation}
where $(x_{s,l,i},y_{s,l,i},z_{s,l,i})$ is the location of $p_{s,l,i}\in P_{s,l}^{\text{surf}}$ in $\mathbb{R}^3$ and $\texttt{G}1$ indicates carving motion.

\subsection{Zigzag Toolpath}
The zigzag toolpath is designed for efficient material removal by systematically traversing the remaining stock within a machining layer.
At each depth level $l$, after the contour toolpath has separated the object from the surrounding material, the zigzag strategy ensures that unremoved regions are cleared.
The toolpath follows a structured back-and-forth motion, covering the machining layer in a sweeping manner.
The traversal path is determined by selecting all points $P_{s,l}^{\text{out}}$ where the signed distance function value is greater than the tool radius:
\begin{equation}
P_{s,l}^{\text{out}} = \{ p \in P_l; \space \text{SDF}_s(p) > r_s^{\text{tool}} \}.
\label{eq:supp:out}
\end{equation}
The path begins at the starting point within the current viewpoint and progresses horizontally to the right.
Once the tool reaches the boundary, it moves one step upward and then traverses back to the left.
This alternating movement continues until all designated points in the layer have been processed, and the G-code is updated as in Equation~\eqref{eq:supp:gcode-g1}.
For this toolpath strategy, $(x_{s,l,i},y_{s,l,i},z_{s,l,i})$ is the location of a point $p_{s,l,i}\in P_{s,l}^{\text{out}}$ in the 3D space $\mathbb{R}^3$.

\subsection{Periphery Toolpath}
The periphery toolpath, similar to the zigzag strategy, is designed for efficient material removal by traversing the remaining stock within a machining layer.
However, unlike the zigzag strategy, where points are visited in a structured back-and-forth motion, the periphery toolpath orders the points in a counterclockwise sequence around the starting point in the layer $l$, and the G-code is updated as in Equation~\eqref{eq:supp:gcode-g1} following the same procedure as the zigzag toolpath.

\subsection{Spiral Toolpath}
Similar to the zigzag and periphery toolpaths, the spiral toolpath is designed for efficient material removal by traversing the remaining stock within a machining layer.
Instead of following a counterclockwise sequence, the tool moves along a continuous spiral trajectory outward from the starting point, and the G-code is updated as in Equation~\eqref{eq:supp:gcode-g1} following the same procedure as the zigzag toolpath.

\section{G-code Simulation}
\label{sec:code-simulation}

\begin{figure}[t]
\renewcommand{\figurename}{Algorithm}
\small
\hrule\vspace{2pt}
\textbf{Simulate Toolpath}\par
\textbf{Input:} voxel stock, $N{\times}4$ waypoint array, tool radius,
sweep step, grid origin/spacing, and tool axis.\par
\begin{enumerate}\setlength{\itemsep}{0pt}\setlength{\parskip}{0pt}
\item Initialize the removal mask to zero and visit consecutive waypoint pairs.
\item Skip rapid (G0) and zero-length segments, then normalize the cutting direction
and construct radial and axial tool-frame directions.
\item Draw swept-volume samples along each cutting segment and within the tool
radius, including samples around both segment endpoints.
\item Map sampled points to voxel indices and mark the intersected cells in the
removal mask.
\item Update the stock by element-wise multiplication with the complement of the
removal mask and return the resulting occupancy.
\end{enumerate}
\vspace{1pt}\hrule
\caption{Swept-volume voxel simulation used by the reported study.}
\label{alg:simulate}
\end{figure}

We simulate machining by carving a voxel stock along a polyline toolpath using a simple swept-volume approximation. The toolpath is a $N{\times}4$ array of waypoints $(x,y,z,f)$, where the flag $f\in\{0,1\}$ marks the ending waypoint of each segment as rapid (G0, $f{=}0$) or cutting (G1, $f{=}1$), and only G1 segments remove material. Safe moves are encoded as G0 segments, and layer height is implicitly determined by the $z$ coordinates of successive G1 segments.
A step-by-step simulation procedure is given in Algorithm~\ref{alg:simulate}.

The simulator models the swept volume of the cutting tool only, and disconnected segments are joined by retract-and-reposition moves, while consecutive points within a segment are joined by G1 cutting moves that may gouge and are penalized by $\mathcal{R}_s$. It does not model the tool shank and holder, the fixture and clamps, the stock beyond the machining envelope, or machine kinematic limits. Programs are additionally inspected in CIMCO, which checks the toolpath but is likewise not a full machine-and-fixture simulation. Because the collision model covers only the cutter, nothing in the reward prevents the agent from selecting a small tool radius for a deep, narrow cavity, a combination that in practice requires a long overhang and can bring the shank or holder into contact with unmachined stock. Two properties limit the exposure in our normalized unit cube. The radius is restricted to a bounded candidate set, and material is removed layer by layer from the top, so stock above the current layer has already been cleared over the cleared footprint. Neither is a guarantee. The natural extension is to add the shank and holder to the swept-volume test of Algorithm~\ref{alg:simulate} as a second, larger-radius cylinder above the flute length and to treat any intersection with remaining stock as an infeasible action, either by masking it out of $\mathcal{A}$ or by adding a penalty term to the reward. We leave this to future work.

\section{Reinforcement Learning Formulation}
\label{sec:supp:rl}

This section formalizes the partially observed decision process summarized in Section~\ref{sec:method}, in particular when an action is taken, what the latent state transition is, and what the policy observes.

\subsection{Episode Structure}
One episode corresponds to machining one object. Given a target shape $\mathcal{S}$, the orientation selection module first fixes an ordered set of machining orientations $V=\{v_1,\dots,v_n\}$. The episode then proceeds for exactly $n$ steps, and the agent takes \textbf{one action per orientation}, not per layer and not per toolpath segment. All layers and all toolpath segments generated for orientation $v_s$ are produced by the single action $a_s$, and the layer loop inside a step is deterministic given that action.

This means the machining model is multi-view 3-axis indexed (3+2 positioning), in which the rotary axes position the part at $v_s$ and are then held fixed while a 3-axis toolpath is executed. It is not continuous 5-axis simultaneous motion, in which all five axes interpolate within a single move. The orientation set therefore defines a sequence of indexed setups rather than a continuous tool-axis trajectory.

\subsection{Latent State and Policy Observation}
At step $s$, the environment state includes the evolving stock $V_{s-1}$, the target, and the indexed orientation. The policy observation is the latent code
\begin{equation}
\mathbf{z}_s = \mathcal{E}(\mathcal{S}_s) \in \mathbb{R}^{1024},
\end{equation}
where $\mathcal{S}_s$ is the visibility-based reconstruction of the target for orientation $v_s$ and $\mathcal{E}$ is a DGCNN encoder applied to $16{,}384$ points sampled from its surface.
This observation encodes the target view, not the changing stock occupancy. The current stock affects simulation and reward but is not explicitly supplied to the policy. Thus, the formulation is a finite-horizon partially observed decision process rather than a fully observed MDP, and the DQN should be read as a target-conditioned selector trained to maximize expected return under the training distribution rather than a stock-adaptive controller. This limitation is explicit, since the policy cannot condition its next action on action-dependent residual-stock deviations that are not predictable from the target view.

\subsection{Action}
The action is the joint choice
\begin{equation}
a_s=\{r_s^{\text{tool}},\, h_s^{\text{layer}},\, s_s^{\text{strategy}}\}\in\mathcal{A},
\end{equation}
a discretized tool radius, layer height, and clearing strategy. Contour is always applied to preserve the surface, so $s_s^{\text{strategy}}$ selects among zigzag, periphery, and spiral. $\mathcal{A}$ is therefore the product of the three discrete sets, which is what keeps the decision space small relative to per-move action spaces.

\subsection{Transition}
Executing $a_s$ emits the G-code block for orientation $v_s$, which the voxel simulator runs against the current stock:
\begin{equation}
V_{s} = V_{s-1}\odot\bigl(1-\text{erase}(a_s, v_s)\bigr),
\end{equation}
starting from a full stock $V_0=\mathbf{1}$. Once the sampled swept-volume points are fixed, the transition is deterministic given $(V_{s-1}, a_s, v_s)$. Sampling is an approximation used to instantiate that transition, while action stochasticity comes from $\epsilon$-greedy exploration. The episode terminates after the $n$-th orientation.

\subsection{Reward and Objective}
Each step returns the preservation term $\mathcal{R}_s$, and the terminal step additionally returns the terminal fidelity term $\mathcal{R}_f$ and the efficiency term $\mathcal{R}_{\text{MRR}}$, combined as in Equation~(\ref{eq:reward}). The network is trained with the standard DQN objective
\begin{equation}
\mathcal{L}(\theta)=\mathbb{E}\Bigl[\bigl(r_s+(1-d_s)\gamma\max_{a'}Q_{\theta^-}(\mathbf{z}_{s+1},a')-Q_\theta(\mathbf{z}_s,a_s)\bigr)^2\Bigr],
\end{equation}
where $r_s$ is the weighted reward of step $s$, $d_s=1$ at the terminal orientation and is zero otherwise, so no value is bootstrapped past termination. We use discount $\gamma=0.95$, a target network $Q_{\theta^-}$, experience replay, and an $\epsilon$-greedy behavior policy decaying from $1.0$ to $0.01$. Training runs for 2,000 episodes in total, each sampling one object with replacement from the corresponding 5,000-object ABC or 35,000-object ShapeNet training pool, rather than making an exhaustive pass through that pool. Each selected orientation produces one environment transition, with at most five orientations per episode. Thus, training generates at most 10,000 environment transitions, and the actual count depends on the number of selected orientations. The $Q$-network is two fully connected layers of $128$ units on top of $\mathbf{z}_s$.

\section{More Qualitative Results}

\begin{figure*}[t]
    \centering

    \begin{subfigure}[b]{0.10\textwidth}
        \includegraphics[clip,width=\textwidth, page=1]{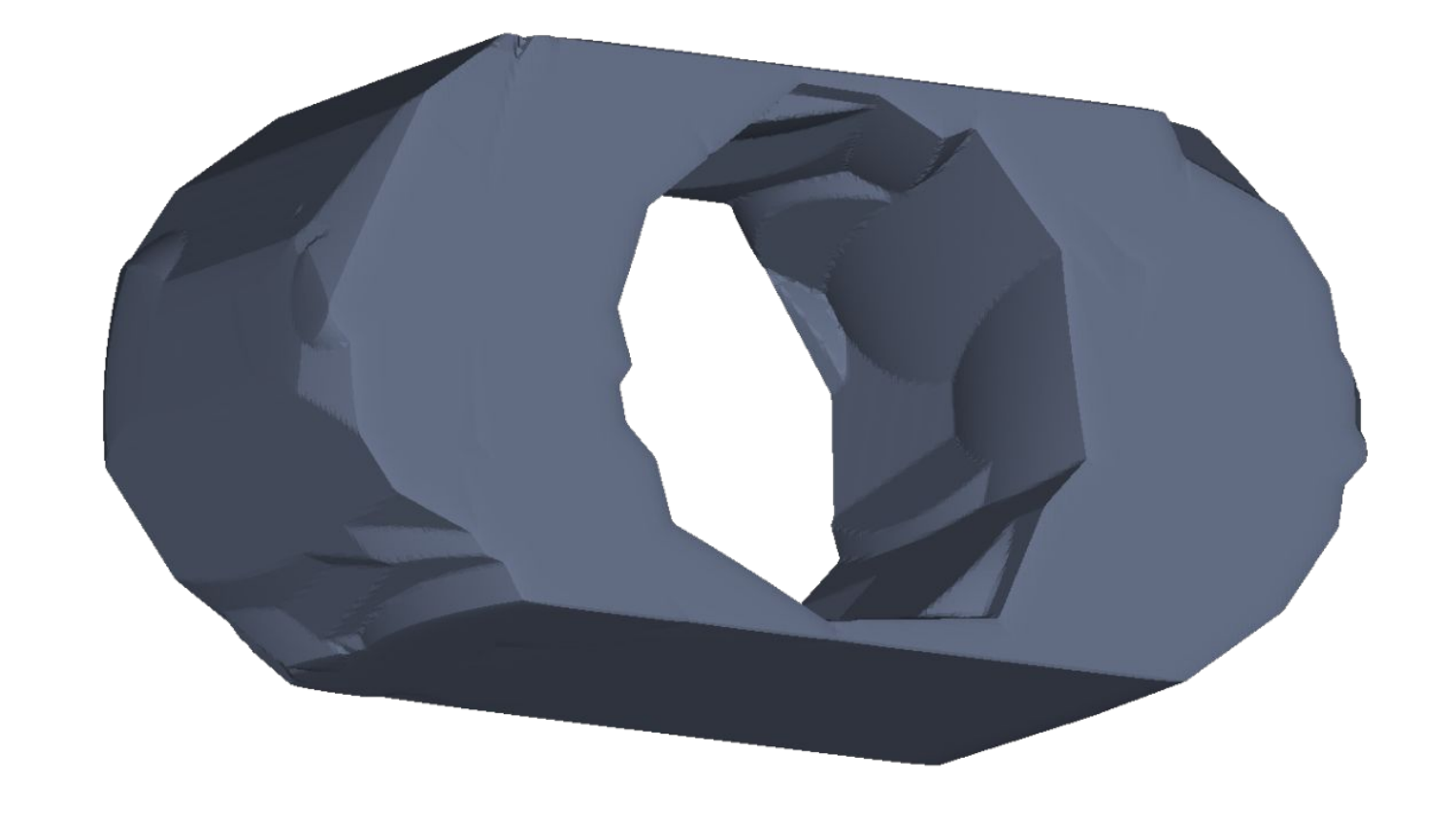}
    \end{subfigure}
    \hfill
    \begin{subfigure}[b]{0.10\textwidth}
        \includegraphics[clip,width=\textwidth, page=2]{figure_assets/supp_abc_qual1.pdf}
    \end{subfigure}
    \hfill
    \begin{subfigure}[b]{0.10\textwidth}
        \includegraphics[clip,width=\textwidth, page=3]{figure_assets/supp_abc_qual1.pdf}
    \end{subfigure}
    \hfill
    \begin{subfigure}[b]{0.10\textwidth}
        \includegraphics[clip,width=\textwidth, page=4]{figure_assets/supp_abc_qual1.pdf}
    \end{subfigure}
    \hfill
    \begin{subfigure}[b]{0.10\textwidth}
        \includegraphics[clip,width=\textwidth, page=5]{figure_assets/supp_abc_qual1.pdf}
    \end{subfigure}
    \hfill
    \begin{subfigure}[b]{0.10\textwidth}
        \includegraphics[clip,width=\textwidth, page=6]{figure_assets/supp_abc_qual1.pdf}
    \end{subfigure}
    \hfill
    \begin{subfigure}[b]{0.10\textwidth}
        \includegraphics[clip,width=\textwidth, page=7]{figure_assets/supp_abc_qual1.pdf}
    \end{subfigure}

    \begin{subfigure}[b]{0.12\textwidth}
        \includegraphics[trim=100 0 80 0,clip,width=\textwidth, page=1]{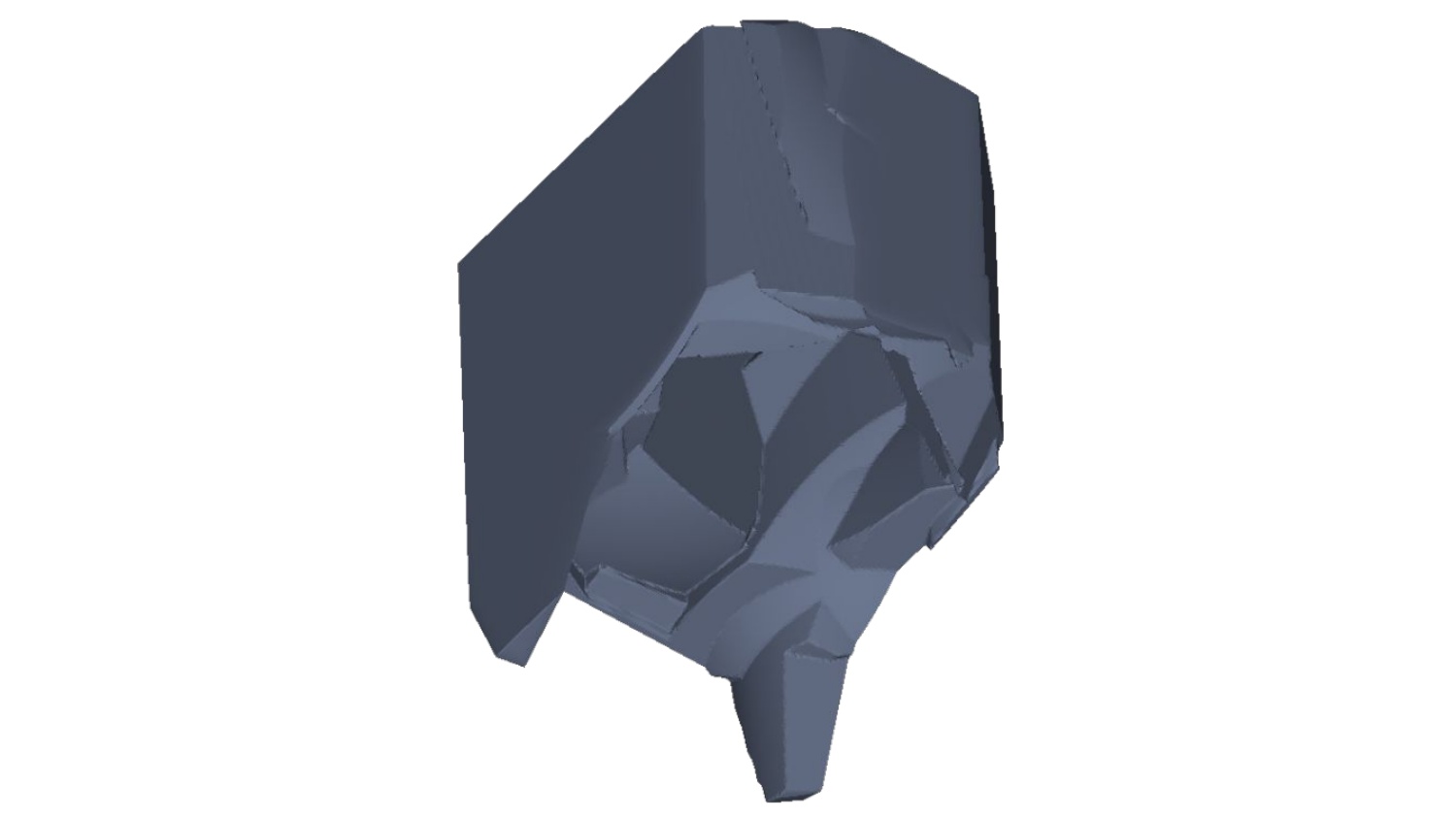}
    \end{subfigure}
    \hfill
    \begin{subfigure}[b]{0.12\textwidth}
        \includegraphics[trim=100 0 80 0,clip,width=\textwidth, page=2]{figure_assets/supp_abc_qual2.pdf}
    \end{subfigure}
    \hfill
    \begin{subfigure}[b]{0.12\textwidth}
        \includegraphics[trim=100 0 80 0,clip,width=\textwidth, page=3]{figure_assets/supp_abc_qual2.pdf}
    \end{subfigure}
    \hfill
    \begin{subfigure}[b]{0.12\textwidth}
        \includegraphics[trim=100 0 80 0,clip,width=\textwidth, page=4]{figure_assets/supp_abc_qual2.pdf}
    \end{subfigure}
    \hfill
    \begin{subfigure}[b]{0.12\textwidth}
        \includegraphics[trim=100 0 80 0,clip,width=\textwidth, page=5]{figure_assets/supp_abc_qual2.pdf}
    \end{subfigure}
    \hfill
    \begin{subfigure}[b]{0.12\textwidth}
        \includegraphics[trim=100 0 80 0,clip,width=\textwidth, page=6]{figure_assets/supp_abc_qual2.pdf}
    \end{subfigure}
    \hfill
    \begin{subfigure}[b]{0.12\textwidth}
        \includegraphics[trim=100 0 80 0,clip,width=\textwidth, page=7]{figure_assets/supp_abc_qual2.pdf}
    \end{subfigure}

    \begin{subfigure}[b]{0.12\textwidth}
        \includegraphics[trim=100 0 80 0,clip,width=\textwidth, page=1]{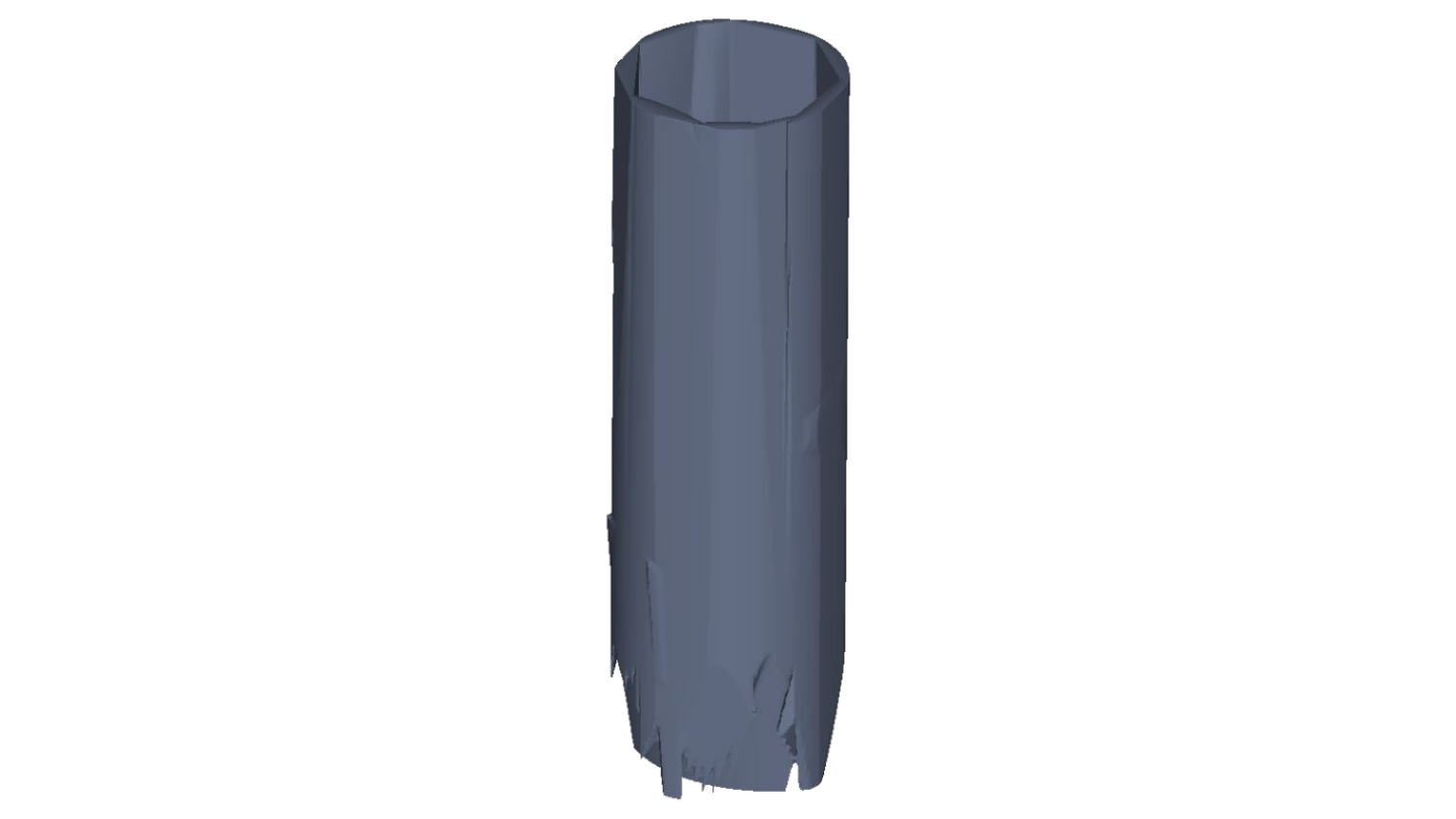}
    \end{subfigure}
    \hfill
    \begin{subfigure}[b]{0.12\textwidth}
        \includegraphics[trim=100 0 80 0,clip,width=\textwidth, page=2]{figure_assets/supp_abc_qual3.pdf}
    \end{subfigure}
    \hfill
    \begin{subfigure}[b]{0.12\textwidth}
        \includegraphics[trim=100 0 80 0,clip,width=\textwidth, page=3]{figure_assets/supp_abc_qual3.pdf}
    \end{subfigure}
    \hfill
    \begin{subfigure}[b]{0.12\textwidth}
        \includegraphics[trim=100 0 80 0,clip,width=\textwidth, page=4]{figure_assets/supp_abc_qual3.pdf}
    \end{subfigure}
    \hfill
    \begin{subfigure}[b]{0.12\textwidth}
        \includegraphics[trim=100 0 80 0,clip,width=\textwidth, page=5]{figure_assets/supp_abc_qual3.pdf}
    \end{subfigure}
    \hfill
    \begin{subfigure}[b]{0.12\textwidth}
        \includegraphics[trim=100 0 80 0,clip,width=\textwidth, page=6]{figure_assets/supp_abc_qual3.pdf}
    \end{subfigure}
    \hfill
    \begin{subfigure}[b]{0.12\textwidth}
        \includegraphics[trim=100 0 80 0,clip,width=\textwidth, page=7]{figure_assets/supp_abc_qual3.pdf}
    \end{subfigure}

    \begin{subfigure}[b]{0.12\textwidth}
        \includegraphics[trim=100 0 80 0,clip,width=\textwidth, page=1]{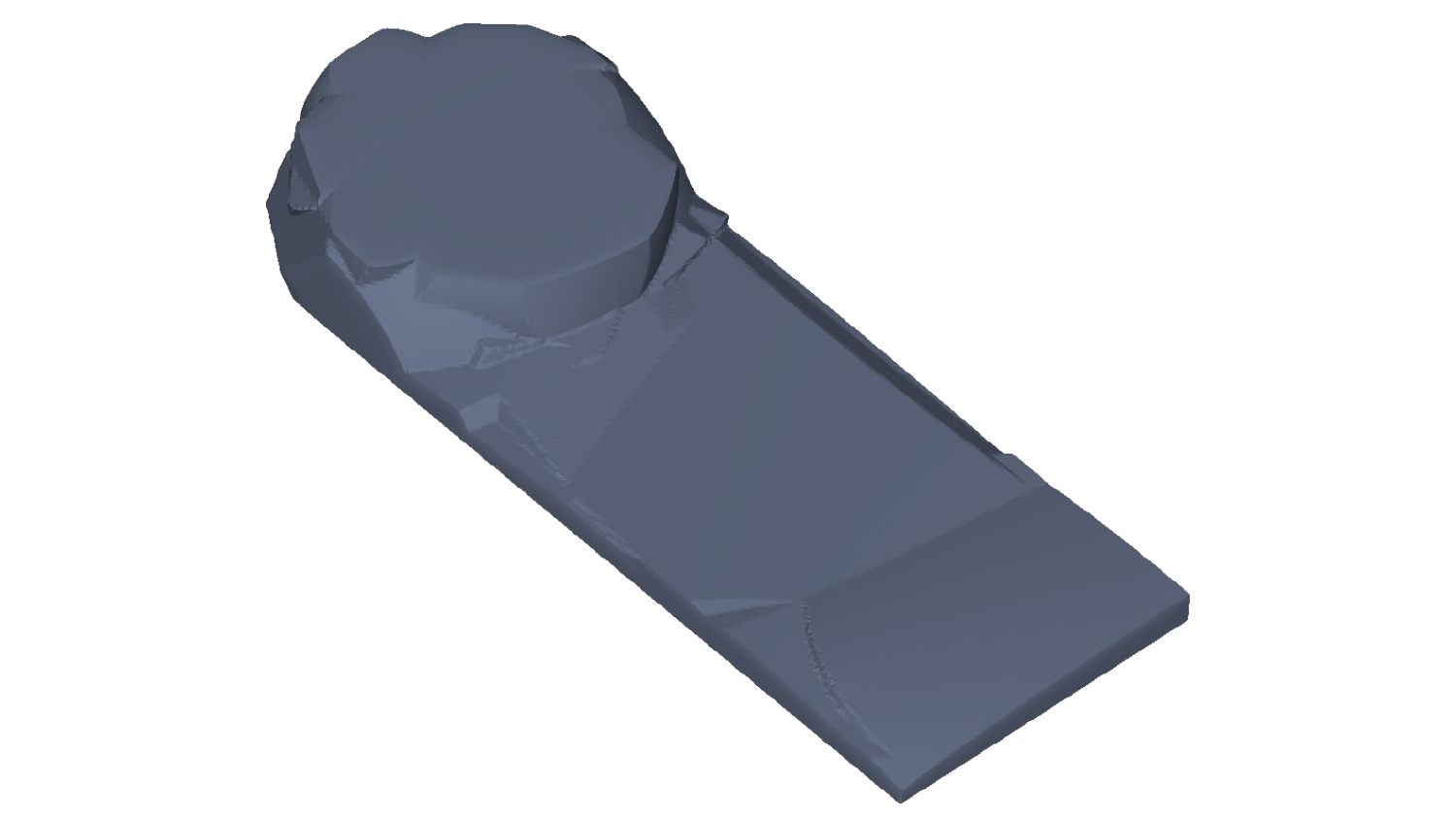}
    \end{subfigure}
    \hfill
    \begin{subfigure}[b]{0.12\textwidth}
        \includegraphics[trim=100 0 80 0,clip,width=\textwidth, page=2]{figure_assets/supp_abc_qual4.pdf}
    \end{subfigure}
    \hfill
    \begin{subfigure}[b]{0.12\textwidth}
        \includegraphics[trim=100 0 80 0,clip,width=\textwidth, page=3]{figure_assets/supp_abc_qual4.pdf}
    \end{subfigure}
    \hfill
    \begin{subfigure}[b]{0.12\textwidth}
        \includegraphics[trim=100 0 80 0,clip,width=\textwidth, page=4]{figure_assets/supp_abc_qual4.pdf}
    \end{subfigure}
    \hfill
    \begin{subfigure}[b]{0.12\textwidth}
        \includegraphics[trim=100 0 80 0,clip,width=\textwidth, page=5]{figure_assets/supp_abc_qual4.pdf}
    \end{subfigure}
    \hfill
    \begin{subfigure}[b]{0.12\textwidth}
        \includegraphics[trim=100 0 80 0,clip,width=\textwidth, page=6]{figure_assets/supp_abc_qual4.pdf}
    \end{subfigure}
    \hfill
    \begin{subfigure}[b]{0.12\textwidth}
        \includegraphics[trim=100 0 80 0,clip,width=\textwidth, page=7]{figure_assets/supp_abc_qual4.pdf}
    \end{subfigure}

    \begin{subfigure}[b]{0.12\textwidth}
        \includegraphics[trim=100 0 80 0,clip,width=\textwidth, page=1]{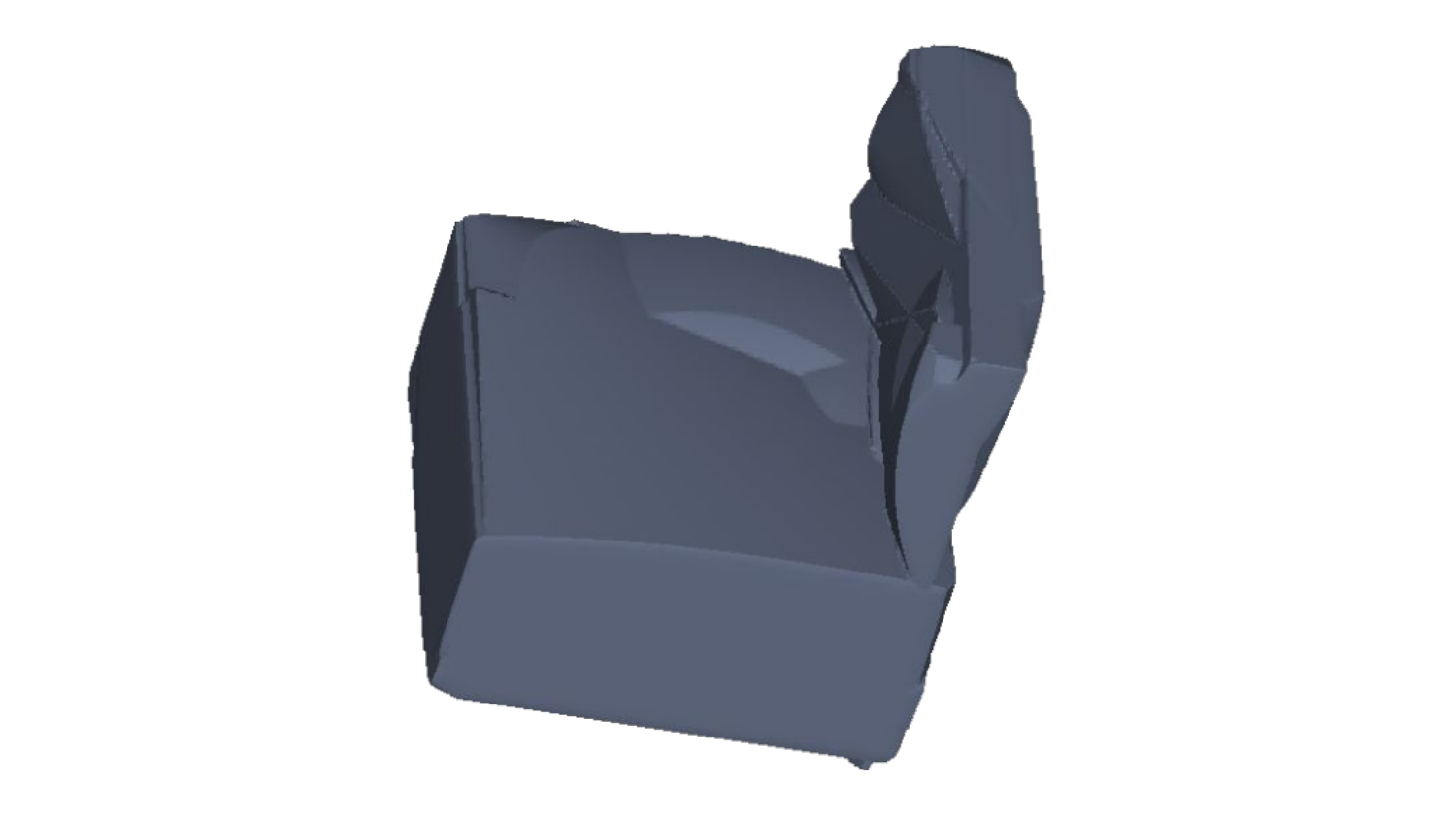}
    \end{subfigure}
    \hfill
    \begin{subfigure}[b]{0.12\textwidth}
        \includegraphics[trim=100 0 80 0,clip,width=\textwidth, page=2]{figure_assets/Geo2Gcode_shapenet_qual3_final.pdf}
    \end{subfigure}
    \hfill
    \begin{subfigure}[b]{0.12\textwidth}
        \includegraphics[trim=100 0 80 0,clip,width=\textwidth, page=3]{figure_assets/Geo2Gcode_shapenet_qual3_final.pdf}
    \end{subfigure}
    \hfill
    \begin{subfigure}[b]{0.12\textwidth}
        \includegraphics[trim=100 0 80 0,clip,width=\textwidth, page=4]{figure_assets/Geo2Gcode_shapenet_qual3_final.pdf}
    \end{subfigure}
    \hfill
    \begin{subfigure}[b]{0.12\textwidth}
        \includegraphics[trim=100 0 80 0,clip,width=\textwidth, page=5]{figure_assets/Geo2Gcode_shapenet_qual3_final.pdf}
    \end{subfigure}
    \hfill
    \begin{subfigure}[b]{0.12\textwidth}
        \includegraphics[trim=100 0 80 0,clip,width=\textwidth, page=6]{figure_assets/Geo2Gcode_shapenet_qual3_final.pdf}
    \end{subfigure}
    \hfill
    \begin{subfigure}[b]{0.12\textwidth}
        \includegraphics[trim=100 0 80 0,clip,width=\textwidth, page=7]{figure_assets/Geo2Gcode_shapenet_qual3_final.pdf}
    \end{subfigure}

    \begin{subfigure}[b]{0.12\textwidth}
        \includegraphics[trim=100 0 80 0,clip,width=\textwidth, page=1]{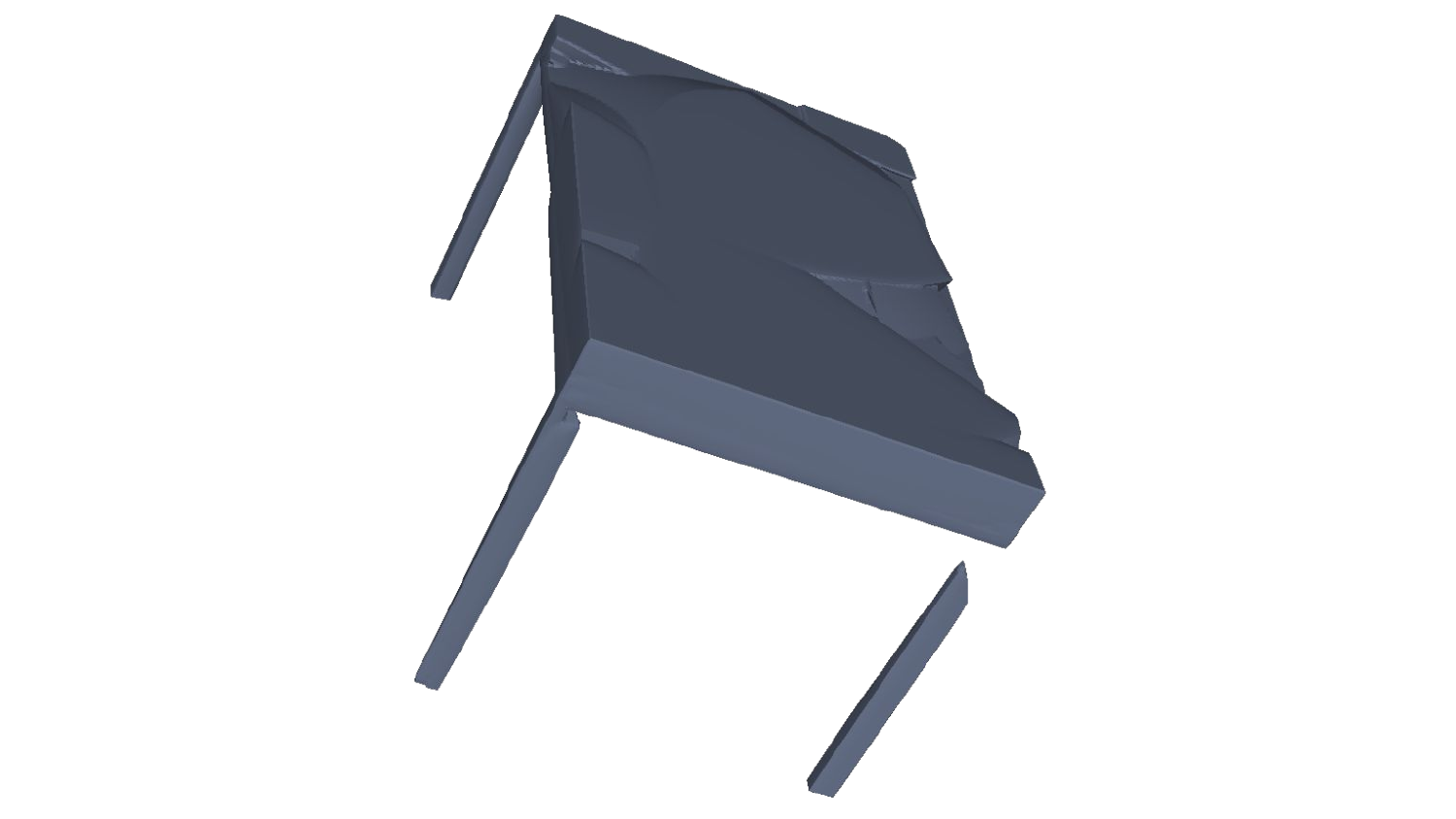}
        \caption*{\footnotesize\mbox{CSG-Stump}}
    \end{subfigure}
    \hfill
    \begin{subfigure}[b]{0.12\textwidth}
        \includegraphics[trim=100 0 80 0,clip,width=\textwidth, page=2]{figure_assets/Geo2Gcode_shapenet_qual5_final.pdf}
        \caption*{\footnotesize\mbox{ExtrudeNet}}
    \end{subfigure}
    \hfill
    \begin{subfigure}[b]{0.12\textwidth}
        \includegraphics[trim=100 0 80 0,clip,width=\textwidth, page=3]{figure_assets/Geo2Gcode_shapenet_qual5_final.pdf}
        \caption*{\footnotesize\mbox{CAPRI-Net}}
    \end{subfigure}
    \hfill
    \begin{subfigure}[b]{0.12\textwidth}
        \includegraphics[trim=100 0 80 0,clip,width=\textwidth, page=4]{figure_assets/Geo2Gcode_shapenet_qual5_final.pdf}
        \caption*{\footnotesize\mbox{SECAD-Net}}
    \end{subfigure}
    \hfill
    \begin{subfigure}[b]{0.12\textwidth}
        \includegraphics[trim=100 0 80 0,clip,width=\textwidth, page=5]{figure_assets/Geo2Gcode_shapenet_qual5_final.pdf}
        \caption*{\footnotesize\mbox{CNC-Net}}
    \end{subfigure}
    \hfill
    \begin{subfigure}[b]{0.12\textwidth}
        \includegraphics[trim=100 0 80 0,clip,width=\textwidth, page=6]{figure_assets/Geo2Gcode_shapenet_qual5_final.pdf}
        \caption*{\footnotesize\mbox{Ours}}
    \end{subfigure}
    \hfill
    \begin{subfigure}[b]{0.12\textwidth}
        \includegraphics[trim=100 0 80 0,clip,width=\textwidth, page=7]{figure_assets/Geo2Gcode_shapenet_qual5_final.pdf}
        \caption*{\footnotesize\mbox{Target (GT)}}
    \end{subfigure}

    \caption{\textbf{Qualitative comparisons on the ABC and ShapeNet datasets.} Our outputs are rendered in the CIMCO G-code simulator. Red highlights the simulated machined surface exposed by the cutting operations, and gray denotes stock that remains after simulation.}
    \label{fig:supp_qual}
\end{figure*}

We extend our qualitative evaluation with comparisons to CSG-Stump, ExtrudeNet, CAPRI-Net, SECAD-Net, and the CNC-based CNC-Net on ABC and ShapeNet. Figure~\ref{fig:supp_qual} shows that the simulated outputs retain the principal structures of the targets across varied shapes. These examples are qualitative evidence from the same simulator used for the quantitative evaluation, but they do not by themselves establish dimensional tolerance or production readiness.

\section{Effect of Machining Precision}
\label{sec:supp:precision}

\begin{figure*}[t]
    \centering

    \begin{subfigure}[b]{0.15\textwidth}
        \includegraphics[width=\textwidth, page=1]{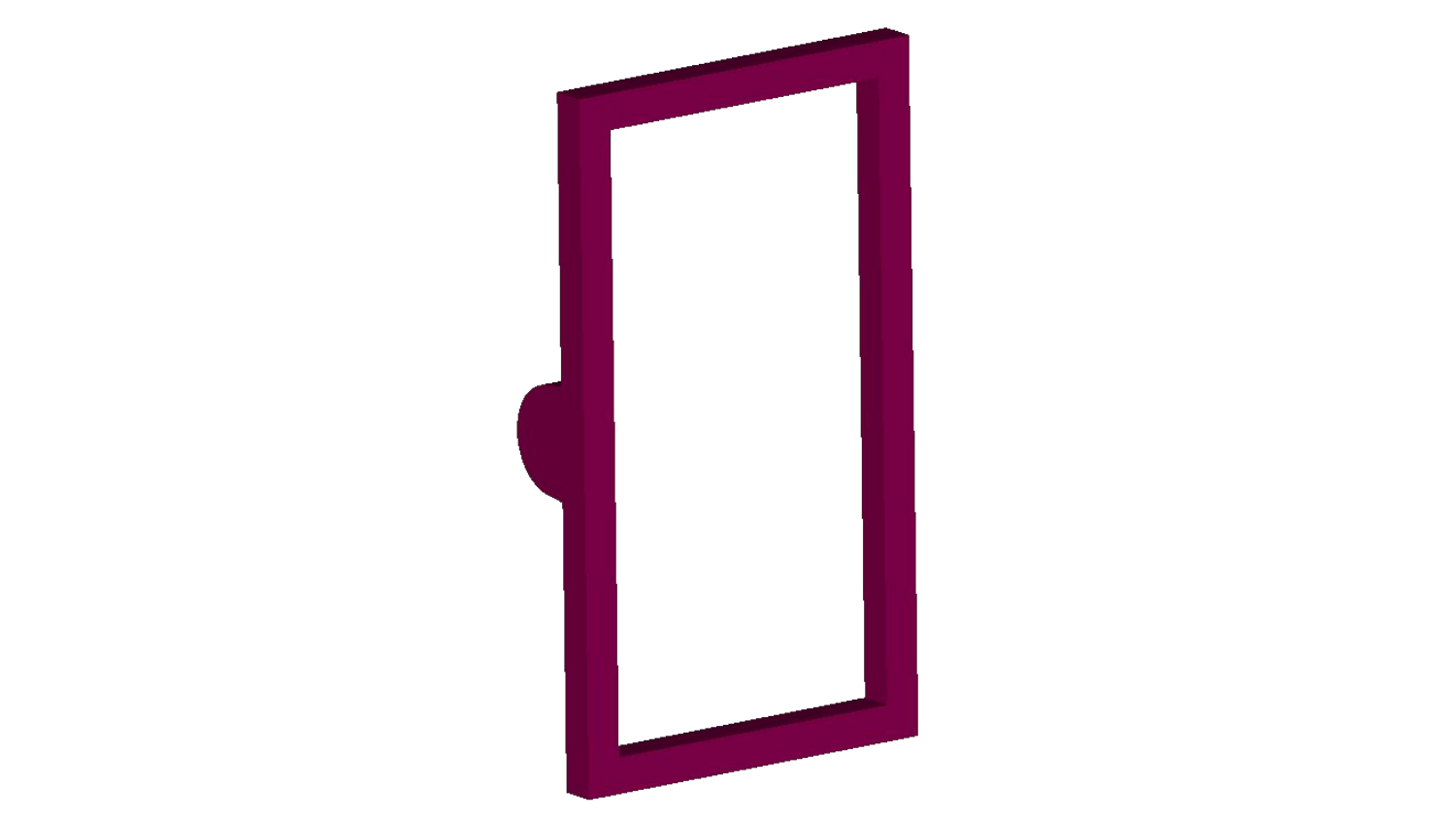}
    \end{subfigure}
    \hfill
    \begin{subfigure}[b]{0.15\textwidth}
        \includegraphics[width=\textwidth, page=2]{figure_assets/supp_abc_compare_high.pdf}
    \end{subfigure}
    \hfill
    \begin{subfigure}[b]{0.15\textwidth}
        \includegraphics[width=\textwidth, page=3]{figure_assets/supp_abc_compare_high.pdf}
    \end{subfigure}
    \hfill
    \begin{subfigure}[b]{0.15\textwidth}
        \includegraphics[width=\textwidth, page=4]{figure_assets/supp_abc_compare_high.pdf}
    \end{subfigure}
    \hfill
    \begin{subfigure}[b]{0.15\textwidth}
        \includegraphics[width=\textwidth, page=5]{figure_assets/supp_abc_compare_high.pdf}
    \end{subfigure}
    \hfill
    \begin{subfigure}[b]{0.15\textwidth}
        \includegraphics[width=\textwidth, page=6]{figure_assets/supp_abc_compare_high.pdf}
    \end{subfigure}

    \vspace{2mm}

    \begin{subfigure}[b]{0.15\textwidth}
        \includegraphics[width=\textwidth, page=7]{figure_assets/supp_abc_compare_high.pdf}
        \caption*{\footnotesize Target (GT)}
    \end{subfigure}
    \hfill
    \begin{subfigure}[b]{0.15\textwidth}
        \includegraphics[width=\textwidth, page=8]{figure_assets/supp_abc_compare_high.pdf}
        \caption*{\footnotesize Fixed parameter}
    \end{subfigure}
    \hfill
    \begin{subfigure}[b]{0.15\textwidth}
        \includegraphics[width=\textwidth, page=9]{figure_assets/supp_abc_compare_high.pdf}
        \caption*{\footnotesize Ours}
    \end{subfigure}
    \hfill
    \begin{subfigure}[b]{0.15\textwidth}
        \includegraphics[width=\textwidth, page=10]{figure_assets/supp_abc_compare_high.pdf}
        \caption*{\footnotesize Target (GT)}
    \end{subfigure}
    \hfill
    \begin{subfigure}[b]{0.15\textwidth}
        \includegraphics[width=\textwidth, page=11]{figure_assets/supp_abc_compare_high.pdf}
        \caption*{\footnotesize Fixed parameter}
    \end{subfigure}
    \hfill
    \begin{subfigure}[b]{0.15\textwidth}
        \includegraphics[width=\textwidth, page=12]{figure_assets/supp_abc_compare_high.pdf}
        \caption*{\footnotesize Ours}
    \end{subfigure}

    \caption{\textbf{Qualitative effect of machining precision.} Examples show the target, fixed coarse parameters (large tool radius and layer height), and our method.}
    \label{fig:param_select}
\end{figure*}

Table~\ref{tab:parameter_controls} reports the quantitative comparison of parameter policies. Fixed (fine) uses the smallest radius and layer height of the action space, Fixed (coarse) the largest, and Fixed (const.\ triple) holds one radius--height--pattern triple constant across all shapes and views. These are conditional parameter comparisons with shared orientations and sequential residual-stock propagation, not a sweep of every fixed setting.

Figure~\ref{fig:param_select} shows the corresponding qualitative behavior. Always choosing the largest parameter setting frequently leads to low-resolution geometry and regions that remain uncarved, because a large tool radius cannot reach narrow concavities and a large layer height coarsely discretizes the depth direction. The learned selector produces an intermediate accuracy--time operating point and also outperforms per-view greedy simulator search (Table~\ref{tab:parameter_controls}).

\section{Comparison with the RL-based Method}

\begin{figure}[t]
\centering
    \includegraphics[width=\columnwidth]{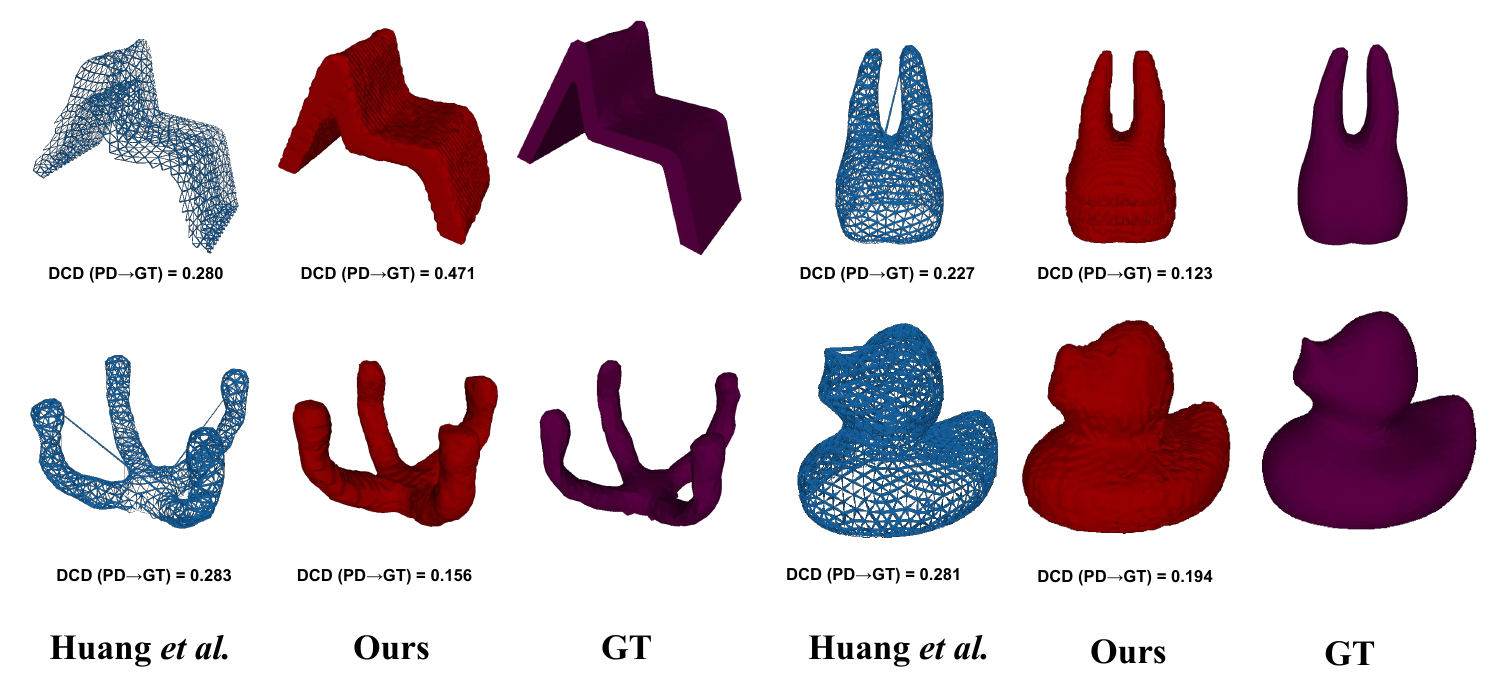}
    \caption{\textbf{Comparison with \citet{huang2024toolpath}.} While their method performs sample-specific fitting on these examples, our method produces competitive results in a zero-shot manner without per-sample optimization.}
    \label{fig:RL_comp}
\end{figure}

We include \citet{huang2024toolpath} as an illustrative cross-task comparison to an RL toolpath method developed for 3D printing, not as a direct machining baseline. We use the samples released by the authors because their pipeline assumes a wireframe-mesh input and is not directly applicable to our mesh-based setting. We therefore report qualitative results and directed Chamfer distance (PD$\rightarrow$GT) on their public examples. Their method performs sample-specific fitting, whereas ours is applied without per-sample fitting, so the comparison demonstrates transfer across task formulations but does not establish superiority under a matched protocol.

\section{Comparison with DeepMill on Inaccessibility Prediction}

\begin{table}[t]
\centering
\small
\setlength{\tabcolsep}{4pt}
\begin{tabular}{lcccc}
\toprule
 & \multicolumn{2}{c}{ABC 7K} & \multicolumn{2}{c}{ABC 15K} \\
\cmidrule(lr){2-3}\cmidrule(lr){4-5}
Method & Acc. & F1 & Acc. & F1 \\
\midrule
DeepMill & 96.3 & 97.2 & 96.3 & 97.3 \\
Ours & \textbf{96.5} & \textbf{98.2} & \textbf{96.8} & \textbf{98.4} \\
\bottomrule
\end{tabular}
\caption{Quantitative comparison with DeepMill on the ABC dataset.}
\label{tab:deepmill_comp}
\end{table}

We further compare our orientation selection module with DeepMill~\cite{zhong2025deepmill} on the inaccessibility prediction task. As shown in Table~\ref{tab:deepmill_comp}, the evaluation uses the remeshed ABC 7K and ABC 15K splits and DeepMill's ball-end cutter setting. This is an auxiliary evaluation distinct from the ray-visibility proxy used to select our orientations. After selection, we apply the benchmark's ball-end collision test to the selected orientations $V$ and mark a surface point accessible when at least one direction reaches it without collision. The resulting mask is compared with the ground-truth accessibility annotation using accuracy and F1.

As reported in Table~\ref{tab:deepmill_comp}, our method achieves slightly better accuracy and consistently higher F1 scores than DeepMill on both ABC 7K and ABC 15K. These results support the orientation module on the two stated ABC benchmark splits. Because this auxiliary protocol compares an exact post-selection collision test with a learned predictor, it is not evidence of general superiority in learned accessibility prediction.

\section{Machining-Relevant Metrics}
\label{sec:supp:machining}

\begin{table*}[t]
\centering
\small
\setlength{\tabcolsep}{5pt}
\begin{tabular}{lcccc}
\toprule
\textbf{Method} & \textbf{Path Smoothness ($^\circ$)}$\downarrow$ & \textbf{Curv.\ consistency}$\downarrow$ & \textbf{G1 segment-length std. (mm)}$\downarrow$ & \textbf{Accel.\ violation}$\downarrow$ \\
\midrule
MeshCAM (non-expert) & 52.4 $\pm$ 9.6 & 1.15 $\pm$ 0.25 & 12.28 $\pm$ 7.21 & \textbf{1.77e-6} \\
MeshCAM (expert) & 39.9 $\pm$ 17.3 & 7.57 $\pm$ 12.2 & 8.23 $\pm$ 4.12 & 4.0e-4 \\
Ours & \textbf{27.1 $\pm$ 14.2} & \textbf{0.69 $\pm$ 0.24} & \textbf{7.75 $\pm$ 2.88} & 1.32e-5 \\
\bottomrule
\end{tabular}
\caption{Machining-relevant metrics comparison on the ABC dataset between our method and MeshCAM under non-expert and expert settings.}
\label{tab:supp:machining_metric}
\end{table*}

We compare the machining quality of our generated G-code with two MeshCAM baselines on the same ABC test set. The path-regularity proxies in this section map each normalized cube to a 100\,mm largest dimension and assume G1 feed 100\,mm/min, G0 rate 3000\,mm/min, and maximum acceleration 500\,mm/s$^2$. These assumptions are distinct from $T$, which is defined alongside Equation~(\ref{eq:reward}) as normalized path length at a uniform feed. We therefore interpret only within-table $T$ ratios and do not present $T$ as seconds.

We compute four machining-relevant metrics directly from the generated toolpaths.
\textbf{Path smoothness} is measured as the turning angle at each junction between consecutive toolpath segments, computed from their direction vectors. A smaller value indicates a smoother path with less abrupt direction change.
\textbf{Curvature consistency} is measured by defining a curvature proxy at each junction as the turning angle divided by the average length of the two adjacent segments, and then reporting the standard deviation of this quantity along the toolpath. Lower values indicate more consistent local curvature.
\textbf{G1 segment-length variation} is the standard deviation of cutting-move lengths. It measures discretization regularity along the emitted polyline, not the geometric step-over distance between neighboring passes.
\textbf{Acceleration violation rate} measures the fraction of junctions where the centripetal acceleration required to follow the path at the assumed feed rate exceeds the machine limit of 500 mm/s$^2$. Lower values indicate better compatibility with practical machine dynamics.

Table~\ref{tab:supp:machining_metric} summarizes the results. Our method has the lowest mean junction angle ($27.1^\circ$), curvature-proxy standard deviation ($0.69$), and G1 segment-length standard deviation ($7.75$ mm) in this comparison. Its acceleration-violation proxy is $1.32 \times 10^{-5}$, between the two MeshCAM settings. These path-level proxies describe the emitted polylines under the stated feed assumptions and do not replace machine-specific dynamics or surface-finish measurements.

The generated program uses a fixed cutting feed for G1 moves and a fixed rapid rate for G0 moves, and the machining-relevant metrics assume $100$\,mm/min and $3000$\,mm/min respectively. Feed is therefore not modulated by instantaneous tool engagement or removal volume, as adaptive or trochoidal strategies would do. Adding engagement-aware feed scheduling is compatible with the current formulation, since it changes the emitted feed words rather than the toolpath geometry, but it is not evaluated here.

\section{CAD-Surface Overcut Proxies}
\label{sec:supp:tolerance}

\begin{table*}[t]
\centering
\small
\setlength{\tabcolsep}{6pt}
\begin{tabular}{lcccc}
\toprule
\textbf{Method} &
\shortstack{\textbf{Local dim.}\\\textbf{variation}$\downarrow$} &
\shortstack{\textbf{Gouging tol.}\\$\leq0.1$\,\textbf{mm} (\%)$\uparrow$} &
\shortstack{\textbf{Gouging tol.}\\$\leq0.2$\,\textbf{mm} (\%)$\uparrow$} &
\shortstack{\textbf{Mean per-part P95}\\\textbf{overcut} (mm)$\downarrow$} \\
\midrule
MeshCAM (non-expert) & 0.678 & 89.61 & 90.03 & 0.249 \\
MeshCAM (expert)     & 0.726 & 92.31 & 92.53 & 0.262 \\
Ours                 & \textbf{0.228} & \textbf{98.57} & \textbf{99.29} & \textbf{0.217} \\
\bottomrule
\end{tabular}
\caption{\textbf{One-sided overcut proxies on CAD-origin ABC target samples (all 50 parts).} Each metric is computed per part (expert setups averaged within the part) and then averaged with equal part weight. P95 is conditional on gouged samples within each part.}
\label{tab:supp:tolerance}
\end{table*}

The metrics of Section~\ref{sec:metrics} (IoU, F1, CD, NC) are computed from meshes and therefore do not express accuracy in physical units. To evaluate accuracy directly against the design intent, we additionally measure the error of the generated G-code with respect to samples of the CAD-origin target mesh used for generation (normalized and scaled to 100\,mm), independently of the $256^3$ simulation grid.

For each target sample, the signed error is its distance to the nearest emitted cutter position minus the tool radius. \textbf{Local dimensional variation} is the standard deviation of neighboring signed errors. \textbf{Gouging compliance} is the one-sided percentage of all sampled surface points whose overcut stays below $0.1$\,mm or $0.2$\,mm. \textbf{P95 overcut} is computed only over positively gouged samples within each part (zero if none) and then averaged across parts. Consequently, a high all-sample compliance rate can coexist with a conditional P95 above the threshold. These are overcut proxies, not two-sided dimensional-tolerance or worst-case guarantees.

Table~\ref{tab:supp:tolerance} reports these measurements. Our method has local variation $0.228$, mean one-sided $0.1$\,mm compliance $98.57\%$, and mean conditional P95 overcut $0.217$\,mm. The table supports a lower-overcut comparison under this proxy, whereas residual stock, absolute error, and finishing tolerance are outside its scope.

\section{Physical CNC Machining Results}
\label{sec:supp:physical}

\begin{figure}[t]
    \centering
    \includegraphics[width=\columnwidth]{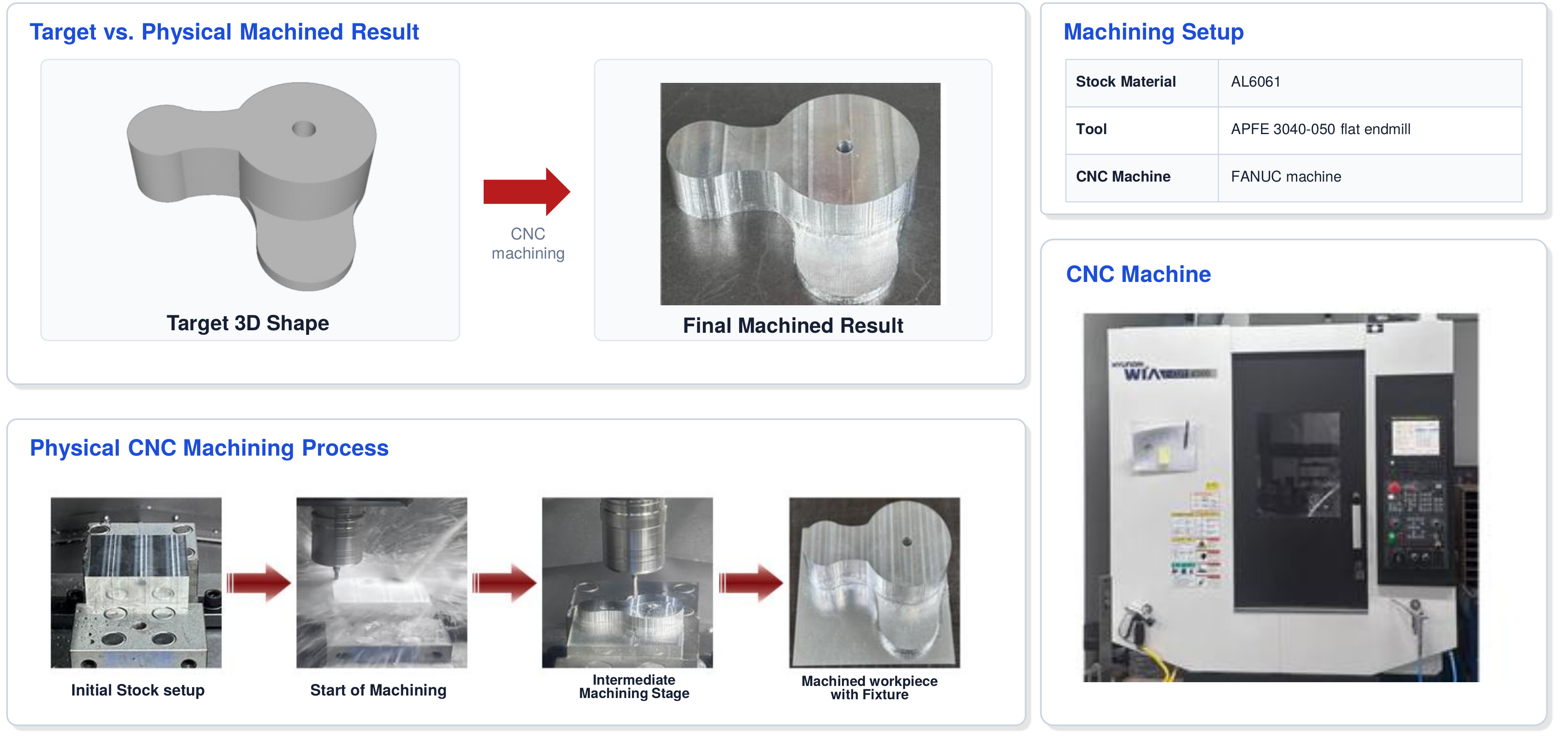}
    \caption{\textbf{Qualitative physical CNC case study.} One part machined from one Geo2Gcode-generated program on a FANUC Robodrill using AL6061 stock and a flat endmill.}
    \label{fig:supp:physical}
\vspace{-3mm}
\end{figure}

All quantitative results use the voxel and CIMCO simulators. We additionally executed one generated program for the pictured part on a FANUC Robodrill with AL6061 stock and a flat endmill. The normalized path was scaled to this job's coordinates, so its physical scale and cutter should not be inferred from the normalized synthetic parameter grid or its time proxy. The program was used without edits to cutting coordinates or path ordering, and an operator performed ordinary setup and machine operation and remained responsible for workholding and machine-envelope clearance. Thus, no learned collision-safety claim is inferred from this run. This single-part case study establishes qualitative controller acceptance and physical feasibility for that example, not dimensional tolerance, universal collision safety, dataset-level verification, or edit-free autonomous manufacturing.

\section{Analysis on Complex Shapes}
\label{sec:supp:complex}

\begin{figure}[t]
\centering
    \includegraphics[width=\columnwidth]{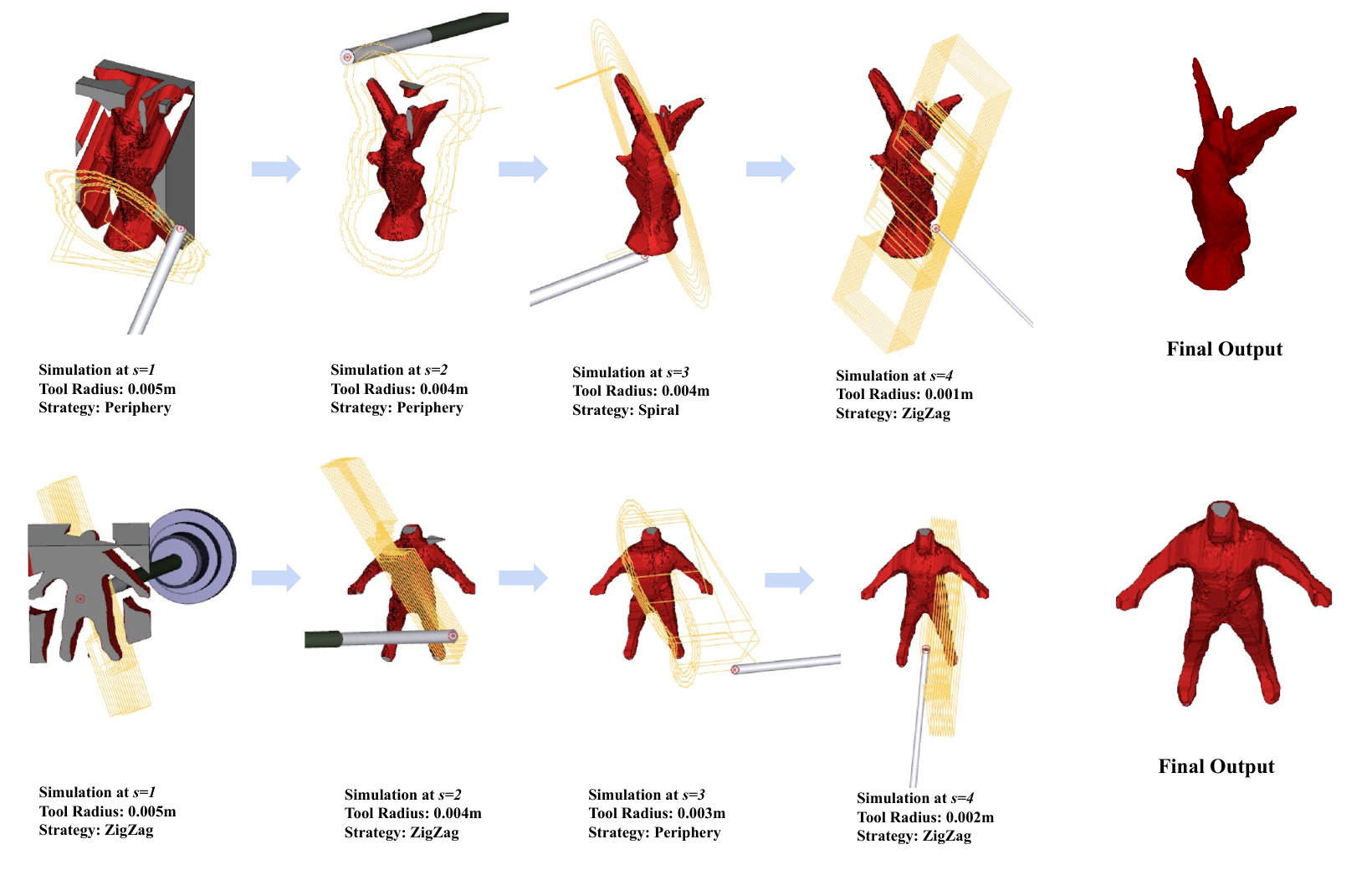}
    \caption{\textbf{Analysis on complex shapes.} The greedy module first fixes the indexed orientations, and at each subsequent simulation step, the policy selects radius, layer height, and clearing strategy for the current orientation.}
    \label{fig:complex_shape}
\end{figure}

Figure~\ref{fig:complex_shape} illustrates the sequential simulation on two complex shapes. The orientation order is produced by the separate greedy module, and the learned policy then chooses radius, layer height, and clearing strategy at each indexed view. The examples exhibit a coarse-to-fine pattern, with larger tools earlier and smaller tools near constrained structures. This is qualitative behavior in these examples rather than a guaranteed scheduling rule.

\section{Voxel Resolution and Laplacian Smoothing}

\begin{table}[t]
\centering
\small
\setlength{\tabcolsep}{4pt}
\begin{tabular}{lcccccc}
\toprule
\textbf{Resolution} & \textbf{Smooth} & \textbf{IoU}$\uparrow$ & \textbf{F1}$\uparrow$ & \textbf{CD}$\downarrow$ & \textbf{NC}$\uparrow$ & \textbf{T}$\downarrow$ \\
\midrule
$64^3$  & \cmark & 0.810 & 0.886 & 0.686 & 0.912 & \textbf{397.6} \\
$128^3$ & \cmark & 0.824 & 0.889 & 0.585 & 0.912 & 409.3 \\
$256^3$ & \cmark & \textbf{0.848} & \textbf{0.902} & \textbf{0.493} & \textbf{0.914} & 423.1 \\
$256^3$ & \xmark & 0.826 & 0.898 & 0.530 & 0.893 & 442.5 \\
\bottomrule
\end{tabular}
\caption{Ablation on voxel resolution and Laplacian smoothing. T is the common normalized simulator-time proxy.}
\label{tab:abl_res_smooth}
\end{table}

Table~\ref{tab:abl_res_smooth} is an archived standalone ABC resolution study. Its run metadata do not establish a controlled replication of the fine-tuned Table~\ref{tab:quant} condition, so the values are used only for paired comparisons within Table~\ref{tab:abl_res_smooth}, not across tables. The grid resolution determines how finely the target geometry is discretized during toolpath generation.

As the resolution increases from $64^3$ to $256^3$, IoU and F1 increase and CD decreases, indicating that higher-resolution grids let the generated toolpaths better approximate the target geometry, while NC stays nearly unchanged. We further compare the effect of Laplacian smoothing at the highest resolution, $256^3$. Applying smoothing improves all geometric metrics over the unsmoothed variant, achieving better IoU, F1, CD, and NC. This indicates that smoothing helps remove local artifacts and produce more accurate and coherent toolpaths.

Toolpaths are generated and simulated on a $256^3$ grid over the normalized bounding cube. For a part whose largest dimension is $100$\,mm this corresponds to a voxel of roughly $0.39$\,mm, which is coarse relative to the $\pm0.01$--$0.05$\,mm tolerances expected of finishing operations. The method should therefore be read as targeting roughing, and a conventional finishing pass is complementary and out of scope. High one-sided gouging compliance is compatible with this scope but does not establish that residual material dominates the unreported two-sided error distribution.

\section{Baseline Protocols}
\label{sec:supp:baselines}

\subsection{Why CAD Reconstruction Methods Are Included}
CAD reconstruction methods are not G-code generators. We include them as geometric context for the mesh-to-CAD stage of a conventional mesh $\rightarrow$ CAD $\rightarrow$ CAM $\rightarrow$ G-code route. The comparison is not symmetric, since CAD methods emit smooth analytic or implicit surfaces, whereas a carved result carries discretization and toolpath structure. CD and NC can therefore favor CAD outputs independently of downstream machining behavior, which is why the machining-oriented comparisons are reported separately.

\subsection{MeshCAM Configurations}
Both MeshCAM configurations start from the same triangulated mesh given to our method, with no CAD model and no feature metadata. The non-expert configuration uses one top setup and a 6\,mm-diameter flat end mill (3\,mm radius, our median radius at 100\,mm scale) for parallel roughing and finishing. In the expert configuration, a machinist tunes the orientation, tool size, and step size per toolpath strategy for each part. Per part, the machinist chooses 1--5 axis-aligned setups (mean 2.0, versus 1.8 views for ours), parallel roughing with flat end mills (mostly 20\,mm diameter, i.e., our largest 10\,mm radius), and parallel, waterline, or pencil finishing (mostly 4--8\,mm diameter). All MeshCAM tools are flat end mills, as simulated, and the complete MeshCAM G-code is mapped to our normalized frame before replay. Both configurations use the conventional roughing and parallel-finishing strategies exposed by the tool, and adaptive or trochoidal high-speed-machining strategies are outside the compared set. Because the expert configuration depends on operator skill and effort, it should be read as a strong but not uniquely defined baseline rather than a reproducible protocol.

\section{Ablations for Tool Orientation Selection}

\begin{table}[t]
    \centering
    \small
    \setlength{\tabcolsep}{6pt}
    \begin{tabular}{lcccc}
        \toprule
        \textbf{Method} & \textbf{IoU}$\uparrow$ & \textbf{F1}$\uparrow$ & \textbf{CD}$\downarrow$ & \textbf{NC}$\uparrow$ \\
        \midrule
        6-axes   & 0.792 & 0.876 & 0.711 & 0.881 \\
        Rand(5)  & 0.740 & 0.839 & 0.710 & 0.869 \\
        Rand(6)  & 0.751 & 0.846 & 0.700 & 0.873 \\
        \midrule
        Ours     & \textbf{0.848} & \textbf{0.912} & \textbf{0.460} & \textbf{0.922} \\
        \bottomrule
    \end{tabular}
    \caption{\textbf{Comparison of tool orientation strategies.}}
    \label{tab:abl_approaches}
\end{table}

We evaluate our orientation selection on the ABC dataset. Following Section~\ref{subsec:tool_ori}, we compare it with Random-5, Random-6, and six principal-axis views. As shown in Table~\ref{tab:abl_approaches}, our selection outperforms all baselines on IoU, F1, CD, and NC.

The orientation set is produced by an iterative greedy procedure that repeatedly adds the direction covering the most currently uncovered surface points. This yields a small set that achieves the coverage target in practice, but it is a heuristic for a set-cover-type problem and is not guaranteed to be of minimum cardinality.
Random-5 and Random-6 are point-estimate controls rather than uncertainty estimates.

\section{Pre-training Time}

\begin{table}[t]
\centering
\small
\setlength{\tabcolsep}{5pt}
\begin{tabular}{lcc}
\toprule
\textbf{Method} & \textbf{Training budget} & \textbf{Total pretraining time (h)} \\
\midrule
CSG-Stump  & 1000 epochs  & 80 \\
ExtrudeNet & 1000 epochs  & 60 \\
CAPRI-Net  & 1000 epochs  & 9  \\
SECAD-Net  & 1000 epochs  & 8  \\
CNC-Net    & 100 epochs   & 50 \\
\textbf{Ours} & 2,000 episodes & 30 \\
\bottomrule
\end{tabular}
\caption{Comparison of training budgets and total pretraining time across different methods.}
\label{tab:training_budget}
\end{table}

Table~\ref{tab:training_budget} reports the training budgets and total pretraining time for several CAD reconstruction methods and our approach. Most existing methods are trained using supervised learning for up to 1000 epochs, leading to training times ranging from several hours to tens of hours depending on model complexity. All pretraining times are measured on a single NVIDIA Quadro RTX 8000 GPU. Our method is trained using reinforcement learning for 2,000 episodes and requires approximately 30 hours of pretraining. This training time is shorter than several prior approaches such as CSG-Stump, ExtrudeNet, and CNC-Net, while remaining within a practical computational budget.

\section{Per-Sample Fine-Tuning Time}

\begin{table}[t]
    \centering
    \small
    \begin{tabular}{lc}
        \toprule
        \textbf{Method} & \textbf{Fine-tuning time per sample} \\
        \midrule
        CSG-Stump   & 60 min \\
        ExtrudeNet  & 30 min \\
        CAPRI-Net   & 3 min  \\
        SECAD-Net   & 3 min  \\
        CNC-Net     & 50 min \\
        \textbf{Ours} & \textbf{10 min} \\
        \bottomrule
    \end{tabular}
    \caption{\textbf{Per-sample fine-tuning time comparison.}}
    \label{tab:finetune_time}
\end{table}

As described in Section~\ref{sec:results}, our method does not require per-sample fine-tuning. Because most prior works adopt per-sample tuning to improve fidelity, we also evaluate our method under this setting. Table~\ref{tab:finetune_time} summarizes the per-sample fine-tuning time. Our method requires 10 minutes (5 episodes) on a single NVIDIA Quadro RTX 8000 GPU, placing it between lightweight CAD-based approaches such as CAPRI-Net and SECAD-Net (approximately 3 minutes) and optimization-heavy baselines including CSG-Stump (60 minutes), CNC-Net (50 minutes), and ExtrudeNet (30 minutes).

\section{Compact Selector Diagnostic}
\label{sec:supp:compact_selector}

\paragraph{Protocol and scope.}
We retain an additional archived diagnostic to examine informed alternatives to learned selection. It is a separate $128^3$ cached-simulation study, not a replication's $256^3$ evaluation. It uses 12 actions and two decisions, permitting enumeration of 144 action sequences. The diagnostic policy observes target points, current stock points, and progress. Unlike the main target-only selector, it is stock-aware. Development uses 15 training and three validation shapes. Each of five seeds trains for 50 epochs, selecting a checkpoint by validation mean terminal IoU. The fixed sequence is selected on the three validation shapes. The held-out cohort contains ten of 46 candidate shapes for which all actions completed, and this eligibility condition can favor easier geometry and limits generalization.

\begin{table}[t]
\centering\small
\setlength{\tabcolsep}{4pt}
\begin{tabular}{lcc}
\toprule
Selector & Terminal IoU$\uparrow$ & Time proxy$\downarrow$ \\
\midrule
Uniform random (expectation) & .5938 & 884.7 \\
Learned (five seeds) & .7457 & 1258.5 \\
Frozen heuristic & .7517 & 1492.8 \\
Validation-selected fixed & .7682 & 945.5 \\
Retrospective oracle & .7834 & 1057.3 \\
\bottomrule
\end{tabular}
\caption{Separate compact selector diagnostic on ten eligible held-out shapes. Time is path length divided by feed rate, not measured machine seconds. The oracle maximizes terminal IoU over all 144 sequences per shape.}
\label{tab:supp:compact_selector}
\end{table}

\paragraph{Results and uncertainty.}
Table~\ref{tab:supp:compact_selector} reports the archived aggregate results. Learned exceeds uniform random by 0.1519 IoU (95\% hierarchical seed-and-shape interval $[0.1001,0.1984]$). Conditional on the selected fixed sequence, Learned--Fixed is $-0.0225$ ($[-0.0538,-0.0020]$), and Learned--Heuristic is $-0.0060$ ($[-0.0307,0.0144]$). A separate 20,000-draw selection-aware bootstrap resamples the three validation shapes, reselects among 144 sequences, samples one of five learned seeds, and resamples the ten held-out shapes. Its mean Learned--Selected Fixed gap is $-0.0129$, with interval $[-0.0668,0.0351]$. This is a resampling summary, not the direct difference between the two table means, and the selection-aware gap is unresolved.

\paragraph{Search and interpretation.}
Retrospective best-of-$K$ search reaches 0.7371, 0.7606, and 0.7722 IoU for $K=4,8,16$, respectively, compared with the exhaustive oracle's 0.7834. These are not equal-cost comparisons. Learned is not Pareto-optimal among the reported operating points, as the validation-selected fixed sequence has higher IoU and lower time proxy. The diagnostic supports improvement over uninformed selection in this compact setting, not a necessity or superiority claim for learning. Its observation, action space, resolution, and selected cohort differ from the main experiment, so the values must not be pooled with main-paper results.

\end{document}

%% file: main_evidence.tex
\begin{table}[t]
\centering\small
\setlength{\tabcolsep}{4pt}
\begin{tabular}{lcccc}
\toprule
\textbf{Method} & \textbf{IoU}$\uparrow$ & \textbf{F1}$\uparrow$ & \textbf{CD}$\downarrow$ & \textbf{NC}$\uparrow$ \\
\midrule
CSG-Stump & 0.751 & 0.856 & 0.645 & 0.858 \\
ExtrudeNet & 0.735 & 0.852 & 0.743 & 0.842 \\
CAPRI-Net & 0.728 & 0.846 & 0.624 & 0.883 \\
SECAD-Net & 0.742 & 0.845 & 0.765 & 0.872 \\
CNC-Net & 0.772 & 0.884 & 1.204 & 0.858 \\
\oursrow Ours & \textbf{0.827} & \textbf{0.899} & \textbf{0.525} & \textbf{0.910} \\
\bottomrule
\end{tabular}
\caption{\textbf{Untuned evaluation on all 1,000 ABC test shapes.} No method uses per-shape tuning.}
\label{tab:large_main}
\vspace{-3mm}
\end{table}

\noindent\textbf{Large-scale untuned evaluation.}
On all 1,000 ABC test shapes without any per-shape tuning (Table~\ref{tab:large_main}), Geo2Gcode leads on all four metrics, with 0.827 IoU versus 0.772 for CNC-Net and a 56\% lower CD.

\begin{table}[t]
    \centering
    \small
    \setlength{\tabcolsep}{3pt}
    \begin{tabular}{lccccc}
        \toprule
        \textbf{Parameter policy} & \textbf{IoU}$\uparrow$ & \textbf{F1}$\uparrow$ & \textbf{CD}$\downarrow$ & \textbf{NC}$\uparrow$ & \textbf{T}$\downarrow$ \\
        \midrule
        Random         & 0.781 & 0.863 & 0.905 & 0.521 & 533.5 \\
        Fixed (fine)   & \textbf{0.856} & \textbf{0.917} & \textbf{0.427} & \textbf{0.936} & 1634.0 \\
        Fixed (coarse) & 0.718 & 0.818 & 0.666 & 0.853 & \textbf{129.1} \\
        Fixed (const.\ triple) & 0.788 & 0.897 & 0.583 & 0.901 & 1219.0 \\
        \midrule
        Fixed (const.\ triple, alt.) & 0.793 & 0.872 & 0.574 & 0.923 & 627.4 \\
        Greedy sim.\ search & 0.821 & 0.907 & 0.471 & 0.918 & 641.3 \\
        \midrule
        \oursrow Learned (no tuning) & 0.837 & 0.907 & 0.493 & 0.914 & 451.3 \\
        \oursrow Learned (5-ep.\ tuning) & 0.848 & 0.912 & 0.460 & 0.922 & 423.1 \\
        \bottomrule
    \end{tabular}
    \caption{\textbf{Parameter-selection policies on 50 ABC shapes.} All rows share the same orientations, path generator, and simulator, and only the per-view parameter choice differs. $T$ is the normalized simulator cutting-time proxy. Alt.: another point of the same constant-triple sweep. Greedy: per view, simulate all 42 actions and keep the one with the highest per-view reward (Eq.~\ref{eq:reward}).}
    \label{tab:parameter_controls}
\vspace{-3mm}
\end{table}

\subsection{Effect of Learned Parameter Selection}
\label{sec:param_select}
Table~\ref{tab:parameter_controls} isolates the selector by holding orientations, compiler, and simulator fixed and changing only the per-view choice of radius, layer height, and clearing pattern. Random samples the action space per view. Fixed (fine/coarse) always use the smallest/largest radius and layer height. Fixed (const.\ triple) is a reported point of a sweep that holds one radius--height--pattern triple constant across all views and shapes.

Even Random parameters reach 0.781 IoU, so most fidelity comes from the compiler, while its NC (0.521) is near the 0.5 of uncorrelated normals. Without any per-shape tuning, the learned selector outperforms Random and the constant-triple setting on every metric while also cutting faster. Among the reported fixed settings, it gives the best balance of fidelity and toolpath length. Fixed-fine remains the most accurate, but the selector retains 97.8\% of its IoU at about a quarter of its cutting time, whereas Fixed-coarse is faster but far less accurate. Five-episode adaptation narrows the remaining gap. Greedy per-view search, which maximizes each view's reward in isolation, reaches lower IoU at higher $T$ than the untuned selector, because a choice that is best for one view can leave poor stock for later views. Crediting each action with its effect on later views avoids this myopia, even though the selector does not observe the stock.

%% file: main.bbl
\begin{thebibliography}{40}
\providecommand{\natexlab}[1]{#1}
\providecommand{\url}[1]{\texttt{#1}}
\expandafter\ifx\csname urlstyle\endcsname\relax
  \providecommand{\doi}[1]{doi: #1}\else
  \providecommand{\doi}{doi: \begingroup \urlstyle{rm}\Url}\fi

\bibitem[Balic and Korosec(2002)]{BALIC20021171}
J Balic and M Korosec.
\newblock Intelligent tool path generation for milling of free surfaces using
  neural networks.
\newblock \emph{Int. J. Mach. Tools Manuf.}, 2002.

\bibitem[Chang et~al.(2015)Chang, Funkhouser, Guibas, Hanrahan, Huang, Li,
  Savarese, Savva, Song, Su, et~al.]{Shapenet}
Angel~X Chang, Thomas Funkhouser, Leonidas Guibas, Pat Hanrahan, Qixing Huang,
  Zimo Li, Silvio Savarese, Manolis Savva, Shuran Song, Hao Su, et~al.
\newblock {ShapeNet}: An information-rich {3D} model repository.
\newblock \emph{arXiv preprint arXiv:1512.03012}, 2015.

\bibitem[Chung et~al.(2021)Chung, Kim, Knyazev, Lee, Taylor, Park, and
  Cho]{Brick-by-brick}
Hyunsoo Chung, Jungtaek Kim, Boris Knyazev, Jinhwi Lee, Graham~W Taylor, Jaesik
  Park, and Minsu Cho.
\newblock Brick-by-brick: Combinatorial construction with deep reinforcement
  learning.
\newblock In \emph{Advances in Neural Information Processing Systems}, pages
  5745--5757, 2021.

\bibitem[{CIMCO A/S}(2025)]{cimco}
{CIMCO A/S}.
\newblock {CIMCO Edit 2025 User Guide}.
\newblock Product documentation, 2025.

\bibitem[Deng et~al.(2020)Deng, Genova, Yazdani, Bouaziz, Hinton, and
  Tagliasacchi]{Cvxnet}
Boyang Deng, Kyle Genova, Soroosh Yazdani, Sofien Bouaziz, Geoffrey Hinton, and
  Andrea Tagliasacchi.
\newblock {CvxNet}: Learnable convex decomposition.
\newblock In \emph{CVPR}, 2020.

\bibitem[Dittrich et~al.(2019)Dittrich, Uhlich, and Denkena]{DITTRICH201949}
Marc-Andr{\'e} Dittrich, Florian Uhlich, and Berend Denkena.
\newblock Self-optimizing tool path generation for 5-axis machining processes.
\newblock \emph{CIRP-JMST}, 2019.

\bibitem[Du et~al.(2018)Du, Inala, Pu, Spielberg, Schulz, Rus, Solar-Lezama,
  and Matusik]{du2018}
Tao Du, Jeevana~Priya Inala, Yewen Pu, Andrew Spielberg, Adriana Schulz,
  Daniela Rus, Armando Solar-Lezama, and Wojciech Matusik.
\newblock {InverseCSG}: Automatic conversion of {3D} models to {CSG} trees.
\newblock \emph{ACM Trans. Graph.}, 37\penalty0 (6):\penalty0 213:1--213:16,
  2018.

\bibitem[Foley et~al.(1996)Foley, van Dam, Feiner, and
  Hughes]{foley1996computer}
James~D. Foley, Andries van Dam, Steven~K. Feiner, and John~F. Hughes.
\newblock \emph{Computer Graphics: Principles and Practice}.
\newblock Addison-Wesley Professional, 2nd edition, 1996.

\bibitem[Genova et~al.(2019)Genova, Cole, Vlasic, Sarna, Freeman, and
  Funkhouser]{genova2019learning}
Kyle Genova, Forrester Cole, Daniel Vlasic, Aaron Sarna, William~T Freeman, and
  Thomas Funkhouser.
\newblock Learning shape templates with structured implicit functions.
\newblock In \emph{ICCV}, 2019.

\bibitem[Hong et~al.(2023)Hong, Yavartanoo, Neshatavar, and Lee]{ACL-SPC}
Sangmin Hong, Mohsen Yavartanoo, Reyhaneh Neshatavar, and Kyoung~Mu Lee.
\newblock {ACL-SPC}: Adaptive closed-loop system for self-supervised point
  cloud completion.
\newblock In \emph{CVPR}, pages 9435--9444, 2023.

\bibitem[Hong et~al.(2026)Hong, Lee, and Lee]{StarryGazer}
Sangmin Hong, Suyoung Lee, and Kyoung~Mu Lee.
\newblock {StarryGazer}: Leveraging monocular depth estimation models for
  domain-agnostic single depth image completion.
\newblock \emph{Signal, Image and Video Processing}, 20:\penalty0 598, 2026.

\bibitem[Hong et~al.(2024)Hong, Zhang, Gu, Bi, Zhou, Liu, Liu, Sunkara, Bui,
  and Tan]{LRM}
Yicong Hong, Kai Zhang, Jiuxiang Gu, Sai Bi, Yang Zhou, Difan Liu, Feng Liu,
  Kalyan Sunkara, Trung Bui, and Hao Tan.
\newblock {LRM}: Large reconstruction model for single image to {3D}.
\newblock In \emph{ICLR}, 2024.

\bibitem[Hsieh and Chu(2013)]{HSIEH20133}
Hsin-Ta Hsieh and Chih-Hsing Chu.
\newblock Improving optimization of tool path planning in 5-axis flank milling
  using advanced pso algorithms.
\newblock \emph{Robot. Comput.-Integr. Manuf.}, 2013.

\bibitem[Huang et~al.(2023)Huang, Zhang, Chen, Li, Zhang, and
  Ni]{huang2023learning}
Xiaoyang Huang, Yi Zhang, Kai Chen, Teng Li, Wenjun Zhang, and Bingbing Ni.
\newblock Learning shape primitives via implicit convexity regularization.
\newblock In \emph{ICCV}, 2023.

\bibitem[Huang et~al.(2024)Huang, Guo, Su, Han, Ding, Zhang, Liu, Wang, Fang,
  Song, Whiting, and Wang]{huang2024toolpath}
Yuming Huang, Yuhu Guo, Renbo Su, Xingjian Han, Junhao Ding, Tianyu Zhang, Tao
  Liu, Weiming Wang, Guoxin Fang, Xu Song, Emily Whiting, and Charlie C.~L.
  Wang.
\newblock Learning based toolpath planner on diverse graphs for 3d printing.
\newblock \emph{ACM Transactions on Graphics (TOG)}, 43\penalty0 (6), 2024.

\bibitem[Kania et~al.(2020)Kania, Zi{\k{e}}ba, and Kajdanowicz]{UCSG}
Kacper Kania, Maciej Zi{\k{e}}ba, and Tomasz Kajdanowicz.
\newblock {UCSG-NET} -- unsupervised discovering of constructive solid geometry
  tree.
\newblock In \emph{Advances in Neural Information Processing Systems}, pages
  8776--8786, 2020.

\bibitem[Koch et~al.(2019)Koch, Matveev, Jiang, Williams, Artemov, Burnaev,
  Alexa, Zorin, and Panozzo]{Abc}
Sebastian Koch, Albert Matveev, Zhongshi Jiang, Francis Williams, Alexey
  Artemov, Evgeny Burnaev, Marc Alexa, Denis Zorin, and Daniele Panozzo.
\newblock {ABC}: A big {CAD} model dataset for geometric deep learning.
\newblock In \emph{CVPR}, 2019.

\bibitem[Kukreja and Pande(2023)]{kukreja2022optimum}
Aman Kukreja and Sanjay~S Pande.
\newblock Optimal toolpath planning strategy prediction using machine learning
  technique.
\newblock \emph{Eng. Appl. Artif. Intell.}, 2023.

\bibitem[Laidlaw et~al.(1986)Laidlaw, Trumbore, and Hughes]{CSG}
David~H Laidlaw, W~Benjamin Trumbore, and John~F Hughes.
\newblock Constructive solid geometry for polyhedral objects.
\newblock In \emph{SIGGRAPH}, 1986.

\bibitem[Li et~al.(2019)Li, Sung, Dubrovina, Yi, and Guibas]{SPFN}
Lingxiao Li, Minhyuk Sung, Anastasia Dubrovina, Li Yi, and Leonidas~J. Guibas.
\newblock Supervised fitting of geometric primitives to {3D} point clouds.
\newblock In \emph{CVPR}, 2019.

\bibitem[Li et~al.(2023)Li, Guo, Zhang, and Yan]{SECAD-Net}
Pu Li, Jianwei Guo, Xiaopeng Zhang, and Dong-Ming Yan.
\newblock Secad-net: Self-supervised cad reconstruction by learning
  sketch-extrude operations.
\newblock In \emph{CVPR}, 2023.

\bibitem[Lorensen and Cline(1987)]{LorensenCline1987MarchingCubes}
William~E. Lorensen and Harvey~E. Cline.
\newblock Marching cubes: A high resolution 3d surface construction algorithm.
\newblock In \emph{Proceedings of the 14th Annual Conference on Computer
  Graphics and Interactive Techniques (SIGGRAPH '87)}, pages 163--169, New
  York, NY, USA, 1987. ACM.

\bibitem[Mescheder et~al.(2019)Mescheder, Oechsle, Niemeyer, Nowozin, and
  Geiger]{onet}
Lars Mescheder, Michael Oechsle, Michael Niemeyer, Sebastian Nowozin, and
  Andreas Geiger.
\newblock Occupancy networks: Learning 3d reconstruction in function space.
\newblock In \emph{CVPR}, 2019.

\bibitem[Mnih et~al.(2013)Mnih, Kavukcuoglu, Silver, Graves, Antonoglou,
  Wierstra, and Riedmiller]{DQN}
Volodymyr Mnih, Koray Kavukcuoglu, David Silver, Alex Graves, Ioannis
  Antonoglou, Daan Wierstra, and Martin Riedmiller.
\newblock Playing atari with deep reinforcement learning.
\newblock In \emph{NIPS Deep Learning Workshop}, 2013.
\newblock arXiv:1312.5602.

\bibitem[Niu et~al.(2018)Niu, Li, and Xu]{niu2018im2struct}
Chengjie Niu, Jun Li, and Kai Xu.
\newblock {Im2Struct}: Recovering {3D} shape structure from a single {RGB}
  image.
\newblock In \emph{CVPR}, 2018.

\bibitem[Paschalidou et~al.(2021)Paschalidou, Katharopoulos, Geiger, and
  Fidler]{Neural_parts}
Despoina Paschalidou, Angelos Katharopoulos, Andreas Geiger, and Sanja Fidler.
\newblock Neural parts: Learning expressive 3d shape abstractions with
  invertible neural networks.
\newblock In \emph{CVPR}, 2021.

\bibitem[Ren et~al.(2021)Ren, Zheng, Cai, Li, Jiang, Cai, Zhang, Pan, Zhang,
  Zhao, and Yi]{CSG-Stump}
Daxuan Ren, Jianmin Zheng, Jianfei Cai, Jiatong Li, Haiyong Jiang, Zhongang
  Cai, Junzhe Zhang, Liang Pan, Mingyuan Zhang, Haiyu Zhao, and Shuai Yi.
\newblock {CSG-Stump}: A learning friendly {CSG}-like representation for
  interpretable shape parsing.
\newblock In \emph{ICCV}, 2021.

\bibitem[Ren et~al.(2022)Ren, Zheng, Cai, Li, and Zhang]{ExtrudeNet}
Daxuan Ren, Jianmin Zheng, Jianfei Cai, Jiatong Li, and Junzhe Zhang.
\newblock Extrudenet: Unsupervised inverse sketch-and-extrude for shape
  parsing.
\newblock In \emph{ECCV}, 2022.

\bibitem[Sharma et~al.(2018)Sharma, Goyal, Liu, Kalogerakis, and Maji]{Csgnet}
Gopal Sharma, Rishabh Goyal, Difan Liu, Evangelos Kalogerakis, and Subhransu
  Maji.
\newblock {CSGNet}: Neural shape parser for constructive solid geometry.
\newblock In \emph{CVPR}, 2018.

\bibitem[Sharma et~al.(2020)Sharma, Liu, Maji, Kalogerakis, Chaudhuri, and
  M{\v{e}}ch]{ParseNet}
Gopal Sharma, Difan Liu, Subhransu Maji, Evangelos Kalogerakis, Siddhartha
  Chaudhuri, and Radom{\'\i}r M{\v{e}}ch.
\newblock {ParSeNet}: A parametric surface fitting network for {3D} point
  clouds.
\newblock In \emph{ECCV}, 2020.

\bibitem[Tulsiani et~al.(2017)Tulsiani, Su, Guibas, Efros, and
  Malik]{Tulsiani_2017_CVPR}
Shubham Tulsiani, Hao Su, Leonidas~J Guibas, Alexei~A Efros, and Jitendra
  Malik.
\newblock Learning shape abstractions by assembling volumetric primitives.
\newblock In \emph{CVPR}, 2017.

\bibitem[Wang et~al.(2018)Wang, Zhang, Li, Fu, Liu, and Jiang]{Pixel2Mesh}
Nanyang Wang, Yinda Zhang, Zhuwen Li, Yanwei Fu, Wei Liu, and Yu-Gang Jiang.
\newblock {Pixel2Mesh}: Generating {3D} mesh models from single {RGB} images.
\newblock In \emph{ECCV}, pages 52--67, 2018.

\bibitem[Wang et~al.(2019)Wang, Sun, Liu, Sarma, Bronstein, and Solomon]{dgcnn}
Yue Wang, Yongbin Sun, Ziwei Liu, Sanjay~E. Sarma, Michael~M. Bronstein, and
  Justin~M. Solomon.
\newblock Dynamic graph cnn for learning on point clouds.
\newblock \emph{ACM Transactions on Graphics (TOG)}, 2019.

\bibitem[Woo(1994)]{Woo1994Visibility}
Tony~C. Woo.
\newblock Visibility maps and spherical algorithms.
\newblock \emph{Computer-Aided Design}, 26\penalty0 (1):\penalty0 6--11, 1994.

\bibitem[Yavartanoo et~al.(2021)Yavartanoo, Chung, Neshatavar, and Lee]{3dias}
Mohsen Yavartanoo, Jaeyoung Chung, Reyhaneh Neshatavar, and Kyoung~Mu Lee.
\newblock {3DIAS}: {3D} shape reconstruction with implicit algebraic surfaces.
\newblock In \emph{ICCV}, 2021.

\bibitem[Yavartanoo et~al.(2024{\natexlab{a}})Yavartanoo, Hong, Neshatavar, and
  Lee]{CNC-Net}
Mohsen Yavartanoo, Sangmin Hong, Reyhaneh Neshatavar, and Kyoung~Mu Lee.
\newblock Cnc-net: Self-supervised learning for cnc machining operations.
\newblock In \emph{CVPR}, 2024{\natexlab{a}}.

\bibitem[Yavartanoo et~al.(2024{\natexlab{b}})Yavartanoo, Hong, Neshatavar, and
  Lee]{Text2CAD}
Mohsen Yavartanoo, Sangmin Hong, Reyhaneh Neshatavar, and Kyoung~Mu Lee.
\newblock {Text2CAD}: Text to {3D} {CAD} generation via technical drawings.
\newblock \emph{arXiv preprint arXiv:2411.06206}, 2024{\natexlab{b}}.

\bibitem[Yu et~al.(2022)Yu, Chen, Li, Sanghi, Shayani, Mahdavi-Amiri, and
  Zhang]{CAPRI-Net}
Fenggen Yu, Zhiqin Chen, Manyi Li, Aditya Sanghi, Hooman Shayani, Ali
  Mahdavi-Amiri, and Hao Zhang.
\newblock {CAPRI-Net}: Learning compact {CAD} shapes with adaptive primitive
  assembly.
\newblock In \emph{CVPR}, 2022.

\bibitem[Zhong et~al.(2025)Zhong, Wang, Wang, Lu, and Zhao]{zhong2025deepmill}
Fanchao Zhong, Yang Wang, Peng-Shuai Wang, Lin Lu, and Haisen Zhao.
\newblock {DeepMill}: Neural accessibility learning for subtractive
  manufacturing.
\newblock In \emph{SIGGRAPH}, 2025.

\bibitem[Zou et~al.(2017)Zou, Yumer, Yang, Ceylan, and Hoiem]{Zou_2017_ICCV}
Chuhang Zou, Ersin Yumer, Jimei Yang, Duygu Ceylan, and Derek Hoiem.
\newblock {3D-PRNN}: Generating shape primitives with recurrent neural
  networks.
\newblock In \emph{ICCV}, 2017.

\end{thebibliography}
